\documentclass[twoside,11pt]{article}

\usepackage{amsfonts}

\usepackage{jmlr2e_x}
\usepackage{times}

\usepackage{amsmath}

 \usepackage{epstopdf}
\usepackage{multirow,bigdelim}
 \usepackage{color}
\newfont{\bboard}{msbm10 scaled\magstephalf}
\def\real{\mbox{\bboard R}}

\def\boardE{\mbox{\bboard E}}

\def\boardZ{\mbox{\bboard Z}}

\newfont{\bboardscript}{msbm8 scaled\magstephalf}
\newcommand{\EVscript}{\mbox{\bboardscript E}}

\newcommand{\argmin}{ \mathop{\mathrm{argmin}} }
\newcommand{\argmax}{ \mathop{\mathrm{argmax}} }
\newcommand{\EV}{\boardE}

\newcommand{\KLp}[2]{\mathrm{KL}\left(#1\left\|#2\right.\right)}

\newcommand{\tr}{\mathrm{tr}}
\newcommand{\detbar}[1]{\left|#1\right|}
\newcommand{\norm}[1]{\left\|#1\right\|}
\newcommand{\Normal}[2]{{\cal N}\left(#1,#2\right)}
\newcommand{\Normalv}[3]{{\cal N}\left(#1\middle|#2,#3\right)}
\newcommand{\Normalvv}[3]{{\cal N}(#1|#2,#3)}

\newcommand{\mbf}[1]{\mathbf{#1}}
\newcommand{\diag}{\mathrm{diag}}

\newcommand{\pdd}[2]{\frac{\partial #1}{\partial #2}}

\newcommand{\var}{\mathrm{var}}

\newcommand{\GP}{{\cal GP}}

\newcommand{\refeqn}[1]{(\ref{#1})}

\newcommand{\comments}[1]{}
\newlength{\myw}

\newcommand{\va}{\mbf{a}}

\newcommand{\vx}{\mbf{x}}
\newcommand{\vf}{\mbf{f}}

\newcommand{\vy}{\mbf{y}}

\newcommand{\vk}{\mbf{k}}
\newcommand{\vz}{\mbf{z}}
\newcommand{\vm}{\mbf{m}}
\newcommand{\vt}{\mbf{t}}
\newcommand{\vu}{\mbf{u}}
\newcommand{\vv}{\mbf{v}}
\newcommand{\vone}{\mbf{1}}
\newcommand{\vzero}{\mbf{0}}
\newcommand{\valpha}{\boldsymbol{\alpha}}
\newcommand{\vgamma}{\boldsymbol{\gamma}}
\newcommand{\vtheta}{\boldsymbol{\theta}}
\newcommand{\veta}{\boldsymbol{\eta}}
\newcommand{\vbeta}{\boldsymbol{\beta}}
\newcommand{\vlambda}{\boldsymbol{\lambda}}
\newcommand{\vmu}{\boldsymbol{\mu}}

\newcommand{\mA}{\mbf{A}}
\newcommand{\mB}{\mbf{B}}
\newcommand{\mC}{\mbf{C}}
\newcommand{\mD}{\mbf{D}}

\newcommand{\mX}{\mbf{X}}
\newcommand{\mW}{\mbf{W}}
\newcommand{\mV}{\mbf{V}}
\newcommand{\mK}{\mbf{K}}

\newcommand{\mI}{\mbf{I}}

\newcommand{\mSigma}{\boldsymbol{\Sigma}}

\newcommand{\mLambda}{\boldsymbol{\Lambda}}

\newcommand{\ttheta}{\tilde{\theta}}
\newcommand{\teta}{\tilde{\eta}}
\newcommand{\tveta}{\tilde{\veta}}
\newcommand{\tZ}{\tilde{Z}}
\newcommand{\tmu}{\tilde{\mu}}
\newcommand{\tsigma}{\tilde{\sigma}}
\newcommand{\tvmu}{\tilde{\vmu}}

\newcommand{\tmSigma}{\tilde{\mSigma}}
\newcommand{\tv}{\tilde{v}}
\newcommand{\tu}{\tilde{u}}
\newcommand{\tw}{\tilde{w}}
\newcommand{\tvu}{\tilde{\vu}}
\newcommand{\tmW}{\tilde{\mW}}
\newcommand{\tvt}{\tilde{\vt}}
\newcommand{\tvv}{\tilde{\vv}}

\newcommand{\noti}{{\neg i}}

\newcommand{\calL}{{\cal L}}

\newcommand{\calE}{{\cal E}}

\newcommand{\sd}[1]{\dot{#1}}
\newcommand{\sdd}[1]{\ddot{#1}}
\newcommand{\sddd}[1]{\dddot{#1}}

\newcommand{\abc}[1]{\textcolor{red}{#1}}

\newcommand{\NOTE}[1]{\textcolor{red}{[NOTE: #1]}}

\newcommand{\subsubsubsection}[1]{\vspace{0.1in}\noindent {\bf #1} --}

\newtheorem{claim}{Claim}

\newcommand{\shape}{$_\mathrm{sh}$}
\newcommand{\scale}{$_\mathrm{sc}$}
\newcommand{\NLP}{NLP}

\jmlrheading{1}{July 12, 2013}{1-48}{7/2013}{??/??}{Lifeng Shang and Antoni B.
Chan}

\ShortHeadings{On Approximate Inference for Generalized Gaussian
Process Models}{Shang and Chan} \firstpageno{1}

\begin{document}

\title{On Approximate Inference for Generalized Gaussian Process Models}

\author{\name Lifeng~Shang \email lshang@cityu.edu.hk \\
       \addr Department of Computer Science\\
       City University of Hong Kong\\
       \AND
       \name Antoni~B.~Chan \email abchan@cityu.edu.hk \\
       \addr Department of Computer Science\\
       City University of Hong Kong
       \AND
      {\rm (This manuscript was submitted for review on July 12, 2013)}
       }

\editor{}

\maketitle

\begin{abstract}A generalized Gaussian process model (GGPM) is a unifying framework
that encompasses many existing Gaussian process (GP) models, such as
GP regression, classification, and counting. In the GGPM framework,
the observation likelihood of the GP model is itself parameterized
using the exponential family distribution (EFD). In this paper, we
consider
efficient algorithms for approximate inference on GGPMs using
the general form of the EFD. A particular GP model and its associated
inference algorithms can then be formed by changing the parameters
of the EFD, thus greatly simplifying its creation for task-specific
output domains. We demonstrate the efficacy of this framework by
creating several new GP models for regressing to non-negative reals
and to real intervals. We also consider %
a closed-form Taylor
approximation for efficient inference on GGPMs, and elaborate on its
connections with other model-specific heuristic closed-form approximations.
Finally, we present a comprehensive set of experiments to compare
approximate inference algorithms on a wide variety of GGPMs.
\end{abstract}


\section{Introduction}

In recent years, Gaussian processes (GPs) \citep{GPML}, a
non-parametric Bayesian approach to regression and classification,
have been gaining popularity in machine learning and computer vision.
For example, recent work \citet{Kapoor2010objects} has demonstrated
promising results on object classification using GP classification
and active learning. GPs have several properties that are desirable
for solving complex regression tasks, such as those found in computer vision.
First, due to the Bayesian
formulation, GPs can be learned robustly from small training sets,
which is important in tasks where the amount of training data is
sparse compared to the dimension of the model (e.g., large-scale
object recognition, tracking, 3d human pose modeling).
Second,  GP regression produces a predictive distribution, not just
a single predicted value, thus providing a probabilistic approach to
judging confidence in the predictions, e.g., for active learning.
Third, GPs are based on kernel functions between the input examples,
which allows for both a diverse set of image representations (e.g.,
bag-of-words, local-feature descriptors), and incorporation of prior
knowledge about the computer vision task (e.g., modeling object
structure).
Finally, in the GP framework, the kernel hyperparameters can be
learned by maximizing the marginal likelihood, or evidence, of the
training data.
This is typically more efficient than standard cross-validation
(which requires a grid search), and allows for more expressive
kernels, e.g., compound kernels that model different trends in the
data, or multiple kernel learning, where features are optimally
combined by weighting the kernel function of each feature.

Because of these advantages, GP regression and classification have
been applied to many computer vision problems, such as
object classification \citep{Kapoor2010objects}, %
human action recognition \citep{Han2009iccv}, %
age estimation \citep{Zhang2010mtwgp}, eye-gaze recognition
\citep{Noris2008eyes},
tracking \citep{Raskin2007gpadf},  %
counting people \citep{Chan2008cvpr, Chan2009iccv}, crowd flow
modeling \citep{Ellis2009iccv}, anomaly detection
\citep{Loy2009bmvc}, stereo vision \citep{Williams2006SGP,
Sinz2004dagm},
interpolation of range data \citep{Plagemann07GBP}, %
non-rigid shape recovery \citep{Zhu2009cvpr},
3d human pose recovery \citep{Bo2010TGP, Urtasun2008cvpr,
Fergie2010bmvc, Zhao2008icpr},
and latent-space models of 3d human pose \citep{Urtasun2005iccv, Wang2008GPDM, Chen2009SGPDM}. %
However, despite their successes, many of these methods attempt to
``shoe-horn'' their computer vision task into the standard GP
regression framework.  In particular, while the standard GP
regresses a continuous {\em real-valued} function, it is often used
to predict {\em discrete} non-negative integers (crowd counts
\citealp[]{Chan2008cvpr} or age \citealp[]{Zhang2010mtwgp}),
non-negative real numbers (disparity \citealp{Williams2006SGP,
Sinz2004dagm}
or depth \citealp{Plagemann07GBP}), %
and real numbers on a fixed interval (pose angles
\citealp{Bo2010TGP, Urtasun2008cvpr, Fergie2010bmvc, Zhao2008icpr}
or squashed optical flow \citealp{Loy2009bmvc}).
Hence, heuristics are often required to convert the real-valued GP
prediction to a {\em valid task-specific output}, which is not
optimal in the Bayesian setting. For example in
\citet{Chan2008cvpr}, the real-valued GP prediction must be
truncated and rounded to generate a proper count prediction, and it
is not obvious how the predictive distribution over real-values can
be converted to one over counts.

Developing a new GP model for each of the above regression tasks
requires first finding a suitable distribution  for the output
variable (e.g., Poisson distribution for counting numbers, Gamma
distribution for positive reals, Beta distribution for a real
interval), and then deriving an approximate inference algorithm.
This task can be simplified considerably with recourse to a unifying
framework, which we call a {\em generalized Gaussian process model}
(GGPM) \citep{Chan2011CVPR}.
The GGPM is inspired by the {\em generalized linear model} (GLM)
\citep{book:GLM}, which aims to consolidate parametric regression
methods (e.g., least-squares regression, Poisson regression,
logistic regression) into a unifying framework.  Similarly, the GGPM
unifies many Bayesian non-parametric regression methods using GP
priors (e.g., GP regression, GP classification, and GP counting)
through the {\em exponential family}.

With the GGPM, the observation likelihood of the output is itself parameterized using
the {\em generic form} of the exponential family distribution.
Approximate inference algorithms can then be derived that  depend
only on these EFD parameter functions, elucidating the terms (e.g.,
derivatives, moments) needed for each algorithm.
Different GP models are then created by simply changing the
parameters of the likelihood function, thus  {\em easing the
development of new GP models for task-specific output domains}.
Note that this is analogous to GLMs \citep{book:GLM}, where a common
iteratively reweighted least-squares (IRLS) algorithm was derived to
estimate all associated regression models.

This paper is intended to both survey existing GP regression models,
as well as develop a unifying regression framework for GP models and
its associated  approximate inference algorithms.
Besides further formalizing the GGPM framework, the contributions of
this paper are 4-fold:
1) we derive a %
closed-form approximate inference method for
GGPMs, based on a Taylor approximation, and show that model-specific
closed-form approximations from
\citet{Kapoor2010objects,Chan2009iccv,Heikkinen2008} are special
cases; 2) we analyze existing approximate inference algorithms
(Laplace approximation, expectation propagation, variational
approximations) using the generic form of the exponential family
distribution; 3) using the GGPM framework, we propose several new GP
models for regressing to non-negative and interval real outputs; 4)
we conduct comprehensive experiments comparing the efficacy of the
approximate inference algorithms on both synthetic and real data
sets.
The remainder of the paper is organized as follows.  In Section
\ref{text:previous}, we first review Gaussian process regression and
related work.  In Section  \ref{text:GGPM}, we introduce the GGPM
framework, in Section \ref{text:examples}, we discuss existing novel GP models within the GGPM framework.
In Section \ref{text:appinf}, we derive efficient approximate inference algorithms.  Next, we compare approximate posteriors in Section \ref{1D_example}, and discuss initialization strategies for hyperparameter estimation in Section \ref{multiple_local_minima}.
Finally, in Section \ref{text:experiments}, we present
experiments to compare the approximate inference algorithms, as well
as demonstrate the efficacy of the new proposed models.

\section{Gaussian processes and related work}
\label{text:previous}

In this section we review Gaussian process regression and other related work.

\subsection{Gaussian process regression}

Gaussian process regression (GPR) \citep{GPML} is a Bayesian
approach to predicting a real-valued function $f(\vx)$ of an input
vector $\vx\in\real^d$ (also known as the regressor or explanatory
variable).  The function value is observed through a noisy
observation (or measurement or output) $y \in \real$,
    \begin{align}
    y = f(\vx) + \epsilon,
    \end{align}
where $\epsilon$ is zero-mean Gaussian noise with variance
$\sigma_n^2$, i.e., $\epsilon \sim \Normal{0}{\sigma_n^2}$.  A
zero-mean {\em Gaussian process} (GP) prior is placed on the
function, yielding the GPR model
    \begin{align}
    f \sim \GP(\vzero, k(\vx,\vx')),
    \quad\quad
    y|f(\vx) \sim \Normal{f(\vx)}{\sigma_n^2}.
    \label{eqn:gpr}
    \end{align}
A GP is a random process that represents a distribution over
functions, and is completely specified by its mean and covariance
functions, $m(\vx)$ and $k(\vx,\vx')$.  For simplicity, we assume
the mean function is zero. The covariance (or kernel) function
$k(\vx,\vx')$ determines the class of functions that $f$ can
represent (e.g., linear, polynomial, etc).
Given any set of input vectors $\mX = [\vx_1,\cdots,\vx_n]$, the GP
specifies that the corresponding function values $\vf = [f(\vx_1),
\cdots, f(\vx_n)]^T$ are jointly Gaussian,
    $\vf|\mX \sim \Normal{0}{\mK}$,
where $\mK$ is the covariance (kernel) matrix with entries
$k(\vx_i,\vx_j)$.

The function $f$ is estimated from a training set of input vectors
$\mX = [\vx_1,\cdots,\vx_n]$ and corresponding {\em noisy}
observations $\vy = [y_1,\cdots, y_n]^T$. First, given the inputs
and noisy outputs $\{\mX, \vy\}$, the posterior distribution of the
corresponding function values $\vf$ is obtained with Bayes' rule,
    \begin{align}
    \label{eqn:postf}
    p(\vf|\mX,\vy) = \frac{p(\vy|\vf) p(\vf|\mX)}{\int p(\vy|\vf) p(\vf|\mX) d\vf}
    \end{align}
where $p(\vy|\vf) = \prod_{i=1}^n p(y_i|f_i)$ is the observation
likelihood, and the denominator is the marginal likelihood,
$p(\vy|\mX) = \int p(\vy|\vf) p(\vf|\mX) d\vf$. To predict a
function value $f_* = f(\vx_*)$ from a novel input $\vx_*$, the
posterior distribution in \refeqn{eqn:postf} is marginalized to
obtain the predictive distribution (i.e., an average over all
possible latent function values),
    \begin{align}
    \label{eqn:predfstar}
    p(f_*|\mX,\vx_*,\vy) = \int p(f_*|\vf, \mX,\vx_*) p(\vf|\mX,\vy) d\vf
    \end{align}
\comments{ the joint distribution of $\{\vf, f_*\}$ and the
conditional distribution of $f_*$ are both Gaussian,
    \begin{align}
    \begin{bmatrix}
    \vf \\ f_*\end{bmatrix} | \mX, \vx_* \sim \Normal{\vzero}{\begin{bmatrix}\mK & \vk_* \\ \vk_*^T & k_{**} \end{bmatrix}},
    \quad\quad
    f_* | \vf, \mX, \vx_* \sim \Normal{\vk_*^T \mK^{-1} \vf}{ k_{**} - \vk_*^T \mK^{-1} \vk_*},
    \end{align}
where $\vk_* = [k(\vx_*, \vx_i)]_i$ and $k_{**} = k(\vx_*,\vx_*)$. }
Finally, the distribution of the predicted noisy observation $y_*$
is obtained by marginalizing over $f_*$,
    \begin{align}
    \label{eqn:predystar}
    p(y_*|\mX,\vx_*,\vy) = \int p(y_*|f_*)p(f_*|\mX,\vx_*,\vy) df_*.
    \end{align}
Since the observation likelihood and posterior are both Gaussian,
the predictive distributions in (\ref{eqn:predfstar},
\ref{eqn:predystar}) are both Gaussian, with parameters that can be
computed in closed-form \citep{GPML},
    \begin{align}
    \mu_* &= \vk_*^T (\mK + \sigma^2\mI)^{-1} \vy ,\quad
    \sigma_*^2 =  k_{**} - \vk_*^T(\mK^{-1} + \sigma^2\mI)^{-1} \vk_* , \\
    p(f_*|\mX,\vx_*,\vy) &= \Normalv{f_*}{\mu_*}{\sigma_*^2}, \quad
    p(y_*|\mX,\vx_*,\vy) = \Normalv{y_*}{\mu_*}{\sigma_*^2+\sigma_n^2}.
    \end{align}
where $\vk_* = [k(\vx_*, \vx_1) \cdots k(\vx_*, \vx_n)]^T$ and
$k_{**} = k(\vx_*,\vx_*)$.

The hyperparameters $\valpha$ of the kernel function and the
observation noise $\sigma_n^2$ are typically estimated by maximizing
the marginal likelihood, or evidence, of the training data (also
called Type-II maximum likelihood),
    \begin{align}
    \{\hat{\valpha}, \hat{\sigma}_n^2\} &= \argmax_{\valpha,\sigma_n^2} \log p(\vy|\mX), \\
    \log p(\vy|\mX) %
    &=  -\tfrac{1}{2}\vy^T(\mK+\sigma_n^2\mI)^{-1}\vy - \tfrac{1}{2} \log \detbar{\mK+\sigma_n^2\mI} - \tfrac{n}{2}\log 2\pi.
	\label{eqn:GPmarg}
    \end{align}
The marginal likelihood measures the data fit, averaged over all
probable functions.  Hence, the kernel
hyperparameters are selected so that each probable latent function will model the
data well.

\subsection{GP classification and other GP models}

For Gaussian process classification (GPC)
\citep{Nickisch2008GPC,Kuss05GPC,GPML}, a GP prior is again placed
on the function $f$, which is then ``squashed'' through a sigmoid
function to obtain the probability of the class $y\in\{0,1\}$,
    \begin{align}
    f \sim \GP(0,k(\vx,\vx')),
    \quad\quad
    p(y=1|f(\vx)) = \sigma(f(\vx)),
    \label{eqn:gpc}
    \end{align}
where $\sigma(f)$ is a sigmoid function, e.g. the logistic or
probit functions, which maps a real number to the range $[0,1]$.  However,
since the observation likelihood  is no longer Gaussian, computing
the posterior and predictive distributions in (\ref{eqn:postf},
\ref{eqn:predfstar}, \ref{eqn:predystar}) is no longer analytically
tractable.  This has led to the development of several approximate
inference algorithms for GPC, such as Markov-chain Monte Carlo
(MCMC) \citep{Nickisch2008GPC}, variational bounds
\citep{Gibbs1997VGPC}, Laplace approximation \citep{Williams1998GPC}, and
expectation propagation (EP) \citep{Minka2001,GPML}. As an
alternative to approximate inference, the classification task itself
can be approximated as a GP {\em regression} problem, where the
observations are set to $y\in\{-1,+1\}$ and standard GPR is applied.
This is a computationally efficient alternative called {\em label
regression} \citep{Nickisch2008GPC} (or {\em least-squares
classification} in \citealp{GPML}, and also discussed in
\citealp{Tresp2000GBCM} as a ``fast version'' for two-class
classification), and has shown promising results in object
recognition \citep{Kapoor2010objects}.

GPR has been extended in several ways for other regression tasks
with univariate outputs. Robust GP regression can be obtained by
replacing the Gaussian observation likelihood with the Laplace or
Cauchy likelihood \citep{Opper2009}, or with a student-t likelihood
\citep{Jylanki2011}.
\citet{Paciorek2004} uses a binomial likelihood to model event
occurrence data.
In spatial statistics, the {\em kriging} method was developed for
interpolating spatial data, and essentially uses the same model as
GP regression. \citet{dregress:Diggle1998,Vehtari2007,Vanhatalo2010}
extend this framework for modeling counting observations, by
assuming a Poisson observation likelihood and a GP spatial prior.
Similarly, \citet{Chan2009iccv} develops a method for Bayesian
Poisson regression, by applying a Gaussian prior on the linear
weights of Poisson regression.  Using a log-gamma approximation and
kernelizing yields a closed-form solution that resembles GPR with
specific output-dependent noise.
Finally, \citet{Savitsky2011,Savitsky2010} presents a Bayesian
formulation of the Cox hazard model, by replacing the linear
covariate with a GP prior, and also studies GP counting models using
the Poisson and negative binomial likelihoods, in the context of
Bayesian variable selection.
\begin{table*}
 \scriptsize

\begin{center}
\begin{tabular}{@{}c|c|c|c|c|c|c|c|c@{\hspace{0.02in}}l@{}l@{}}
  & %
  & %
  & \multicolumn{6}{c}{inference methods}
\\
method
  & likelihood
  & $\cal Y$
  & MCMC
  & LA
  & EP
  & KLD
  & VB
  & TA$^\#$ &
\\ \cline{1-10}
regression
  & Gaussian %
  & $\real$ %
  & - %
  & - %
  & - %
  & - %
  & - %
  & \textbf{RW}06 & (exact)
  &
  \rdelim\}{14}{3mm}[{\bf GGPMs}]
\\ \cline{1-10}
\multirow{2}{*}{classification}
  & \multirow{2}{*}{logit/probit} %
  & \multirow{2}{*}{$\{0,1\}$} %
  & \multirow{2}{*}{\textbf{N}97}  %
  & {\textbf{WB}98} %
  & \multirow{2}{*}{\textbf{M}01} %
  & \multirow{2}{*}{\textbf{NR}08} %
  & \multirow{2}{*}{\textbf{GM}97} %
  & \multirow{2}{*}{\textbf{NR}08} & \multirow{2}{*}{(label regression)}\\
  & & & &\textbf{KG}06& & & & &%
\\
\comments{ classification
  & probit?%
  & $\pm1$  %
  & \cite{Neal97}  - %
  & - %
  & - %
  & - %
  & - %
  & - & %
\\
} \cline{1-10}
robust regression
  & Laplace  %
  & $\real$ %
  & - %
  & - %
  & \textbf{K}06 %
  & \textbf{MA}09 %
  & \textbf{RN}10$^\dagger$ %
  & - & %
\\ \cline{1-10}
\multirow{3}{*}{counting}
  & \multirow{3}{*}{Poisson} %
  & \multirow{3}{*}{$\boardZ_+^{\infty}$} %
  & \textbf{D}98 %
  & \multirow{3}{*}{\textbf{V}10} %
  & \multirow{3}{*}{\textbf{V}10} %
  & \multirow{3}{*}{\#} %
  & \multirow{3}{*}{-} %
  & \multirow{3}{*}{\textbf{CV}09} & \multirow{3}{*}{(BPR)}\\
  & & & \textbf{VV}07 & &&&&&\\
  & & & \textbf{S}11& &&&&& %
\\ \cline{1-10}
counting
  & COM-Poisson %
  & $\boardZ_+^{\infty}$ %
  & - %
  & \# %
  & \# %
  & \# %
  & - %
  & \# &
\\ \cline{1-10}
robust counting
  & neg. binomial %
  & $\boardZ_+^{\infty}$ %
  & \textbf{S}11 %
  & \textbf{V}11$^\dagger$ %
  & \textbf{V}11$^\dagger$ %
  & \# %
  & - %
  & \# & %
\\ \cline{1-10}
occurrence
  & binomial %
  & $\boardZ_+^{N}$ %
  & \textbf{PS}04 %
  & \textbf{V}11$^\dagger$ %
  & \textbf{V}11$^\dagger$ %
  & \# %
  & - %
  & \# & %
  \\\cline{1-10}
\comments{ survival time
  & Weibull %
  & $\real_+$ %
  & \cite{GPstuffcode}$^\dagger$ %
  & \cite{GPstuffcode}$^\dagger$ %
  & \cite{GPstuffcode}$^\dagger$ %
  & - %
  & - %
  & - & %
\\
}
\comments{ robust regression
  & Cauchy (student-t w/ 1dof) %
  & $\real$  %
  & - %
  & - %
  & - %
  &   %
  & - %
  & - &%
\\
}\comments{
\\
robust regression
  & sech$^2$ %
  & $\real$ %
  & -  %
  & \cite{GPMLcode}$^\dagger$ %
  & \cite{GPMLcode}$^\dagger$ %
  & - %
  & \cite{GPMLcode}$^\dagger$ %
  & - & %
}
range regression
  & beta %
  & $[0, 1]$ %
  &  %
  & \# %
  & \# %
  & \# %
  & - %
  & \# & %
\\\cline{1-10}
non-negative
  & Gamma  %
  & $\real_{+}$ %
  &   %
  & \# %
  & \# %
  & \# %
  & - %
  & \# & \\ \cline{1-10}
non-negative
  & Inv. Gaussian
  & $\real_{+}$ %
  &   %
  & \# %
  & \# %
  & \# %
  & - %
  & \# &
\\ \cline{1-10}
 \multirow{2}{*}{robust regression}
  & \multirow{2}{*}{student-t}  %
  & \multirow{2}{*}{$\real$} %
  & \textbf{N}97  %
  & \textbf{V}09 %
  & \multirow{2}{*}{\textbf{J}11} %
  & \textbf{K}06 %
  & \multirow{2}{*}{\textbf{RN}10$^\dagger$} %
  & \multirow{2}{*}{-} & \multirow{2}{*}{}\\
  & & &\textbf{K}06 &\textbf{RN}10 & &\textbf{MA}09$^*$ & & &
\\\cline{1-10}
\end{tabular}
\caption{Bayesian regression methods using GP priors.}
\label{tab:priorgp}
\end{center}

\vspace{-0.1in}
{%
\footnotesize Abbreviations -- Inference: MCMC (Markov-chain Monte
Carlo),
LA (Laplace approximation),  EP (expectation propagation); KLD (KL
divergence minimization),
VB (variational bounds), TA (Taylor approximation). %
Citations:  \textbf{CV}09 \citep{Chan2009iccv}, \textbf{D}98
\citep{dregress:Diggle1998}, \textbf{GM}97 \citep{Gibbs1997VGPC},
\textbf{J}11 \citep{Jylanki2011}, \textbf{K}06 \citep{KussThesis},
\textbf{KG}06 \citep{Kim2006emep}, \textbf{M}01 \citep{Minka2001},
\textbf{MA}09 \citep{Opper2009}, \textbf{N}97 \citep{Neal1997},
\textbf{NR}08 \citep{Nickisch2008GPC}, \textbf{PS}04
\citep{Paciorek2004}, \textbf{RN}10 \citep{GPMLcode}, \textbf{RW}06
\citep{GPML}, \textbf{S}11 \citep{Savitsky2011}, \textbf{V}09
\citep{Vanhatalo2009}, \textbf{V}10 \citep{Vanhatalo2010},
\textbf{V}11 \citep{GPstuffcode}, \textbf{VV}07 \citep{Vehtari2007},
\textbf{WB}98 \citep{Williams1998GPC}
Other:
$\boardZ_+^N = \{0,1,2,\ldots,N\}$,
BPR (Bayesian Poisson regression);
$^\dagger$part of a software  toolbox.  %
$^*$considered the Cauchy distribution (Student-t with 1 d.o.f.).
$\#$ introduced in this paper by our model.
}
\end{table*}

Table \ref{tab:priorgp} summarizes the previous work on Bayesian
regression models using GP priors, along with the approximate
inference algorithms proposed for them. The goal of this paper is to
generalize many of these models into a unified framework.
Finally, GP models have also be extended to model {\em multivariate
observations}, i.e, vector outputs.
\citet{Chu2005GPOR} proposes GP ordinal regression (i.e., ranking)
using a multi-probit likelihood, while multiclass classification is
obtained using a probit  \citep{Girolami05VBM,Kim2006emep}  or
softmax \citep{Williams1998GPC} sigmoid function.
\citet{Teh05SPLFM} linearly mixes independent GP priors to obtain a
{\em semiparametric latent factor model}. In this paper, we only
consider a univariate outputs with a single GP prior; extending the
GGPM to multivariate outputs is a topic of future work.

\subsection{Related work on GGPMs}

Previous works on GGPMs include \citet{dregress:Diggle1998}, which
focuses on geostatistics (extending kriging) using Poisson- and
binomial-GGPMs,
\citet{Paciorek2004}, which uses binomial-GGPM in the context of
testing non-stationary covariance functions, and
\citet{Savitsky2011}, which is mainly interested in variable
selection by adding priors to the kernel hyperparameters of a GGPM.
All these works
\citep{dregress:Diggle1998,Paciorek2004,Savitsky2011} perform
inference using MCMC, by plugging in  different likelihood functions
without exploiting the exponential family form.  MCMC tends to be
slow, and convergence problems were observed in
\citet{dregress:Diggle1998}.
In contrast,  this paper focuses on {\em efficient} algorithms for
approximate inference, and derives their general forms by exploiting
the exponential family form.
By doing so, we can create a ``plug-and-play'' aspect to GP models,
which we exploit later to create several novel GP models with very
little extra work.
\citet{Seeger2004GPML,Tresp2000GBCM,GPFRbook} also briefly mention
the connection between the assumed output likelihoods of GPR/GPC and
the exponential family, but do not study approximate inference or
new models in depth.
The GGPM  can be interpreted as a Bayesian approach to {\em
generalized linear models} (GLMs) \citep{book:GLM}, where a Gaussian
prior is placed on the linear weights (or equivalently a GP prior
with linear kernel is placed on the systemic component, i.e., latent
function).
Previous works on Bayesian GLMs consider different priors.
A typical approach is to form a Bayesian hierarchical GLM by
applying a conjugate prior on the parameters \citep[e.g.,
][]{Albert1988BHGLM,Das2007,Bedrick1996}.
Recent work  focuses on inducing sparsity in the latent function,
e.g.,
\citet{Seeger07ECML} and \citet{Nickisch2009VBGLM} assume %
a factorial heavy-tailed prior distribution, but are not
kernelizable due to the factorial assumption.
\citet{Hannah2010DPGLM} proposes a mixture of GLMs, based on a
Dirichlet process, to allow different regression parameters in
different areas of the input space, and performs inference using
MCMC.

When used with a non-linear kernel, the GGPM is a Bayesian
kernelized GLM for non-linear regression.
\citet{Zhang2010BGKM} also proposes a Bayesian kernelized GLM %
using a hierarchical model with a sparse prior (a mixture of point
mass and Silverman's g-prior) and evaluates their models on
classification problems.
Finally, \citet{Cawley2007GKM} proposes a non-Bayesian version of a
GLM, called a {\em generalised kernel machines} (GKM),  which is
based on kernelizing the iterated-reweighted least-squares algorithm
(IRWLS).  The GGPM is a Bayesian formulation of the GKM.

With respect to our previous work \citep{Chan2011CVPR}, this paper
presents more algorithms for approximate inference (e.g.,
variational methods), and in more depth.  We also propose novel GP
models for regression to non-negative reals and real intervals, and
show more connections with heuristic methods using the Taylor
approximation.  Furthermore, comprehensive experiments are presented
to compare the performance of the inference algorithms on a wide
variety of likelihood functions.

\section{Generalized Gaussian process  models}

\label{text:GGPM}

In this section, we introduce the generalized Gaussian process
model, a non-parametric Bayesian regression model that encompasses
many existing GP models.

\subsection{Exponential family distributions}
We first note that different GP models are obtained by changing the
form of the observation likelihood $p(y|f)$.  The standard GPR
assumes a Gaussian observation likelihood, while GPC  essentially uses a Bernoulli distribution, and \citet{Chan2009iccv} uses a
Poisson likelihood for counting.
These likelihood functions are all instances of the single-parameter
{\em exponential family distribution} \citep{book:Duda}, with
likelihood given by
    \begin{align}
    p(y|\theta,\phi) = \exp \left\{ \frac{1}{a(\phi)}\left[T(y) \theta - b(\theta)\right] + c(\phi,y) \right\},
    \label{eqn:expo}
    \end{align}
where $y \in {\cal Y}$ is the observation from a set of possible
values ${\cal Y}$ (e.g., real numbers, counting numbers, binary
class labels), $\theta$ is the natural parameter of the exponential
family distribution, and $\phi$ is the dispersion parameter. $T(y)$
is the sufficient statistic (e.g. the logit function for beta
distribution), $a(\phi)$ and $c(\phi,y)$ are known functions, and
$b(\theta)= \log \int \exp ( \frac{1}{a(\phi)}T(y) \theta +
c(\phi,y)) dy$ is the log-partition function,
which normalizes the distribution.  The mean and variance of the
sufficient statistic $T(y)$ are functions of $b(\theta)$ and
$a(\phi)$,
    \begin{align}
    \mu = \EV[T(y)|\theta] = \sd{b}(\theta), \quad
    \var[T(y)|\theta] = \sdd{b}(\theta) a(\phi),
    \label{eqn:expomeanvar}
    \end{align}
where $\sd{b}(\theta)$ and $\sdd{b}(\theta)$ are the first and
second derivatives of $b$ w.r.t.~$\theta$.
The exponential family  generalizes a wide variety of distributions for different output domains, %
and hence a unifying framework can be created by analyzing
a GP model where the likelihood takes the {\em generic form} of
\refeqn{eqn:expo}.

\subsection{Generalized Gaussian process models}

\comments{ We now consider a Bayesian framework for GLMs, which adds
a prior distribution on the weight parameters of the GLM.  In
general, we can consider two approaches. The first is a {\em
weight-space} approach, where we assume a prior distribution on the
linear weights, i.e. $\vbeta \sim \Normal{0}{\mSigma_p}$.  Given a
novel input $\vx_*$, a Gaussian approximation to the posterior
$p(\theta_* | \mX, \vx_*, \vy)$ will lead to kernelization and
tractable approximate inference algorithms. On the other hand, the
second approach is to employ a {\em function-space} view, where we
assume a functional prior on the systemic function $\eta(\vx)$, i.e.
$\eta(\vx) \sim \GP(0,k(\vx,\vx'))$, where $\GP$ is a zero-mean
Gaussian process prior with covariance function $k(\vx,\vx')$.  In
other words, we directly start with a kernelized prior (through the
GP covariance function $k(\vx,\vx')$), and tractable approximate
inference is again possible due to a Gaussian approximation on the
posterior of $\theta_*|\mX,\vx_*,\vy$. It turns out that both
approaches will yield the same generalized GPR model, albeit the
latter requires significantly less derivation and is thus the focus
of this paper. }

We now consider a framework for a generic Bayesian model that
regresses from inputs $\vx\in \real^d$ to outputs $y \in {\cal Y}$,
which encompasses many popular GP models.
Following the formulation of GLMs \citep{book:GLM},
the model is composed of
three components:
\begin{enumerate}
\item a latent function, $\eta(\vx) \sim \GP(0,k(\vx,\vx'))$, which is a function of the inputs, modeled with a  GP prior.
\item a random component, $p(y|\theta,\phi)$, that models the output as an exponential family distribution with parameter
$\theta$ and dispersion $\phi$.
\item a link function, $\eta = g(\mu)$, that relates the {\em mean} of the sufficient statistic with the latent function.
\end{enumerate}
Formally, the GGPM is specified by
    \begin{align}
    \eta(\vx) \sim \GP(0, k(\vx,\vx')),\quad
    y \sim p(y|\theta,\phi),
    \quad
    g(\EV[T(y)|\theta]) = \eta(\vx).
    \label{eqn:GGPR1}
    \end{align}
The mean of the sufficient statistic is related to the latent function $\eta(\vx)$ through the inverse-link function, i.e. $\mu = g^{-1}(\eta(\vx))$.   %
The advantage of the link function is that it allows
{\em direct specification of prior knowledge} about the functional relationship
 between the  output mean and the latent
function $\eta(\vx)$.
On the other hand, the effect of the GP kernel function is to
adaptively warp (or completely override) the link function to fit
the data.  While many trends can be represented by the GP kernel
function (e.g., polynomial functions), it is important to note that
some functions (e.g., logarithms)  %
{\em cannot be naturally
represented by a kernel function},  due to its positive-definite
constraint.  Hence, directly specifying the link function is
necessary for these cases.

Substituting \refeqn{eqn:expomeanvar} for the mean, we obtain the
parameter $\theta$ as a function of the latent function,
    \begin{align}
    \eta(\vx) %
    = g(\EV[T(y)|\theta]) = g( \sd{b}(\theta) )
    \ \ \Rightarrow\  \
    g^{-1}(\eta(\vx)) = \sd{b}(\theta)
    \ \ \Rightarrow\  \
    \theta(\eta(\vx)) = \sd{b}^{-1} (g^{-1} (\eta(\vx))),
    \label{eqn:thetaeta}
    \end{align}
where $\sd{b}^{-1}$ is the inverse of the first derivative of
$b(\cdot)$.
Using \refeqn{eqn:thetaeta}, another form of GGPM that directly relates the latent function with the parameter is
    \begin{align}
    \eta(\vx) \sim \GP(0, k(\vx,\vx')),
    \quad
    y \sim p(y|\theta(\eta(\vx)),\phi),
    \quad
    \theta(\eta(\vx))  = \sd{b}^{-1} (g^{-1}( \eta(\vx))).
    \label{eqn:GGPR2}
    \end{align}
The GGPM unifies many Bayesian  regression methods using GP priors
(e.g., all of Table \ref{tab:priorgp} except the student-t distribution), with each
model arising from a specific instantiation of the parameter
functions, $\calE = \{a(\phi), b(\theta), c(\phi,y),
\theta(\eta), T(y)\}$.
Given a set of training examples and a novel input, the predictive
distribution is obtained by marginalizing over the posterior of the
latent function $\eta(\vx)$, similar to standard GPR/GPC \citep{GPML}.
The dispersion $\phi$ is treated as a hyperparameter, which can be
estimated along with the kernel hyperparameters by maximizing the
marginal likelihood.

\subsection{Canonical link function}

One common link function is to select $g(\cdot)$ such that $\theta(\eta(\vx)) =
\eta(\vx)$.   This is called the {\em canonical link function}, and is obtained with $g(\cdot) = \sd{b}^{-1}(\cdot)$ and $g^{-1}(\cdot) = \sd{b}(\cdot)$.
For the canonical link function, the GGPM simplifies to
	\begin{align}
	    \eta(\vx) \sim \GP(0, k(\vx,\vx')),
	    \quad
	    y \sim p(y|\theta=\eta(\vx),\phi),
	    \label{eqn:GGPR-canonical}
	\end{align}
and the output mean is related to the latent function by
	$\mu = \sd{b}(\eta(\vx))$.
\subsection{Inference and prediction}
\label{text:inference}

Inference on GGPMs follows closely to that of standard GPR/GPC \citep{GPML}.
Given a set of training examples, input vectors $\mX =
[\vx_1,\cdots,\vx_n]$ and corresponding observations $\vy =
[y_1,\cdots, y_n]^T$, the goal is to generate a {\em predictive
distribution} of the output $y_*$ corresponding to a novel input
$\vx_*$ . The predictive distribution is obtained using two steps.  The first step is to calculate a posterior distribution
of the latent function values $\veta$, which best explains the
training examples $\{\mX,\vy\}$.  This corresponds to the training
or parameter estimation phase of a regression model.  In the second
step, the distribution of the latent function value $\eta(\vx_*)$
for the novel input $\vx_*$ is calculated, followed by the
predictive distribution for $y_*$.

Formally, given the training inputs $\mX = [\vx_1,\cdots,\vx_n]$,
the latent  values $\eta_i = \eta(\vx_i)$ are jointly
Gaussian, according to the GP prior,
    $p(\veta | \mX) = \Normalv{\veta}{0}{\mK}$,
where $\veta = [\eta_1,\cdots, \eta_n]^T$ and $[\mK]_{i,j} =
k(\vx_i, \vx_j)$ is the kernel matrix.
The posterior distribution of $\veta$ is obtained by further
conditioning on the training outputs $\vy$, and applying Bayes'
rule,
    \begin{align}
    p(\veta|\mX,\vy) = \frac{p(\vy|\vtheta(\veta)) p(\veta|\mX)}{p(\vy|\mX)},
    \label{eqn:GGPRpost}
    \end{align}
where $p(\vy|\vtheta(\veta))$ is the observation
likelihood\footnote{To reduce clutter, we will not write the
dependency on the dispersion parameter $\phi$ or the kernel
hyperparameters, unless we are explicitly optimizing them.}, and
$p(\vy|\mX)$ is the marginal likelihood, or evidence,
    \begin{align}
    p(\vy|\mX) = \int p(\vy|\vtheta(\veta)) p(\veta|\mX) d\veta.
    \label{eqn:GGPRmarg}
    \end{align}
The posterior $p(\veta|\mX,\vy)$ in \refeqn{eqn:GGPRpost} is the
distribution of the latent  values for all inputs $\mX$ that
can best describe the provided input/output pairs.

Next, the predictive distribution of $y_*$ for a novel
input $\vx_*$ is obtained by first predicting the latent function value $\eta_*
= \eta(\vx_*)$. Conditioned on all the inputs $\{\mX,\vx_*\}$, the
joint distribution of the latent function values $\{\veta,\veta_*\}$
is also Gaussian, via the GP prior, and the
the conditional distribution is obtained using the conditional
Gaussian theorem,
    \begin{align}
    p(\eta_* | \veta, \mX, \vx_*) = \Normalv{\eta_*}{\vk_*^T \mK^{-1} \veta}{ k_{**} - \vk_*^T \mK^{-1} \vk_*},
    \end{align}
where $\vk_* = [k(\vx_1,\vx_*), \cdots k(\vx_n,\vx_*)]^T$  and
$k_{**} = k(\vx_*, \vx_*)$. The predictive distribution of $\eta_*$
is then obtained by marginalizing over the posterior distribution in
\refeqn{eqn:GGPRpost} (i.e., averaging over all possible latent
functions),
    \begin{align}
    p(\eta_*|\mX, \vx_*, \vy) = \int p(\eta_*|\veta,\mX,\vx_*) p(\veta|\mX, \vy) d\veta .
    \label{eqn:GGPRpoststar}
    \end{align}
 Finally, the predictive distribution of $y_*$  is obtained by marginalizing over $\eta_*$,
    \begin{align}
    p(y_*|\mX, \vx_*, \vy) = \int p(y_*|\theta(\eta_*)) p(\eta_*|\mX, \vx_*, \vy) d\eta_* .
    \label{eqn:GGPRpredstar}
    \end{align}
Note that for most non-Gaussian likelihoods, the posterior and
predictive distributions in (\ref{eqn:GGPRpost}, \ref{eqn:GGPRmarg},
\ref{eqn:GGPRpoststar}, \ref{eqn:GGPRpredstar}) cannot be computed
analytically in closed-form.  We will discuss efficient approximate inference
algorithms in Section \ref{text:appinf}.

\subsection{Learning the hyperparameters}
As in GPR \citep{GPML}, the kernel hyperparameters $\valpha$ and the
dispersion $\phi$, can be estimated from the data %
 by maximizing the marginal likelihood in
\refeqn{eqn:GGPRmarg},
    \begin{align}
    \{\valpha^*,\phi^*\} = \argmax_{\valpha, \phi} \int p(\vy|\veta,\phi) p(\veta|\mX,\valpha) d\veta,
    \end{align}
where we now note the dependence on the hyperparameters. The
marginal likelihood measures the data fit, averaged over all
probable latent functions.  Hence, the criteria selects the kernel
hyperparameters such that each probable latent function will model
the data well.

\section{Example GGPMs}
\label{text:examples}

In this section, using the GGPM framework, we propose several {\em
novel} models for GP regression on different output domains.
We obtain Bayesian regression for non-negative real number outputs
by adopting the Gamma and inverse-Gaussian distributions, and propose a Beta-GGPM that regresses to real
numbers on the $[0,1]$ interval.
We also consider existing GP models within the GGPM framework (e.g.,
Poisson- and binomial-GGPMs from Table \ref{tab:priorgp}).
We also discuss the role of the link function and its selection criteria. %

\subsection{Summary}

Table~\ref{tab:ExpoDistLink} (top) presents some examples of EFDs (distributions and parameters), while
Table~\ref{tab:ExpoDistLink} (bottom) shows their expectations and link functions of their corresponding GGPM.
The table encompasses both existing and novel GP models.
By  changing the parameters of the EFD to form a specific
observation likelihood (i.e., selecting the functions $\{a(\phi),
b(\theta), c(\phi,y), \theta(\eta), T(y)\}$,
we can easily obtain a wide range of GP models with
different types of outputs, e.g., Gamma and Inverse Gaussian
for non-negative reals, Beta for $[0,1]$ interval reals, etc.

\begin{table}[h]\scriptsize
\begin{center}
\begin{tabular}{@{}l|l@{\hspace{0.05in}}c@{}c@{}||c@{}c@{}c@{\hspace{0.05in}}c@{}}
\hline Model & $y$ & $p(y)$ & $\{\theta, \phi\}$ & $T(y)$ &
$a(\phi)$ & $b(\theta)$ & $c(\phi, y)$
\\
\hline
\hline \hbox{Gaussian} & $\mathbb{R}$ &
$\frac{1}{\sqrt{2\pi\sigma^2}}e^{\frac{-1}{2\sigma^2}(y-\mu)^2}$ &
$\{\mu, \sigma^2\}$ & $y$ & $\phi$ & $\frac{1}{2}\theta^2$ &
$-\frac{\log(2\pi\phi)}{2}-\frac{y^2}{2\phi}$\\

\hline \hbox{Gamma\shape} & $\mathbb{R}_+$ &
$\frac{(\nu/\mu)^{\nu}}{\Gamma(\nu)}y^{\nu-1}e^{-y\nu/\mu}$ &
$\{\frac{-1}{\mu}\, \frac{1}{\nu}\}$ & $y$ & $\phi$&
$-\log(-{\theta})$ &
$\log\frac{y^{\phi^{-1}-1}{\phi^{-1}}^{\phi^{-1}}}{\Gamma(\phi^{-1})}$\\

\hline \hbox{Gamma\scale} & $\mathbb{R}_+$ &
$\frac{s^{-\nu}}{\Gamma(\nu)}y^{\nu-1}e^{-y/s}$ & $\{\nu, s\}$ &
$\log y$ & $\phi$& $\theta \log \phi + \phi\log \Gamma(\frac{\theta}{\phi})$ &
$-y/\phi-\log
y$\\

\hline \hbox{Inv. Gauss.} & $\mathbb{R}_+$ &
$\left[\frac{\lambda}{2\pi
y^3}\right]^{\frac{1}{2}}e^{\frac{-\lambda(y-\mu)^2}{2\mu^2 y}}$ &
$\{\frac{-1}{2\mu^2}\, \frac{1}{\lambda}\}$ & $y$ &
$\phi$ & $-\sqrt{-2\theta}$ &
$\log\left(\frac{1}{2\pi
y^3\phi}\right)^{\frac{1}{2}}-\frac{1}{2y\phi}$\\

\hline \hbox{Neg. Bino.} & $\mathbb{Z}_+$ &
$\frac{\Gamma(y+\alpha^{-1})p^{\alpha^{-1}}(1-p)^y}{\Gamma(y+1)\Gamma(\alpha^{-1})}$
& $\{\alpha\log(1-p), \alpha\}$ & $y$ & $\phi$&
$-\log{(1-e^{\phi^{-1}\theta})}$ & $\log
\frac{\Gamma(y+\phi^{-1})}{\Gamma(y+1)\Gamma(\phi^{-1})}$\\

\hline \hbox{Poisson} & $\mathbb{Z}_+$ & $\frac{1}{y!}\lambda^{y}
e^{-\lambda}$ & $\{\log \lambda, 1\}$ & $y$ & $\phi$ & $e^{\theta}$ &
$-\log(y!)$\\

\hline \hbox{COM-Po.} & $\mathbb{Z}_+$ &
$\frac{1}{S(\mu,\nu)}\left[\frac{\mu^y}{y!}\right]^{\nu}$ & $\{\log
\mu, \nu\}$ & $y$ & $\tfrac{1}{\phi}$ %
& $\phi^{-1}\log
S(e^{\theta},
\phi)$ & $-\phi\log(y!)$\\

\hline \hbox{Binomial} &  $\tfrac{1}{N}\mathbb{Z}_+^N$  %
&
$\binom{N}{Ny}\pi^{Ny}(1-\pi)^{N-Ny}$ & $\{\log \frac{\pi}{1-\pi},
\frac{1}{N}\}$ & $y$ & $\phi$& $\log(1+e^{\theta})$ & $\log
\binom{\phi^{-1}}{\phi^{-1}y}$\\

\hline \hbox{Beta} & $[0, 1]$ & $\frac{\Gamma(\nu)y^{\mu
\nu-1}(1-y)^{(1-\mu)\nu-1}}{\Gamma(\mu \nu)\Gamma((1-\mu) \nu)}$ & $\{\mu,
\frac{1}{\nu}\}$ & $\log \frac{y}{1-y}$ & $\phi$&
$\phi\log\Gamma(\frac{\theta} {\phi})\Gamma(\frac{1-\theta}{\phi})$ &
$\log\frac{\Gamma(\frac{1}{\phi})(1-y)^{\frac{1}{\phi}-1}}{y}$
\\

\hline
\end{tabular}
\end{center}
\vspace{-0.25in}
\begin{center}
\begin{tabular}{@{}l@{\hspace{0.1in}}|cc||c|c@{}}
\hline Model & $\sd{b}(\theta) = \mathbb{E}[T(y)]$ & $\hbox{var}[T(y)]$ &
$g(\mathbb{E}[T(y)])$& $\theta(\eta)$
\\
\hline
\hline \hbox{Gaussian} &$\theta$ & $\phi$ & $\mu$& $\eta$\\

\hline \hbox{Gamma\shape} &$-\theta^{-1}$ & $\theta^{-2}\phi$
&
$\log \mu$ & $-e^{-\eta}$\\

\hline \hbox{Gamma\scale} & $\log \phi + \psi_0(\theta/\phi)$ &
$\psi_1(\theta/\phi)$&$\log \psi_0^{-1}(\mu-\log\phi)+\log\phi$&
$e^{\eta}$\\

\hline \hbox{Inv. Gauss.} & ${(-2\theta)}^{-\frac{1}{2}}$ &
$\phi{(-2\theta)}^{-\frac{3}{2}}$ & $2\log\mu + \log2$&
$-e^{-\eta}$\\
\hline \hbox{Neg. Bino.}&
$\frac{\phi^{-1}e^{\phi^{-1}\theta}}{1-e^{\phi^{-1}\theta}}$ &
$\frac{\phi^{-1}e^{\phi^{-1}\theta}}{(1-e^{\phi^{-1}\theta})^2}$
& $-\log \left[\left(\frac{\mu\phi}{1+\mu\phi}\right)^{-\phi}-1\right]$& $-\log (1+e^{-\eta})$\\

\hline \hbox{Poisson} & $e^{\theta}$ & $e^{\theta}$ & $\log \mu$& $\eta$\\

\hline \hbox{Linear Poisson} & $e^{\theta}$ & $e^{\theta}$ & $\log(e^{\mu}-1)$& $\log(\log(1+e^{\eta}))$\\
\hline \hbox{COM-Poisson} &
$e^{\theta}+\frac{1}{2\phi}-\frac{1}{2}$ & $-$ & $\log(\mu-\frac{1}{2\phi}+\frac{1}{2})$& $\eta$\\

\hline \hbox{Linear COM-Poisson} &
$e^{\theta}+\frac{1}{2\phi}-\frac{1}{2}$ & $-$ & $\log(e^{\mu-\frac{1}{2\phi}+\frac{1}{2}}-1)$ & $\log(\log(1+e^{\eta}))$\\

\hline \hbox{Binomial} & $\frac{e^{\theta}}{1+e^{\theta}}$ &
$\frac{\phi e^{\theta}}{(1+e^{\theta})^2}$ & $\log \frac{\mu}{1-\mu}$ & $\eta$\\

\hline \hbox{Beta} &
$\psi_0(\frac{\theta}{\phi})-\psi_0(\frac{1-\theta}{\phi})$&
$\psi_1(\frac{\theta}{\phi})+\psi_1(\frac{1-\theta}{\phi})$ & $\mu$& $\frac{e^{\eta}}{1+e^{\eta}}$\\

\hline
\end{tabular}
\end{center}
\vspace{-0.25in}
\caption{(top) Exponential family distributions;
(bottom) expectations and link functions for GGPM}
\label{tab:ExpoDistLink}
\end{table}

\comments{
\subsection{Gaussian likelihood}

The Gaussian distribution for $y\in\real$, $p(y|\mu, \sigma^2) = (2
\pi \sigma^2)^{-\frac{1}{2}} e^{\frac{-1}{2\sigma^2}(y-\mu)^2}$,
\comments{
    \begin{align}
    p(y|\mu, \sigma^2) = \frac{1}{\sqrt{2 \pi \sigma^2}} e^{\frac{-1}{2\sigma^2}(y-\mu)^2},
    \end{align}
} can be rewritten in exponential family form by setting
$\theta=\mu$ and $\phi = \sigma^2$, with
    \begin{align}
    a(\phi) = \phi,\quad
    b(\theta) = \tfrac{1}{2}\theta^2,  \quad
    h(y,\phi) = \tfrac{1}{\sqrt{2\pi\phi}}\exp(-\tfrac{y_i^2}{2\phi}).
    \nonumber
    \end{align}
Using the canonical link function, where $\theta(\eta) = \eta$, we
have
    \begin{align}
    \EV[y] = g^{-1}(\eta)&= b'(\eta) = \eta, \quad
    \Rightarrow g(\mu) = \mu.
    \end{align}
\comments{ The derivative functions are
    \begin{align}
    u(\eta,y) &= \frac{1}{\sigma^2} (y-\eta), \\
    w(\eta,y) &= \sigma^2, \\
    \pdd{}{\phi} \log h(y_i,\phi) &= \frac{y_i^2}{2\phi^2} - \frac{1}{2\phi}.
    \end{align}
} The Gaussian-GGPM is the standard GPR model \citep{GPML}. }

\subsection{Binomial distribution}

The binomial distribution models the probability of a certain number
of events occurring in $N$ independent trials, the event probability
in an individual trial is $\pi$, and $y \in
\{\frac{0}{N},\frac{1}{N},\cdots,\frac{N}{N}\}$ is the fraction of
events. Assuming the canonical link function, then
    \begin{align}
    \mu = \EV[y|\theta] %
    = g^{-1}(\eta)
    = \tfrac{e^{\eta}}{1+e^{\eta}},
    \end{align}
and hence the mean is related to the latent space through the
logistic function.  For $N=1$, the binomial-GGPM (or Bernoulli-GGPM) is equivalent to
the GPC model using the logistic function. For $N>1$, the model can
naturally accommodate uncertainty in the labels by using fractional
$y_i$, e.g., for $N=2$ there are three levels $y \in
\{0,\frac{1}{2},1\}$.
Furthermore in the $N=1$ case, by changing the link function to the probit function,
we obtain GPC using the probit likelihood,
    \begin{align}
    \nonumber
    g(\mu) = \Phi^{-1}(\mu)
    \quad    \Rightarrow\quad
    \mu = g^{-1}(\eta) = \Phi(\eta)
    \end{align}
where $\Phi(\eta)$ is the cumulative distribution of a Gaussian.
Substituting into the GGPM, we have
    \begin{align}
    \theta(\eta) = \log \tfrac{\Phi(\eta)}{1-\Phi(\eta)},
    \quad
    b(\theta(\eta)) = -\log(1-\Phi(\eta)).
    \nonumber
    \end{align}
A common interpretation of the probit GPC is that the class probability arises from a noisy Heaviside step function \citep{Seeger2004GPML}.  Here, the GGPM framework provides further insight that the probit and logistic GPC models correspond to Bernoulli-GGPMs using different link functions.
\subsection{Counting GP models}

In this section, we consider counting regression models (Poisson, negative binomial, Conway-Maxwell Poisson), and show how the link function can be changed to better model the mean trend.

\subsubsection{Poisson distribution}

The Poisson distribution is a model for counting data, where the outputs $y
\in \boardZ_+ = \{0,1, \cdots\}$  are counts, and $\lambda$ is the
arrival-rate (mean) parameter. For the canonical link function,
    \begin{align}
    \EV[y|\theta] = %
    g^{-1}(\eta) %
    = e^{\eta} = \lambda,
    \ \ g(\mu) = \log \mu.
    \end{align}
Hence, the mean of the Poisson is the exponential of the latent
value.  The Poisson-GGPM is a Bayesian regression model for
predicting counts $y$ from an input vector $\vx$, and has been
previously studied
\citep{Chan2009iccv,dregress:Diggle1998,Vehtari2007}.

\setlength{\myw}{0.23\linewidth}
\begin{figure}[tbh]
\centering
\begin{tabular}{@{\hspace{-0.1in}}c@{}c@{}c@{}c}
\hspace{0.2in}{\footnotesize (a) Poisson} & \hspace{0.2in}{\footnotesize
(b) linear Poisson} & \hspace{0.2in}{\footnotesize (c) COM-Poisson} &
\hspace{0.2in}{\footnotesize (d) linear COM-Poisson}
\\
\includegraphics[width=1.1\myw]{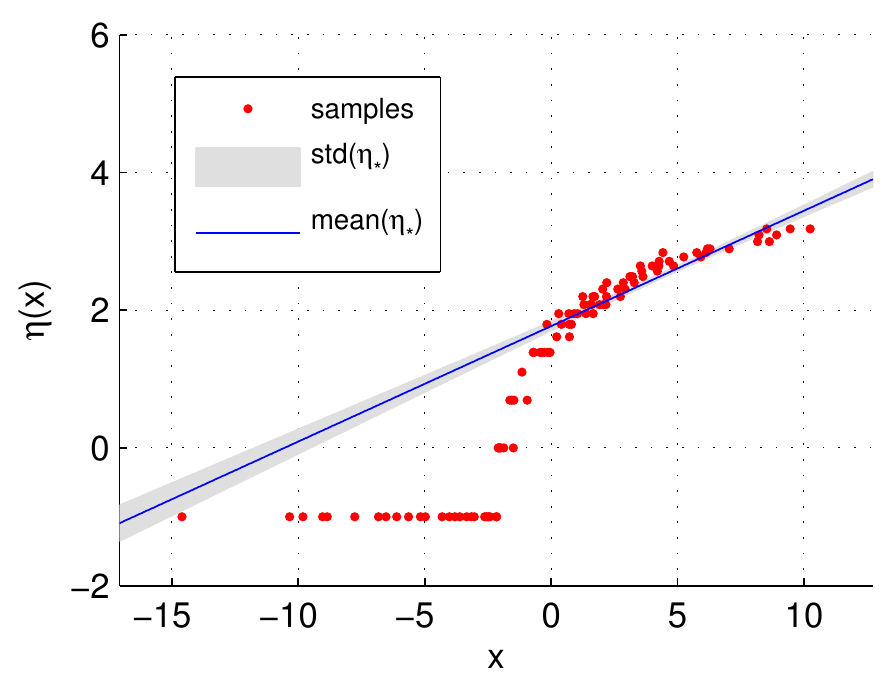}
&
\includegraphics[width=1.1\myw]{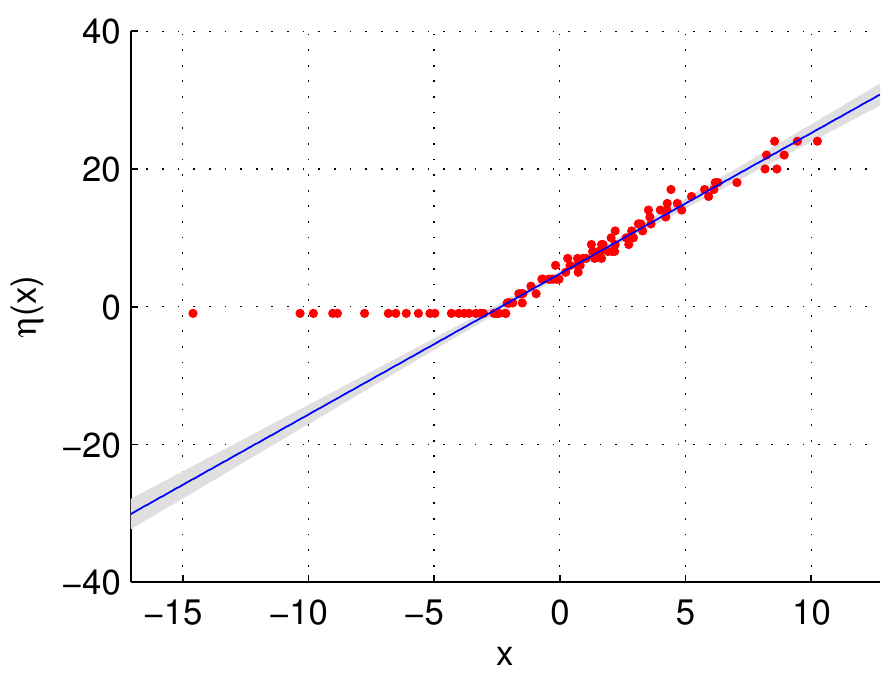}
&
\includegraphics[width=1.1\myw]{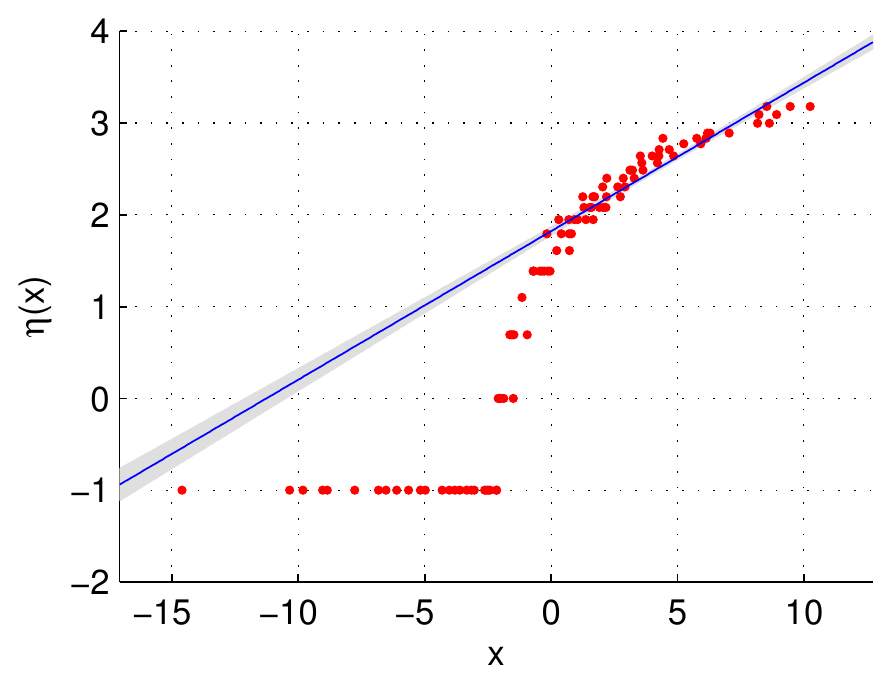}
&
\includegraphics[width=1.1\myw]{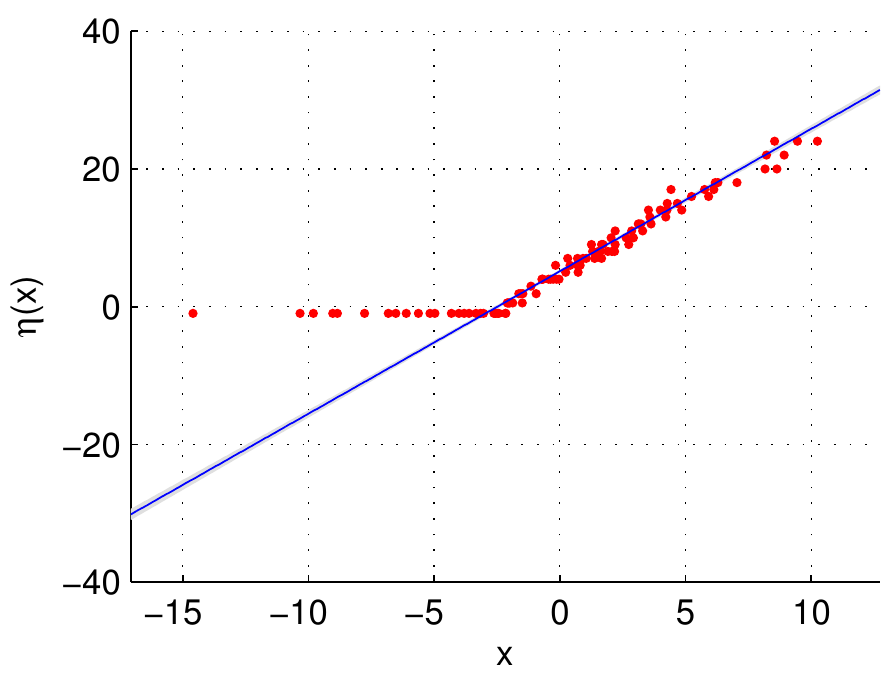}
\vspace{-0.08in}
\\
\includegraphics[width=1.1\myw]{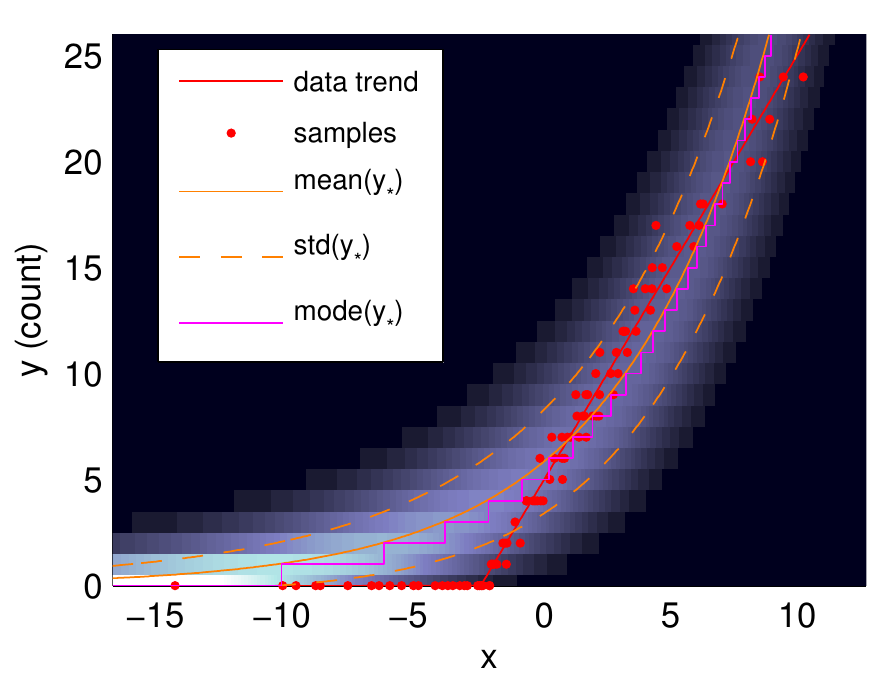}
&
\includegraphics[width=1.1\myw]{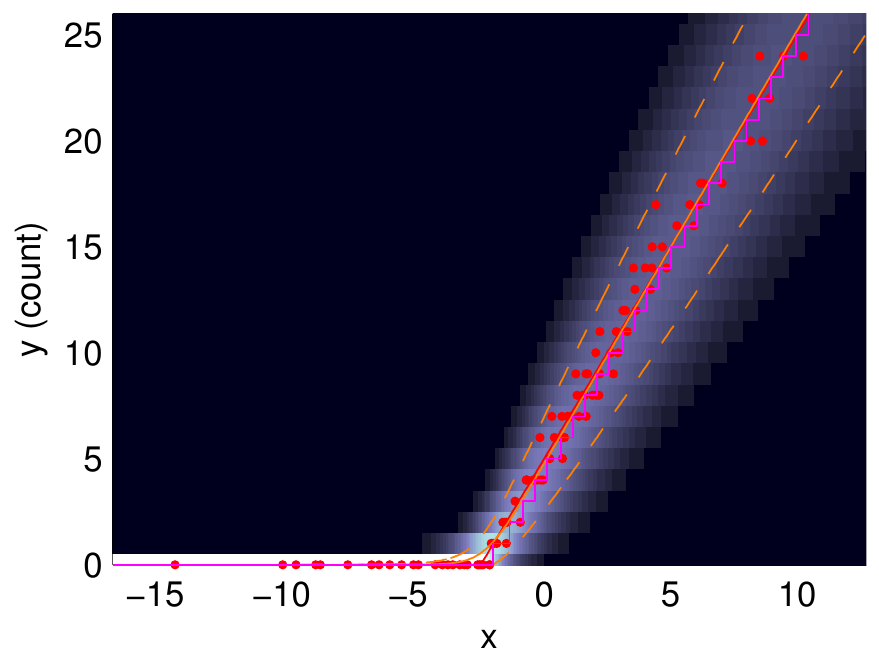}
&
\includegraphics[width=1.1\myw]{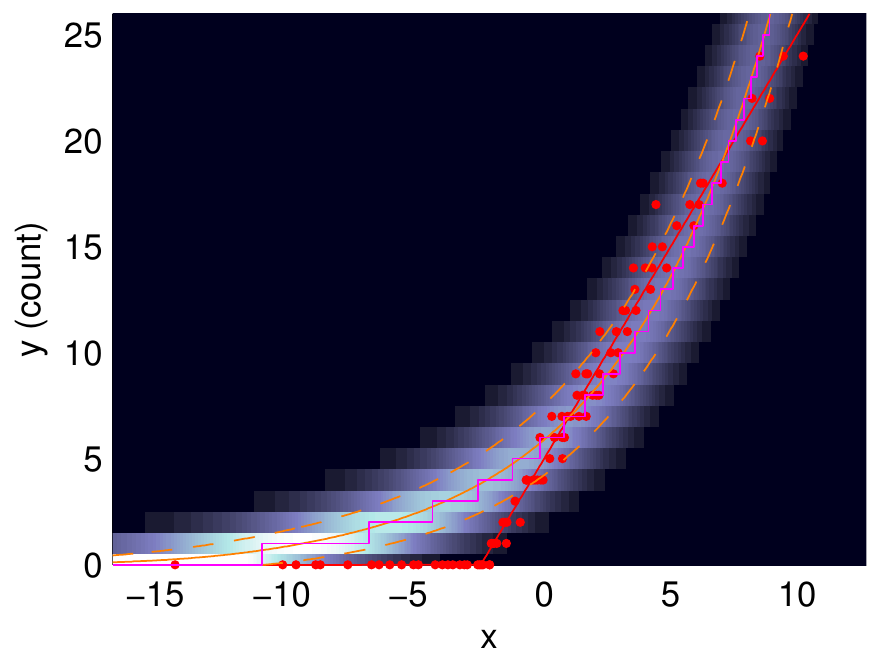}
&
\includegraphics[width=1.1\myw]{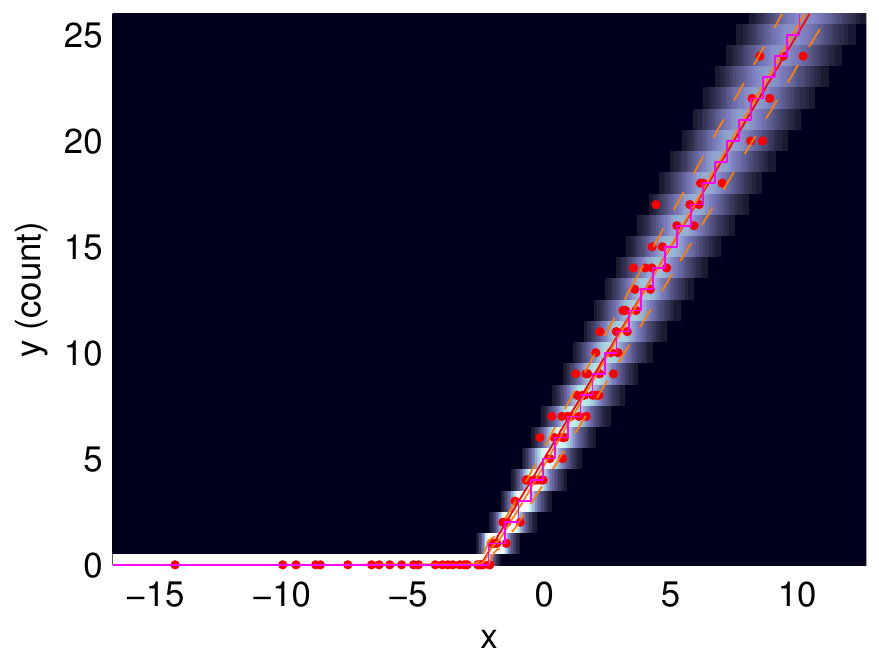}
\end{tabular}
\caption{ Examples of count regression using GGPM with linear kernel and different
likelihood functions: a) Poisson; b) linearized Poisson; c)
COM-Poisson; d) linearized COM-Poisson.  The data follows a linear
trend and is underdispersed. The top row shows the learned latent function, and the bottom row shows the
predictive distributions.  The background color indicates the count
probability (white most probable, black least probable)}
\label{fig:Poissons}
\end{figure}

\begin{figure}[tbh]
\centering
\begin{tabular}{@{}c@{}c}
\includegraphics[width=0.3\linewidth]{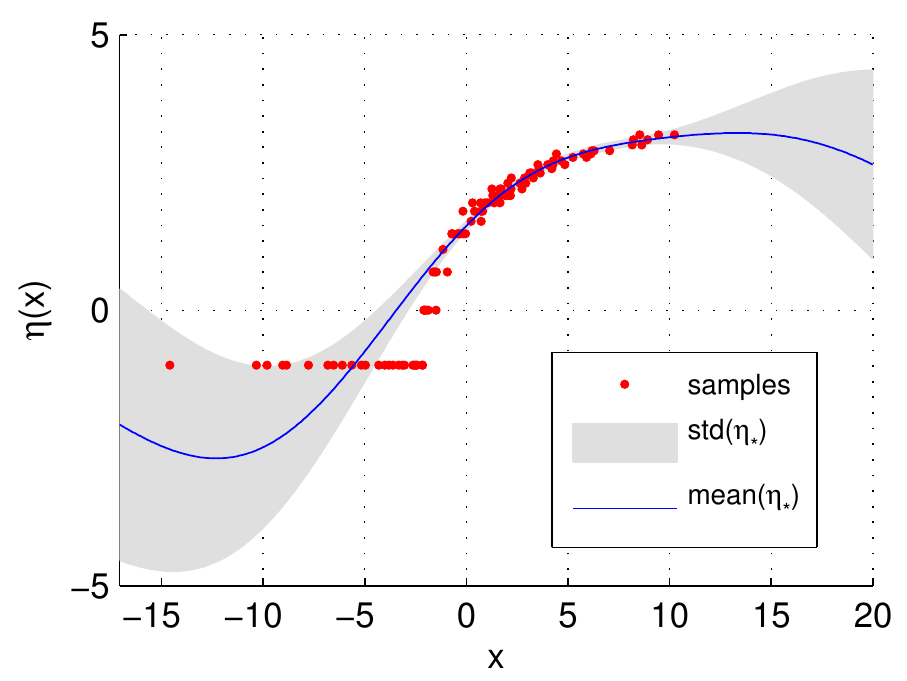}
&
\includegraphics[width=0.3\linewidth]{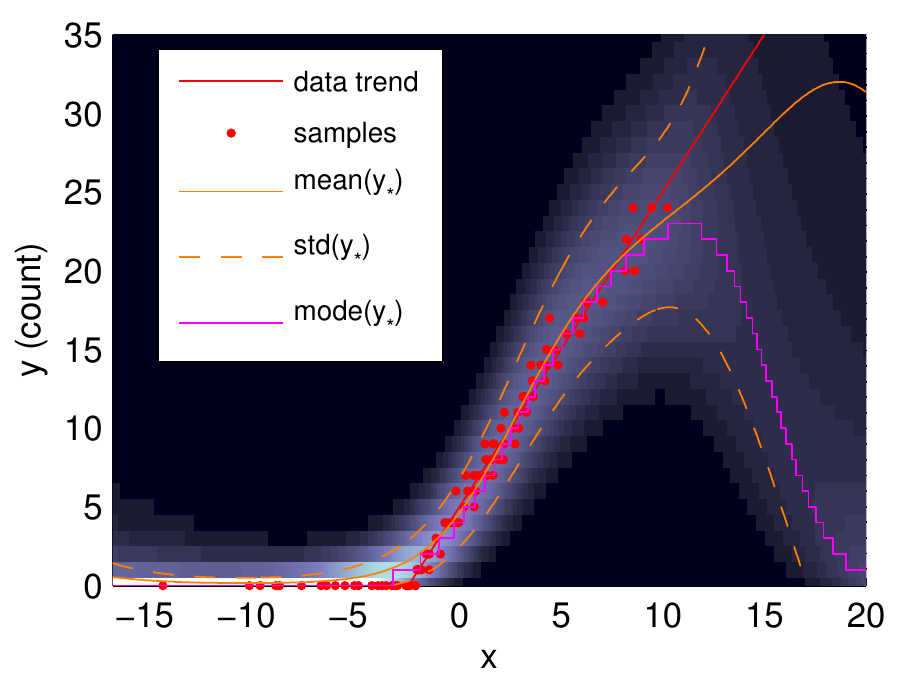}
\end{tabular}
\caption{Example of regressing a linear trend using a Poisson-GGPM with RBF
kernel.} \label{fig:PoissonRBF}
\end{figure}

\subsubsection{Linearized mean}
\label{text:linearizedmean}

The canonical link function assumes that the mean is the exponential
of the latent function.  This may cause problems when
the actual mean trend is different,
as illustrated in Figure \ref{fig:Poissons}a, where the count
actually follows a linear trend. One way to address this problem is
to use a non-linear kernel function to counteract the
exponential link function.
In this case, the ideal kernel function should be a logarithm function.  However, there is no such positive definite kernel, and hence changing the kernel cannot recover a linear trend exactly.
Furthermore, using the RBF kernel  has poor extrapolation capabilities due to
the limited extent of the RBF function, as illustrated in Figure \ref{fig:PoissonRBF}.

Alternatively, the mean can be {\em directly linearized} by changing
the link function of the Poisson-GGPM to represent a linear trend.   For this
purpose, we use the logistic error function,
    \begin{align}
\EV[y|\theta] =  g^{-1}(\eta) = \log(1+e^\eta) \ \Rightarrow \
    g(\mu) = \log (e^{\mu} - 1), \mu > 0.
    \nonumber
    \end{align}
For large values of $\eta$, the link function is linear, while for negative values of $\eta$, the function approaches zero.  %
The parameter function and new partition function are
    \begin{align}
    \theta(\eta) = \log ( \log( 1+e^\eta)),
    \quad
    b(\theta(\eta)) = \log(1+e^\eta).
    \label{eqn:poislin:theta}
    \end{align}
Figures \ref{fig:Poissons}a and \ref{fig:Poissons}b illustrate the
difference between the standard and linearized Poisson GGPMs.  The
standard Poisson-GGPM cannot correctly model the linear trend,
resulting in a poor data fit at the extremes, while the linearized
Poisson follows the linear trend.

One limitation with the Poisson distribution is that it models an
equidispersed random variable, i.e. the variance is equal to the
mean. However, in some cases, the actual random variable is {\em
overdispersed} (with variance greater than the mean) or {\em
underdispersed} (with variance less than the mean).
We next consider two other counting distributions that can handle overdispersion and underdispersion.

\subsubsection{Negative Binomial}

The negative binomial distribution is a model for counting
data, which is overdispersed (variance larger than the mean).
The mean is given by $\mu = \frac{1-p}{\alpha p}$,
where $p\in(0, 1)$ and $\alpha \geq0$ is the scale parameter that adjusts the variance.
The variance can be
written as a function of mean, and always exceeds the mean,
   $\var(y) = \EV[y] + \alpha \EV[y]^2$.
When $\alpha \rightarrow0$, the negative binomial reduces to the Poisson distribution.

Since $\theta=\alpha\log(1-p) < 0$ for valid parameters, the choice of $\theta(\eta)$ must satisfy a non-positive constraint.
Hence, we use a flipped log-loss function,
    \begin{align}
    \theta(\eta) = -\log(1+e^{-\eta}),
    \end{align}
with the corresponding link function,
    \begin{align}
    \EV[y|\theta] %
    = g^{-1}(\eta) %
    = \frac{\alpha^{-1}(1+e^{-\eta})^{-\alpha^{-1}}}{1-(1+e^{-\eta})^{-\alpha^{-1}}}\ \Rightarrow \
    g(\mu) = -\log \left[\left(\frac{\mu\alpha}{1+\mu\alpha}\right)^{-\alpha}-1\right].
    \end{align}
When $\alpha = 1$, the link function reduces to $\log \mu$, and the negative binomial reduces to a  Geometric distribution.

\subsubsection{Conway-Maxwell-Poisson distribution}

Another alternative distribution for count data, which represents
different dispersion levels, is the Conway-Maxwell-Poisson
(COM-Poisson) distribution
\citep{Conway1962,Shmueli2005,Guikema2008},
    \begin{align}
    p(y|\mu, \nu) = \frac{1}{S(\mu, \nu)} \left[\frac{\mu^y}{y!}\right]^{\nu},
    \ \  S(\mu, \nu) = \sum_{n=0}^{\infty} \left[\frac{\mu^n}{n!}\right]^{\nu},
    \end{align}
where $y\in \boardZ_+$, $\mu$ is (roughly) the mean parameter, and
$\nu$ is the dispersion parameter.
The COM-Poisson is a smooth interpolation between three
distributions: geometric ($\nu=0$), Poisson ($\nu=1$), and Bernoulli
($\nu\rightarrow\infty$). The distribution is overdispersed for
$\nu<1$, and underdispersed for $\nu>1$.
The partition function $S(\mu, \nu)$ has no closed-form expression, but can be estimated numerically up to any precision \citep{Shmueli2005}.
Note that $b_{\phi}(\theta)$ is now also a function of $\phi$, which only affects optimization of the dispersion $\phi$
(details in Appendix \ref{app:taylor_bphi} and \citet{Chan2013tr}).
For the canonical link function, %
    \begin{align}
    \EV[y] \approx %
    e^{\eta} + \tfrac{1}{2\nu} - \tfrac{1}{2}  = g^{-1}(\eta)
    \ \ \Rightarrow \  \
    g(\mu) = \log(\mu - \tfrac{1}{2\nu}+\tfrac{1}{2}).
    \end{align}
Alternatively the parameter function in \refeqn{eqn:poislin:theta}
can be used to model a linear trend in the mean.

The COM-Poisson GGPM includes a dispersion hyperparameter that
decouples the variance of the Poisson from the mean, thus allowing
more control on the observation noise of the output.
Figures \ref{fig:Poissons}c and \ref{fig:Poissons}d show examples of
using the COM-Poisson-GGPM on underdispersed counting data with a
linear trend.  Note that the variance of the prediction is much
lower for the COM-Poisson models than for the Poisson models
(Figures \ref{fig:Poissons}a and \ref{fig:Poissons}b), thus
illustrating that the COM-Poisson GGPM can effectively estimate the
dispersion of the data.
A COM-Poisson GLM (with canonical link) was proposed in
\citet{Guikema2008}, and thus the COM-Poisson GGPM is a non-linear
Bayesian extension using a GP prior on the latent function.

\subsection{GP regression to non-negative real numbers}

In this section, we consider GGPMs with non-negative real number outputs.  Two are based on different parameterizations of the Gamma distribution, and the other is based on the inverse Gaussian.
The main difference between the likelihood functions is the amount of observation noise.  In particular, the  variance of the Gamma distribution is $\phi \mu^2$, whereas the inverse Gaussian is more dispersed, with variance $\phi \mu^3$.

\subsubsection{Gamma distribution (mean parameter, shape hyperparameter)}
The Gamma distribution is a model for non-negative real number observations.
The distribution is parameterized by the mean $\mu>0$, and the
\emph{shape} parameter $\nu>0$.
The exponential distribution is a special case of the Gamma distribution
when $\nu =1$.
For the GGPM, we use $\mu$ as the distribution parameter and
$\nu$ as the hyperparameter.
Since $\theta$ must implicitly be negative, we set the function $\theta(\eta) =
-e^{-\eta}$, and the link function is
    \begin{align}
   \EV[y|\theta] =  g^{-1}(\eta) = -1/\theta(\eta) = e^{\eta}
    \ \Rightarrow \
    g(\mu) = \log \mu.
    \end{align}
Hence the link function models $\eta$ as the log-mean of the Gamma.
We denote this likelihood as {\em Gamma-shape} (Gamma\shape), since it uses the shape as the hyperparameter.

\subsubsection{Gamma distribution (shape parameter, scale hyperparameter)}

Alternatively, the Gamma distribution can be parameterized by the shape parameter $\nu$, and a \emph{scale hyperparameter} $s = \frac{\mu}{\nu}$.
Fixing the scale hyperparameter to $s = 2$, the Gamma distribution reduces to
the Chi-square distribution.
Since $\theta$ must be positive, one candidate of $\theta(\eta)$ is
$e^{\eta}$ and the link function is
    \begin{align}
    \EV[T(y)|\theta] = g^{-1}(\eta) = \log \phi + \psi_0(e^{\eta}/\phi)
    \ \Rightarrow \
    g(\mu) = \log \psi_0^{-1}(\mu -\log \phi) +\log \phi,
    \end{align}
where $\psi_k(x) = \pdd{^{k+1}}{x^{k+1}}\log \Gamma(x)$ is the polygamma function, and $\psi_k^{-1}(z)$ is its inverse.
Noting that $\psi_0 \approx \log(x)$, we can approximate the
link function as $g(\mu)\approx\mu=\EV[T(y)] = \EV[\log(y)]$.
We denote this likelihood as {\em Gamma-scale} (Gamma\scale), since it uses the scale as the hyperparameter.

\subsubsection{Inverse Gaussian distribution}
The inverse Gaussian distribution is another model for non-negative real
number outputs, where $\mu$ is the mean parameter and $\lambda$ is the inverse
dispersion.
Since $\theta$ is implicitly negative, we set
$\theta(\eta)=-e^{-\eta}$, and the link function is
    \begin{align}
    \EV[y|\theta] = g^{-1}(\eta) = 1/\sqrt{2e^{-\eta}}
    \ \  \Rightarrow \ \
    g(\mu) = 2\log\mu+\log2.
    \end{align}
The link function models $\eta$ as the log-mean of the inverse Gaussian.

\subsection{Beta likelihood}
The beta distribution is a model for output over a real interval, $y \in [0, 1]$,
where $\mu$ is the distribution mean and $\nu$ is the shape parameter.
To enforce the restriction $0<\theta<1$, we apply the logistic function
   $\theta(\eta) = \frac{e^{\eta}}{1+e^{\eta}}$,
   resulting in the link function,
    \begin{align}
    \EV[T(y)|\theta] = \psi_0(\tfrac{\theta}{\phi}) - \psi_0(\tfrac{1-\theta}{\phi}) \approx \log \tfrac{\theta}{1-\theta} =
    \eta
    \quad\Rightarrow\quad
    g(\mu) \approx \mu,
    \end{align}
where we use the approximation to the digamma function, $\psi_0(x) \approx \log x$.
The Beta-GGPM maps observations from the interval $[0, 1]$ to real values
in the latent space using the logit function.
A similar idea has been explored in the study of estimating
reflectance spectra from RGB values \citep{Heikkinen2008}, where the
reflectance spectra output is regressed using a {\em
logit-transformed} GP. Specifically, the output values in $[0,1]$
are first transformed to real values using the logit
function\footnote{Equivalently, the inverse hyperbolic tangent
function (arctanh) is applied to the data scaled to $[-1,1]$.}, and
a standard GP model is estimated on the transformed output.
We show the relationship between this logit-transformed GP and the Beta-GGPM in
Section \ref{text:specialcases}.

\subsection{Choosing the link function}

\begin{figure}[tbh]
\centering
\begin{tabular}{@{}c@{}c}
\includegraphics[scale=0.42]{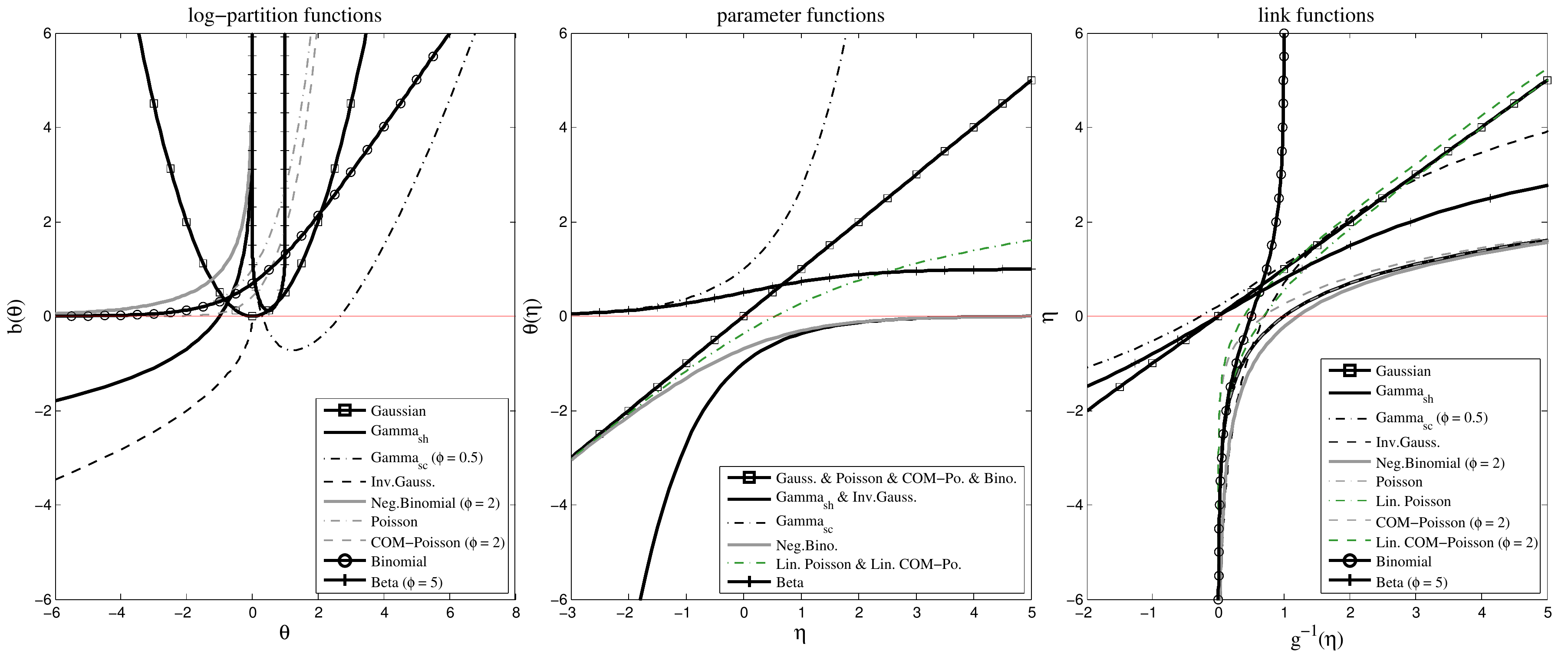}
\end{tabular} \caption{Log-partition, parameter, and link functions.}
\label{fig:functions}
\end{figure}

In the previous examples, several strategies have been used in selecting the link and parameter functions, $g(\mu)$ and $\theta(\eta)$, to obtain a mapping from the latent space to the parameters.
Using the canonical link function simplifies the calculations, but may make undesired assumptions about the mapping between latent space and the distribution mean (as in the exponential mapping for Poisson).  Modifying the link function allows the desired mean trend (e.g., the linearized Poisson).
When selecting the mapping via the parameter function, the main 2 hurdles are: 1) to select a function that satisfies the implicit constraints on the parameter $\theta$ imposed by the log-partition function $b(\theta)$;
2) to select a function that is defined for all values of input $\eta\in\real$ to accommodate the GP prior.
Figure \ref{fig:functions} plots the log-partition functions and link functions for each model.
For example, there is an implicit negative constraint on $\theta$ for the Gamma-shape likelihood, due its the log-partition function.

Finally, it is worth noting the difference between a {\em warped GP}
\citep{Snelson04WGP} and GGPMs. The warped GP learns a function to
warp the output of  GPR. The warping function plays a similar role
as the link function in the GGPM, with one notable difference. With
\citet{Snelson04WGP}, the warping is applied directly to the output
variable $y$, $z=f(y)$, and a GPR is applied on the resulting
$z$.  As a result, the predictive distribution of $z$ is Gaussian,
whereas that of $y=f^{-1}(z)$ is arbitrary (and perhaps
multimodal). On the other hand, with GGPM, the link function applies
warping between the latent value $\eta$ and the mean parameter
$\theta$, and hence the form of the output distribution is
preserved.
In addition, the warping function is learned in
\citet{Snelson04WGP}, whereas in this paper we assume the link
function is fixed for a given GGPM. Certainly, in principle, the
link function could be learned.  However, we note that  the link
function and the kernel function both  control the mapping between
latent space and parameter space.  Hence, if a link function were to
be learned, it would have to be over functions that are orthogonal
to those functions learnable by the kernel function in order to avoid duplicate parameterization.

\section{Approximate inference for GGPMs}
\label{text:appinf}

In this section, we derive approximate inference algorithms for
GGPMs based on the {\em general form} of the exponential family
distribution in \refeqn{eqn:expo}. %
One method of approximate inference is to use MCMC to draw samples
from the posterior $p(\veta|\mX,\vy)$, but this can be
computationally intensive \citep{Nickisch2008GPC}.
Instead, we consider methods that approximate the posterior with a Gaussian.

We %
consider a closed-form approximation to inference based on a Taylor approximation, 
and
we show that the Taylor approximation  can justify common
heuristics or pre-processing steps as principled inference; we
prove that {\em label regression} \citep{Nickisch2008GPC} is a
Taylor approximation to GP classification (Bernoulli-GGPM),
the closed-form Bayesian Poisson regression \citep{Chan2009iccv} is a Taylor approximation of a Poisson-GGPM,
GP regression on {\em log-transformed outputs} is actually a Taylor
approximation to the Gamma-GGPM, and GP regression on {\em logit-transformed outputs} \citep{Heikkinen2008} is a Taylor approximation to the Beta-GGPM.

Finally, we consider several other approximate inference algorithms that have
been previously proposed for various GP models, and generalize them
for GGPMs.

\subsection{Gaussian approximation to the posterior}

As noted in \citet{Nickisch2008GPC}, most  inference approximations on GPC work by finding a Gaussian approximation to the true posterior.  Similarly, for GGPMs approximate inference
also finds  suitable Gaussian approximation $q(\veta|\mX,\vy)$ to the true posterior, i.e.,
    \begin{align}
    p(\veta|\mX,\vy) \approx q(\veta|\mX,\vy) = \Normalvv{\veta}{\hat{\vm}}{\hat{\mV}}
    \end{align}
where the parameters $\{\hat{\vm}, \hat{\mV}\}$ are determined by
the type of approximation.
Substituting the approximation $q(\veta|\mX,\vy)$ into
\refeqn{eqn:GGPRpoststar}, the approximate posterior for $\eta_*$ is
    \begin{align}
    p(\eta_*|\mX,\vx_*,\vy) &\approx
    q(\eta_*|\mX,\vy_*,\vy) = \Normalv{\eta_*}{\hat{\mu}_{\eta}}{\hat{\sigma}_{\eta}^2},
    \end{align}
where the mean and variance are
    \begin{align}
    \label{eqn:approxpoststar:meanvar}
    \hat{\mu}_{\eta} &= \vk_*^T \mK^{-1} \hat{\vm},
    \quad
    \hat{\sigma}_{\eta}^2
    = k_{**} - \vk_*^T(\mK^{-1} - \mK^{-1}\hat{\mV} \mK^{-1} )\vk_*.
    \end{align}
In many inference approximations,  $\{\hat{\vm},\hat{\mV}\}$  take
the form
    \begin{align}
    \label{eqn:approxpost:common}
    \hat{\mV} =( \mK^{-1} + \mW^{-1} )^{-1},
    \quad
    \hat{\vm} = \hat{\mV}\mW^{-1}\vt,
    \end{align}
where $\mW$ is a  diagonal matrix, and $\vt$ is a target vector.  In
these cases, \refeqn{eqn:approxpoststar:meanvar} can be rewritten
    \begin{align}
    \hat{\mu}_{\eta}
    = \vk_*^T \left(\mK + \mW \right)^{-1} \vt
    ,\
    \hat{\sigma}_{\eta}^2
    = k_{**} - \vk_*^T (\mK + \mW)^{-1} \vk_* .
    \end{align}
Note that these are equivalent to the standard equations for GPR,
but with an ``effective'' observation noise $\mW$ and target $\vt$ determined
by the particular approximate inference algorithm.
\subsection{Taylor approximation}
\label{text:Taylor}

In this section, we present a novel closed-form approximation to
inference, which is based on applying a Taylor approximation of the likelihood term
\citep{Chan2011CVPR}.
We first define the following  derivative functions of the
observation log-likelihood,
    \begin{align}
    u(\eta, y) &= \pdd{}{\eta} \log p(y|\theta(\eta))
    =\frac{1}{a(\phi)} \sd{\theta}(\eta)\left[T(y) - \sd{b}(\theta(\eta))\right],
    \\
    w(\eta,y)
    &= - \left[\pdd{^2}{\eta^2} \log p(y|\theta(\eta))\right]^{-1}
    =
    a(\phi) \left\{ \sdd{b}(\theta(\eta)) \sd{\theta}(\eta)^2
    -   \left[T(y) - \sd{b}(\theta(\eta))\right]\sdd{\theta}(\eta) \right\}^{-1}
    \end{align}
For the canonical link function, these derivatives simplify to
    \begin{align}
    u(\eta,y) = \frac{1}{a(\phi)}[T(y) - \sd{b}(\eta)],
    \quad
    w(\eta,y) =
    \frac{a(\phi)}{\sdd{b}(\eta)}.
    \end{align}

\subsubsection{Joint likelihood approximation}

We first consider approximating the joint likelihood
of the data and latent values,
    \begin{align}
    \log p(\vy,\veta|\mX) = \log p(\vy|\vtheta(\veta)) + \log p(\veta|\mX) .
    \label{eqn:cfa:joint}
    \end{align}
The data likelihood term $p(y_i|\theta(\eta_i))$ is the main hurdle for tractable integration over $\veta$.
Hence, we approximate the data log-likelihood term   a 2nd-order Taylor expansion at the expansion point $\teta_i$,
    \begin{align}
    \log p(y_i|\theta(\eta_i))& %
    \approx \log p(y_i|\theta(\teta_i)) + \tu_i (\eta_i-\teta_i) - \frac{1}{2} \tw_i^{-1} (\eta_i-\teta_i)^2
    \label{eqn:cfa:taylor}
    \end{align}
where $\tu_i = u(\teta_i,y_i)$ and $\tw_i = w(\teta_i,y_i)$ are the derivatives evaluated at $\teta_i$.
\comments{
    \begin{align}
    \tv_i = \log p(y_i|\theta(\teta_i)), \
    = v(\teta_i, y_i)
    = \frac{1}{a(\phi)} \left[y_i\theta(\teta_i)  - b(\theta(\teta_i))\right] + \log h(y_i,\phi),
    \tu_i = u(\teta_i,y_i), \quad
    \tw_i = w(\teta_i,y_i).
    \end{align}
with $u(\eta,y)$ and $w(\eta,y)$ are defined in
\refeqn{eqn:likederiv1} and \refeqn{eqn:likederiv2}. } Defining
$\tvu=[\tu_1,\cdots,\tu_n]^T$ and $\tmW =
\diag(\tw_1,\ldots,\tw_n)$, the joint likelihood in
\refeqn{eqn:cfa:joint} can then be approximated as (see Appendix \ref{app:Taylor:joint} for derivation)
    \begin{eqnarray}
    \begin{split}
    \log q(\vy,\veta|\mX)
    &=
    \log p(\vy|\theta(\tveta))
    - \frac{1}{2} \log \detbar{\mK} - \frac{n}{2}\log 2\pi
    \\
    &\quad
    -\frac{1}{2}\norm{\veta - \mA^{-1}\tmW^{-1} \tvt}^2_{\mA^{-1}}
    -\frac{1}{2}\norm{\tvt}_{\mW+\mK}^2
    + \frac{1}{2}\tvu^T\tmW\tvu
    \end{split}
    \label{eqn:cfa:approxjoint}
    \end{eqnarray}
where $\mA = \tmW^{-1}+\mK^{-1}$, $\tvt = \tveta+\tmW\tvu$ is the
target vector, and the individual targets are
$\tilde{t}_i = \teta_i + \tw_i \tu_i$.
The approximate joint log-likelihood in \refeqn{eqn:cfa:approxjoint} will be used to find the approximate posterior and marginal likelihood.

\subsubsection{Approximate posterior}

Removing terms in \refeqn{eqn:cfa:approxjoint} that do not depend on
$\veta$, the posterior of $\veta$ is approximately Gaussian,
    \begin{align}
    \log q(\veta|\mX, \vy) \propto
    \log q(\vy,\veta|\mX)
    \propto
    -\frac{1}{2}\norm{\veta - \mA^{-1}\tmW^{-1}\vt}^2_{\mA^{-1}}
    \Rightarrow
    q(\veta|\mX, \vy)=\Normalvv{\veta}{\hat{\vm}}{\hat{\mV}},
    \end{align}
where,
    $\hat{\mV} =( \tmW^{-1} + \mK^{-1})^{-1}$, and
    $\hat{\vm} = \hat{\mV}\tmW^{-1}\tvt .$
These are of the form in \refeqn{eqn:approxpost:common}, and hence,
the approximate posterior of $\eta_*$ has parameters
    \begin{align}
    \hat{\mu}_{\eta}
    = \vk_*^T (\mK + \tmW )^{-1} \tvt, \ \ \
    \hat{\sigma}_{\eta}^2
    = k_{**} - \vk_*^T (\mK + \tmW)^{-1} \vk_* .
    \nonumber
    \end{align}
The Taylor approximation is a closed-form (non-iterative)
approximation, that can be interpreted as performing GPR on a set of
targets $\tvt$ with  target-specific, non-i.i.d.~observation noise
$\tmW$. The targets $\tvt$ are a function of the the expansion point
$\tveta$, which can be selected as a non-linear transformation of the
observations $\vy$.
$\tmW$ can be interpreted as a Gaussian approximation of the
observation noise in the transformation space of $\tvt$, where the
noise is dependent on the expansion points.
For a standard GPR, this is iid noise, i.e., $\mW
= \sigma_n^2 \mI$.  In other cases, the noise is dependent on the
expansion points, and the particular properties of the observation likelihood.
Instances of the closed-form Taylor approximation for different GP models are further explored in Section \ref{text:examples}.
One advantage with the Taylor approximation is that it is an {\em
efficient non-iterative} method with the same complexity as standard GPR.
In Section \ref{multiple_local_minima}, we exploit
its efficiency to speed up other approximate
inference methods (e.g. EP) during hyperparameter estimation.

\subsubsection{Choice of expansion point}

The targets $\tvt$ are a function of the the expansion point
$\tveta$, which can be chosen as a non-linear transformation of the
observations $\vy$.  Assuming that the output $y_i$ occurs close to
the mean of the distribution, a reasonable choice of the expansion point is
$\teta_i = g(T(y_i))$, which we denote as the \emph{canonical expansion point}. The derivatives and targets
are then simplified to (see Appendix \ref{app:Taylor:expansion})
    \begin{align}
    \tu_i &= u(g(T(y_i)),y_i) = 0 \quad
    \ \ \Rightarrow\ \
    \tilde{t}_i = \teta_i = g(T(y_i)),
    \label{eqn:specialexpansion}
    \\
    \tw_i &=
    w(g(y_i),y_i) = \frac{a(\phi)}{\sdd{b}(\theta(g(T(y_i)))) \sd{\theta}(g(T(y_i)))^2}.
    \label{eqn:specialexpansion1}
    \end{align}
Hence, the Taylor approximation becomes GPR on the transformation of
the input $g(T(y_i))$, with an appropriate non-i.i.d. noise term
given by $\tw_i$.

This formulation gives some further insight on common preprocessing
transformations, such as $\log(y_i)$, used with GPR.   Using the
Taylor approximation, we can show that some forms of preprocessing
are actually making specific assumptions on the output noise.  In
addition, several heuristic methods (e.g., label regression) can be
shown to be instances of approximate inference using the Taylor
approximation.  This is further explored for specific cases in
Section \ref{text:specialcases}.

Finally, it is worth noting the relationship between the GGPM Taylor
approximation and warped GPs \citep{Snelson04WGP}.  Warped GPs also
apply a warping from $y_i$ to $t_i$, and apply GPR on $t_i$.  The
main difference is that with warped GPs, the noise in the
transformed space of $t_i$ is modeled as i.i.d. Gaussian, resulting
in an arbitrary predictive distribution of $y_i$.  On the other hand,
the Taylor approximation models the noise according to the warping
function (i.e., as in $\tw_i$), and hence the predictive
distribution of $y_i$ is preserved.

\subsubsection{Approximate Marginal}

\comments{ The approximate marginal is obtained by substituting the
approximate joint in \refeqn{eqn:cfa:approxjoint}
    \begin{align}
    \log p(\vy|\mX) %
    \approx \log \int \exp (\log q(\vy,\veta|\mX))d\veta,
    \end{align}
resulting in the approximate marginal, }

The approximate marginal likelihood is obtained by integrating out
$\veta$ in \refeqn{eqn:cfa:approxjoint}, yielding  (see Appendix \ref{app:Taylor:marginal} for derivation)
    \begin{align}
    \log q(\vy|\mX)
    &=
    -\frac{1}{2} \tvt^T(\tmW + \mK)^{-1}\tvt
    - \frac{1}{2} \log \detbar{\tmW + \mK} + r(\phi)
    \label{eqn:cfa:marginal}
    \end{align}
where
    $r(\phi) =
    \log p(\vy|\theta(\tveta))
    + \frac{1}{2}\tvu^T\tmW\tvu
    + \frac{1}{2}\log|\tmW|$.
The approximate marginal is similar to that of standard GPR, but
uses the modified targets and noise terms, as discussed earlier.
There is one additional penalty
term $r(\phi)$ on the dispersion $\phi$, which
arises from the non-Gaussianity of the observation noise.
The derivatives of (\ref{eqn:cfa:marginal}) for using conjugate gradient are derived in Appendix~\ref{app:Taylor:marginal}.

\subsubsection{Transformed GPs as Taylor approximate inference}
\label{text:specialcases}

We now show that {\em heuristic} methods using GPs on {\em transformed outputs} can be explained as Taylor approximate inference on GGPMs with specific likelihoods.
The derivations, including derivative functions and targets
 used in Taylor approximate inference, are given in Appendix~\ref{text:Taylor:DeFun}.

\subsubsubsection{Binomial/Bernoulli}
\comments{We first consider the Taylor approximation for the binomial-GGPM. The
derivative functions are
    \begin{align}
    u(\eta,y) = N(y-\tfrac{e^\eta}{1+e^\eta}),
    \ \
    w(\eta,y) %
    = \tfrac{(1+e^\eta)^2}{N e^\eta}.
    \nonumber
    \end{align}
Thus, \abc{for a given expansion point $\teta_i$,} the target and effective noise are
    \begin{align}
    \tilde{t}_i = \teta_i + \tfrac{(1+e^{\teta_i})^2}{e^{\teta_i}} (y_i - \tfrac{e^{\teta_i}}{1+e^{\teta_i}} ), \
    \tilde{w}_i = \tfrac{(1+e^{\teta_i})^2}{N e^{\teta_i}}.
    \nonumber
    \end{align}}
An agnostic choice of expansion point is $\teta_i=0$, which ignores
the training classes,
leading to
    \begin{align}
    \tilde{t}_i = 4(y_i-0.5), \quad
    \tilde{w}_i =4/N.
    \end{align}
Hence, the Taylor approximation for binomial-GGPM is equivalent to
GPR in the latent space of the binomial model, with targets $\tilde{t}_i$
scaled between $[-2, +2]$ and an effective noise term $\tilde{w}_i=4/N$.
When $y_i \in \{0,1\}$, the target values are $\{-2, +2\}$, which is
equivalent to label regression \citep{GPML,Nickisch2008GPC,
Kapoor2010objects}, up to a scale factor.  Hence, {\em label regression
can be interpreted as a Taylor approximation to GPC inference.}  The
scaling of the targets ($\pm2$ or $\pm1$) is irrelevant when the latent space is only used for classifying
based on the sign of $\eta_*$.
However, this scaling is important when computing the actual
label probabilities using the predictive distribution.
The above interpretation explains why label regression tends to work well in practice, e.g., in \cite{Kapoor2010objects}.

\subsubsubsection{Poisson}
Based on the canonical expansion point, we have
$\teta_i = \log (y_i+c)$, where $c\geq0$ is a constant to
prevent taking the logarithm of zero, and hence
the target and effective noise for the Taylor approximation are
    \begin{align}
    \tilde{t}_i    %
    = \log(y_i+c) - \tfrac{c}{y_i+c},
    \ \
    \tilde{w}_i %
    = \tfrac{1}{y_i+c}.
    \end{align}
For $c=0$, the Taylor  approximation is exactly the closed-form
approximation proposed for Bayesian Poisson regression in
\citet{Chan2009iccv}, which was derived in a £¨£©different way using an approximation to the
log-gamma distribution.

\subsubsubsection{Gamma (shape hyperparameter)}
\comments{The derivatives of the Gamma\shape likelihood \abc{(mean parameter, shape hyperparameter)} are
    \begin{align}
    u(\eta,y) = \nu e^{-\eta}(y-e^{\eta}) = \nu(ye^{-\eta}-1),
    \ \
    w(\eta,y) %
    = \frac{1}{\nu}(1+ye^{-\eta}-1)^{-1}=\frac{1}{\nu ye^{-\eta}}.
    \nonumber
    \end{align}
Thus, given an expansion point $\teta_i$, the target and effective noise are
    \begin{align}
    \tilde{t}_i = \teta_i + \tfrac{1}{\nu y_i e^{-\teta_i}}\nu (y_i e^{-\teta_i}-1) = \teta_i+1-\tfrac{1}{y_i e^{-\teta_i}}, \quad
    \tilde{w}_i = \tfrac{1}{\nu y_i e^{-\teta_i}}.
    \end{align}
    }
Using the canonical expansion point, $\teta_i = \log y_i$,
yields the target and effective noise,
    \begin{align}
    \tilde{t}_i    %
    = \log y_i,
    \ \
    \tilde{w}_i %
    = \tfrac{1}{\nu} = \phi.
    \end{align}
Note that this is equivalent to using a standard GP on the log of
the outputs \citep{dregress:Diggle1998, Snelson04WGP}, which is standard practice
in the statistics literature when the observations are only positive values.  %
Hence, this practice of applying a GP on log-transformed outputs
is equivalent to assuming a Gamma likelihood and using Taylor approximate inference.

\subsubsubsection{Inverse Gaussian}
Using the canonical expansion point, $\teta_i = \log(2y_i^2)$,
yields the target and effective noise,
    \begin{align}
    \tilde{t}_i    %
    = \log (2y_i^2)=2\log y_i+\log 2,
    \ \
    \tilde{w}_i %
     = 4\phi y_i.
    \end{align}
This is equivalent to using a standard GP on the linear transform of log of
the outputs, and the noise $\tilde{w}_i$ is monotone in the value of output.

\subsubsubsection{Beta} %
\comments{
Consider an agnostic choice of the expansion point, $\teta_i=0$, and hence $\ttheta_i = \theta(\teta_i) = \frac{1}{2}$.  We also have,
    \begin{align}
    \sd{\theta}(\teta_i) = \frac{e^{\teta_i}}{(1+e^{\teta_i})^2} = \frac{1}{4},
    \quad
    \sdd{\theta}(\teta_i) = \frac{e^{\teta_i}(e^{\teta_i}-1)}{(1+e^{\teta_i})^3} = 0.
    \end{align}
Looking at the 1st and 2nd derivatives of $b(\theta)$ at $\ttheta_i$, we have
    \begin{align}
    \sd{b}(\ttheta_i) &= \psi_0(\tfrac{\ttheta_i}{\phi}) - \psi_0(\tfrac{1-\ttheta_i}{\phi})
    = \psi_0(\tfrac{1}{2\phi}) - \psi_0(\tfrac{1}{2\phi}) = 0, \\
    \sdd{b}(\ttheta_i) &= \tfrac{1}{\phi}\psi_1(\tfrac{\ttheta_i}{\phi}) + \tfrac{1}{\phi}\psi_1(\tfrac{1-\ttheta_i}{\phi})
    = \tfrac{2}{\phi} \psi_1(\tfrac{1}{2\phi}).
    \end{align}
Using the above results, we can now calculate the derivative functions at $\teta_i=0$,
    \begin{align}
    \tu_i = u(\teta_i,y_i) &= \frac{1}{\phi} \sd{\theta}(\teta_i) \left[ T(y_i) - \sd{b}(\ttheta_i)\right]
        = \frac{1}{4\phi} T(y_i) = \frac{1}{4\phi}\log \frac{y_i}{1-y_i},
        \\
    \tw_i = w(\teta_i,y_i) &= \frac{\phi}{\sdd{b}(\ttheta_i) \sd{\theta}(\teta_i)^2 - 0}
        = \frac{\phi}{\tfrac{2}{\phi}\psi_1(\tfrac{1}{2\phi})\tfrac{1}{4^2}}
        = \frac{8\phi^2}{\psi_1(\tfrac{1}{2\phi})}.
    \end{align}
which}
Consider an agnostic choice of the expansion point, $\teta_i=0$, yields the targets and noise,
    \begin{align}
    \tilde{t}_i %
    = \frac{2\phi}{\psi_1(\tfrac{1}{2\phi})} \log \frac{y_i}{1-y_i},
    \quad
        \tw_i = \frac{8\phi^2}{\psi_1(\tfrac{1}{2\phi})}.
    \end{align}
Using the approximation to the trigamma function, $\psi_1(x) \approx 1/x$, the targets are approximately
    \begin{align}
    \tilde{t}_i \approx \log \frac{y_i}{1-y_i},
    \quad
    \tilde{w}_i \approx 4\phi.
    \end{align}
Hence the GP on logit-transformed outputs of \citet{Heikkinen2008} is equivalent to using Taylor approximate inference and a Beta likelihood, with the above approximation to the trigamma function.
Alternatively, another choice is the canonical expansion point, $\teta_i = \log \tfrac{y_i}{1-y_i}$, which we used in our experiments.

\comments{\abc{
Now consider a tighter approximation, $\psi_1(x)\approx \frac{1}{x} + \frac{1}{2x^2}$,
and hence $\psi_1(\tfrac{1}{2\phi}) \approx 2\phi(1+\phi)$.
The targets of the Taylor approximation are now
    \begin{align}
    \tilde{t}_i \approx \frac{1}{1+\phi} \log \frac{y_i}{1-y_i},
    \quad
    \tilde{w}_i \approx \frac{4\phi}{1+\phi}.
    \end{align}
Since $\phi>0$, the effective noise is bounded $0<\tilde{w}_i < 4$,
and for large dispersion parameter, the targets are scaled down
towards 0. }
}

\subsection{Laplace approximation}

The Laplace approximation is a Gaussian approximation of the
posterior $p(\veta|\mX,\vy)$  at its maximum (mode). Hence, the
Laplace approximation is a specific case of the closed-form Taylor
approximation in Section \ref{text:Taylor}, where the expansion point
$\tveta$ is set to the maximum of the true posterior,
    \begin{align}
    \hat{\veta} = \argmax_{\veta} \log p(\veta|\mX, \vy).
    \end{align}
The true posterior mode is obtained iteratively using the
Newton-Raphson method, where in each iteration \citep{GPML},
    \begin{align}
    \hat{\veta}^{(new)} &= \hat{\veta} - \left[\pdd{}{\veta\veta^T} \log p(\hat{\veta}|\mX,\vy)\right]^{-1}  \pdd{}{\veta} \log p(\hat{\veta}|\mX,\vy) %
    = (\hat{\mW}^{-1} + \mK^{-1})^{-1}\hat{\mW}^{-1} \hat{\vt},
    \end{align}
where $\hat{\vu}$ and $\hat{\mW}$ are evaluated at $\hat{\veta}$,
and $\hat{\vt} = \hat{\mW}\hat{\vu}+\hat{\veta}$.
In each iteration the expansion point $\hat{\veta}$ is moved closer to the maximum,
and the target vector  $\hat{\vt}$ is updated.
Note that the update for $\hat{\veta}$ is of the same form as the mean $\hat{\vm}$ in the closed-form Taylor approximation.   Hence, the Taylor approximation could also be considered a one-iteration Laplace approximation, using the expansion point $\tveta$ as the initial point.

The parameters of the approximate posterior of $\veta$ and $\eta_*$ are
    \begin{align}
    \hat{\vm} &= \hat{\veta}, & \hat{\mV} &= ( \hat{\mW}^{-1} + \mK^{-1})^{-1}, \\
    \hat{\mu}_{\eta} &= \vk_*^T \mK^{-1} \hat{\veta} =\vk_*^T \hat{\vu},
    &
    \hat{\sigma}_{\eta} &= k_{**} - \vk_*^T (\mK + \hat{\mW})^{-1} \vk_* .
    \label{eqn:laplace:pred}
    \end{align}
The mode is unique when the log posterior is concave,
or equivalently when $\mW^{-1}$ is positive definite, i.e., $\forall y,\eta$,
    \begin{align}
    \nonumber
    w(\eta,y)^{-1}
    &= \tfrac{1}{a(\phi)}
    \left\{ \sdd{b}(\theta(\eta)) \sd{\theta}(\eta)^2
    -   \left[T(y) - \sd{b}(\theta(\eta))\right]\sdd{\theta}(\eta) \right\} > 0,
    \\
    &\quad \Rightarrow
    \sdd{b}(\theta(\eta)) \sd{\theta}(\eta)^2 >
    \left[T(y) - \sd{b}(\theta(\eta))\right]\sdd{\theta}(\eta).
    \nonumber
    \end{align}
For a canonical link function, this simplifies to
     $\sdd{b}(\eta) >   0$,
i.e., a unique maximum exists when $b(\eta)$ is convex.
Finally, the Laplace approximation for the marginal likelihood is
    \begin{align}
    \log q(\vy|\mX)
    &= \log p(\vy|\theta(\hat{\veta}))
     - \frac{1}{2} \hat{\veta}^T\mK^{-1} \hat{\veta} - \frac{1}{2}\log \detbar{\hat{\mW}^{-1}\mK+\mI}
     \\
    &= \log p(\vy|\theta(\hat{\veta})) - \frac{1}{2} \hat{\vu}^T\mK\hat{\vu} - \frac{1}{2}\log \detbar{\hat{\mW}^{-1}\mK+\mI}.
    \label{eqn:lap:approxmarg}
    \end{align}
where the last line follows from the first-derivative condition at
the maximum.
Note that $\hat{\veta}$ is dependent on the kernel matrix $\mK$ and
$\phi$.  Hence, at each iteration during optimization of
$q(\vy|\mX)$, we have to recompute its value. Derivatives of the
marginal are presented in the supplemental \citep{Chan2013tr}.
\subsection{Expectation propagation}

Expectation propagation (EP) \citep{Minka2001} is a general
algorithm for approximate inference, which has been shown to be
effective
for GPC \citep{Nickisch2008GPC}.
EP approximates each likelihood term $p(y_i|\theta(\eta_i))$ with %
an unnormalized Gaussian $t_i = \tZ_i
\Normalv{\eta_i}{\tmu_i}{\tsigma^2_i}$ (also called a site
function), yielding an approximate data likelihood
    \begin{align}
    \nonumber
    q(\vy|\theta(\veta)) = \prod_{i=1}^n t_i(\eta_i|\tZ_i, \tmu_i, \tsigma^2_i) = \Normalv{\veta}{\tvmu}{\tmSigma}\prod_{i=1}^n \tZ_i,
    \end{align}
where $\tvmu = [\tmu_1,\cdots \tmu_n]^T$ and $\tmSigma =
\diag([\tsigma^2_1,\cdots, \tsigma^2_n])$. Using the site functions,
the posterior approximation is
    \begin{align}
    \nonumber
    q(\veta|\mX,\vy) = \frac{1}{Z_{EP}} \prod_{i=1}^n t_i(\eta_i) p(\veta|\mX) = \Normalv{\veta}{\hat{\vm}}{\hat{\mV}}
    \end{align}
where $\{\hat{\vm},\hat{\mV}\}$ are in the form of
\refeqn{eqn:approxpost:common}, and hence
    \begin{align}
    \hat{\vm} &= \hat{\mV}\tmSigma^{-1}\tvmu, &
    \hat{\mV} &= (\mK^{-1} + \tmSigma^{-1})^{-1},
    \label{eqn:ep:postparams}
    \\
    \hat{\mu}_{\eta}
    &= \vk_*^T (\mK + \tmSigma )^{-1} \tvmu
    ,&
    \hat{\sigma}_{\eta}^2
    &= k_{**} - \vk_*^T (\mK + \tmSigma)^{-1} \vk_* .
    \end{align}
The normalization constant is also the EP approximation to the
marginal likelihood \citep{Nickisch2008GPC,GPML},
    \begin{align}
    \log Z_{EP} &= \log q(\vy|\mX) = \log \int q(y|\theta(\veta)) p(\veta|\mX) d\veta
    \\ &=
     -\frac{1}{2} \tvmu^T(\mK+\tmSigma)^{-1} \tvmu - \frac{1}{2}\log \detbar{\mK+\tmSigma}
     + \sum_i \log \tZ_i.
    \label{eqn:ep:marg}
    \end{align}
Derivatives of \refeqn{eqn:ep:marg} are presented in the
supplemental \citep{Chan2013tr}.

\subsubsection{Computing the site parameters}

Instead of computing the optimal site parameters all at once, EP
works by iteratively  updating each individual site using the other
site approximations \citep{Minka2001,GPML}.  In particular,
to update site $t_i$, we first compute the {\em cavity
distribution}, which is the marginalization over all sites except
$t_i$,
    \begin{align}
    q_{\noti}(\eta_i)=\Normalv{\eta_i}{\mu_{\noti}}{\sigma^2_{\noti}} &\propto \int p(\veta|\mX) \prod_{j\neq i} t_j(\eta_j|\tZ_j, \tmu_j, \tsigma^2_j) d\eta_j,
    \end{align}
where the notation $\noti$ indicates the sites without $t_i$,
and
$q_{\noti}(\eta_i)$ is an approximation to the posterior
distribution of $\eta_i$, given all observations except $y_i$.
 Since both terms are Gaussian, this integral can be computed in closed-form. %
Next, the site parameters of $t_i$ are selected to match the moments
(mean, variance, and normalization) between $\hat{q}(\eta_i) =
p(y_i|\theta(\eta_i))q_{\noti}(\eta_i)$ and
$t_i(\eta_i)q_{\noti}(\eta_i)$. %
This requires first calculating the moments of $q(\eta_i) =
\frac{1}{\hat{Z}_i} p(y_i|\theta(\eta_i))
\Normalv{\eta_i}{\mu_{\noti}}{\sigma^2_{\noti}}$,
    \begin{align}
    \label{eqn:ep:momentfunc}
    \hat{\mu}_i &=   \EV_q[\eta_i], \quad
    \hat{\sigma}^2_i =  \var_q(\eta_i), \quad
    \hat{Z}_i = \int p(y_i|\theta(\eta_i)) q_{\noti}(\eta_i)
    d\eta_i, %
    \end{align}
followed by ``subtracting'' the cavity distribution and then
yielding the site updates.
    \begin{align}
    \tmu_i &= \tsigma_i^2 (\hat{\sigma}_i^{-2} \hat{\mu}_i - \sigma^{-2}_{\noti} \mu_{\noti}),
    \quad
    \tsigma_i^2 = (\hat{\sigma}_i^{-2} - \sigma_{\noti}^{-2})^{-1} ,
    \\
    \log \tZ_i &= \log \hat{Z}_i + \frac{1}{2}\log 2\pi (\sigma^2_{\noti} + \tsigma^2_i) + \frac{(\mu_{\noti}-\tmu_i)^2}{2(\sigma^2_{\noti} + \tsigma^2_i)}.
    \end{align}
EP iterates over each of the site $t_i$, i.e. each observation
$y_i$, iteratively until convergence. Note that in general, EP is
not guaranteed to converge.  Although it is usually well behaved
when the data log-likelihood $\log p(y_i|\theta(\eta_i))$ is
concave and the approximation is initialized to the prior~\citep{Nickisch2008GPC,Jylanki2011}.
Finally, these moments may not be analytically tractable (in fact,
$q(\eta_i)$ is the same form as the predictive distribution), so
approximate integration is usually required.  In general, the
convergence of EP also depends on the accuracy of the moment
approximations.

\subsection{KL divergence minimization}
In this section we discuss a variational approximation that
maximizes a lower-bound of the marginal likelihood by minimizing the
KL divergence between the approximate posterior and the true
posterior. This type of approximate inference was first applied to
robust GP regression in \citet{Opper2009}, and later to GP
classification in \citet{Nickisch2008GPC}.  In this paper, we extend
it to the GGPM.

As with other approximations, the approximate posterior is assumed to be Gaussian,
    \begin{align}
    p(\veta|\mX, \vy) \approx q(\veta|\mX, \vy) = \Normalv{\veta}{\vm}{\mV}
    \end{align}
for some $\vm$ and $\mV$.  To obtain the best approximate posterior,
the KL divergence is minimized between the approximation and the
true posterior,
    \begin{align}
    \{\vm^*,\mV^*\} = \argmin_{\vm,\mV} \KLp{\Normalv{\veta}{\vm}{\mV}}{p(\veta|\mX, \vy)}.
        \label{eqn:KL:KLD}
    \end{align}
As shown in
\citet{Nickisch2008GPC}, minimizing the KL in \refeqn{eqn:KL:KLD} is
equivalent to maximizing the lower bound of $\log p(\vy|\mX)$
 \begin{align}
    \calL
    &= \log p(\vy|\mX) - \KLp{\Normalv{\veta}{\vm}{\mV}}{p(\veta|\mX, \vy)}\\
    &=f(\vm,\vv)
    + \frac{1}{2}\log \detbar{\mK^{-1}\mV} - \frac{1}{2}\tr(\mK^{-1}\mV) - \frac{1}{2}\vm^T\mK^{-1}\vm  + \frac{n}{2}.
    \label{eqn:KL:ZLBfinal}
 \end{align}
The function $f(\vm,\vv)$ is the expectation of the observation log-likelihood,
 \begin{align}
     f(\vm, \vv) \equiv \int \Normalv{\veta}{\vm}{\mV} \log p(\vy|\theta(\veta)) d\veta = \sum_{i=1}^n \EV_{\eta_i|m_i,v_i} \left[\log p(y_i|\theta(\eta_i))\right],
    \end{align}
where $m_i = [\vm]_i$ and $v_i = [\mV]_{ii}$, $\vv = \diag(\mV)$, and
$\EV_{\eta_i|m_i,v_i}[\cdot]$ is the expectation with respect to
$\Normalv{\eta_i}{m_i}{v_i}$. The first term of
\refeqn{eqn:KL:ZLBfinal} emphasizes the fit to the data (via the
expectation), while remaining terms (KL divergence terms) penalize
the posterior from being too different from the prior distribution.

At a local maximum of \refeqn{eqn:KL:ZLBfinal}, the first derivative
conditions yield the following constraints,
    \begin{align}
    \hat{\vm} = \mK\hat{\vgamma},
    \quad
     \hat{\mV} = (\mK^{-1} + \hat{\mLambda} )^{-1}
    \label{eqn:varkl:meancov}
    \end{align}
where the optimal $\hat{\vgamma}$ and $\hat{\mLambda}$ satisfy
    \begin{align}
    \hat{\vgamma} = \pdd{f(\vm,\vv)}{\vm},
    \quad
    \hat{\mLambda} = -2 \pdd{f(\vm,\vv)}{\mV}
    \label{eqn:varkl:Lambda}
    \end{align}
Since the mean and covariance have the forms in
(\ref{eqn:varkl:meancov}), the optimization problem can be reformulated with the variational parameters
$\{\vgamma,\vlambda\}$, such that
    \begin{align}
    \vm = \mK\vgamma,
    \quad
    \mV = (\mK^{-1} + \mLambda)^{-1} = \mK(\mI + \mLambda\mK)^{-1},
    \quad
    \mLambda = \diag(\vlambda).
    \end{align}
This parameterization also avoids inverting the kernel
matrix.
Substituting into \refeqn{eqn:KL:ZLBfinal},
    \begin{align}
    \calL
    &= f(\mK\vgamma,\vv)
    + \frac{1}{2}\log \detbar{(\mI+\mLambda\mK)^{-1}} - \frac{1}{2}\tr((\mI+\mLambda\mK)^{-1}) - \frac{1}{2}\vgamma^T\mK\vgamma  + \frac{n}{2}.
    \label{eqn:KL:ZLBfinal_var}
    \end{align}
Finally, the optimal approximate posterior can be obtained by
maximizing \refeqn{eqn:KL:ZLBfinal_var} with respect to the
variational parameters $\{\vgamma, \vlambda\}$ using standard
optimization techniques (e.g., the conjugate gradient method).
Note that $\calL$ is a lower bound on the marginal likelihood, as in
\refeqn{eqn:KL:ZLBfinal}.  Hence, the model hyperparameters can also
be estimated by maximizing \refeqn{eqn:KL:ZLBfinal_var}.
In practice, the model hyperparameters and approximate posterior can
be estimated at the same time, by jointly maximizing
\refeqn{eqn:KL:ZLBfinal_var} with respect to all parameters (again,
e.g., using conjugate gradient).

\subsubsection{Expectation terms}

The computation of  \refeqn{eqn:KL:ZLBfinal_var} and its derivatives
requires calculating $f(\vm,\vv)$ and its derivatives,
    \begin{align}
    f(m_i,v_i) &= \EV_{\eta_i|m_i,v_i} \left[\log p(y_i|\theta(\eta_i))\right] ,  &
    \pdd{f(m_i,v_i)}{\phi}  &= \pdd{}{\phi}\EV_{\eta_i|m_i,v_i} \left[\log p(y_i|\theta(\eta_i))\right] ,
    \\
    \pdd{f(m_i,v_i)}{m_i} &= \pdd{}{m_i}\EV_{\eta_i|m_i,v_i} \left[\log p(y_i|\theta(\eta_i))\right] , &
    \pdd{f(m_i,v_i)}{v_i} &= \pdd{}{v_i}\EV_{\eta_i|m_i,v_i} \left[\log p(y_i|\theta(\eta_i))\right].
    \end{align}
Plugging in for the exponential family form, the first two terms can
be rewritten as
    \begin{align}
    f(m_i,v_i) &= %
    \frac{1}{a(\phi)}\left\{ T(y_i) \EV_{\eta_i|m_i,v_i}[\theta(\eta_i)] -
        \EV_{\eta_i|m_i,v_i}[b(\theta(\eta_i))] \right\} + c(\phi, y_i),
        \\
    \pdd{f(m_i,v_i)}{\phi} &= %
    \frac{-\sd{a}(\phi)}{a(\phi)^2}\left\{ T(y_i) \EV_{\eta_i|m_i,v_i}[\theta(\eta_i)] -
        \EV_{\eta_i|m_i,v_i}[b(\theta(\eta_i))] \right\} + \sd{c}(\phi,y_i).
    \end{align}
Hence, the expectations $\EV[\theta(\eta)]$ and $\EV[b(\theta(\eta))]$
under a Gaussian distribution are required. Expressions of the last
two terms can be obtained by directly taking the derivative,
    \begin{align}
    \label{eqn:df_dm}
    \pdd{f(m_i,v_i)}{m_i} &=
    \int \pdd{\Normalv{\eta_i}{m_i}{v_i}}{m_i} \log p(y_i|\theta(\eta_i)) d\eta_i
    =
    \EV_{\eta_i|m_i,v_i}\left[ \frac{\eta_i-m_i}{v_i}\log p(y_i|\theta(\eta_i))\right] ,
    \\
    \label{eqn:df_dv}
    \pdd{f(m_i,v_i)}{v_i} &=
    \int \pdd{\Normalv{\eta_i}{m_i}{v_i}}{v_i} \log p(y_i|\theta(\eta_i)) d\eta_i
    =
    \EV_{\eta_i|m_i,v_i}\left[\frac{(\eta_i-m_i)^2-v_i}{2v_i^2}\log p(y_i|\theta(\eta_i))\right].
    \end{align}
Hence, these two derivatives require the expectations $\EV[\eta^k
\theta(\eta)]$ and $\EV[\eta^k b(\theta(\eta))]$, where
$k\in\{1,2\}$. For certain likelihood and link functions (e.g.,
Poisson with canonical link), the above expectations have a closed
form solutions.  In other cases, they need to be approximated.

Alternative expressions to (\ref{eqn:df_dm}, \ref{eqn:df_dv}) can be
obtained by performing a change of variable in the expectation $\eta
= \tfrac{\bar{\eta}-m}{\sqrt{v}}$,
    \begin{align}
    \pdd{f(m_i, v_i)}{m_i} &=
        \pdd{}{m_i} \EV_{\bar{\eta}_i|0,1}[ \log p(y_i|\theta(\sqrt{v_i}\bar{\eta}_i+m_i))]
        = \EV_{\eta_i|m_i,v_i}[ u(\eta_i,y_i) ] ,
        \label{eqn:KL:df_dm_u}
        \\
    \pdd{f(m_i,v_i)}{v_i} &=
        \pdd{}{v_i} \EV_{\bar{\eta}|0,1}[ \log p(y_i|\theta(\sqrt{v_i}\bar{\eta}+m_i)) ]
        =  \frac{1}{2}\EV_{\eta_i|m_i,v_i}\left[ \frac{\eta_i-m_i}{v_i} u(\eta_i,y_i) \right].
        \label{eqn:KL:df_dv_u}
    \end{align}
Hence, alternatively the expectations
 $\EV[\eta^k\sd{\theta}(\eta)]$ and
$\EV[\eta^k\sd{b}(\theta(\eta))\sd{\theta}(\eta)]$, $k\in\{0,1\}$,
are required under a Gaussian. The alternative forms in
(\ref{eqn:KL:df_dm_u}, \ref{eqn:KL:df_dv_u}) allow an intuitive
comparison  between the KLD method and the Laplace approximation,
given in the next section.

\subsubsection{Approximate posterior}

After maximizing $\calL$, resulting in optimal variational
parameters $\{\hat{\vgamma}, \hat{\vlambda}\}$, the approximate
posteriors have parameters
    \begin{align}
     \hat{\vm} &= \mK\hat{\vgamma},
    &
    \hat{\mV} &= (\mK^{-1} + \hat{\mLambda} )^{-1} ,
    \\
    \hat{\mu}_{\eta} &= \vk_*^T \hat{\vgamma},
    &
    \hat{\sigma}^2_{\eta}   &= k_{**} - \vk_*^T (\mK + \hat{\mLambda}^{-1})^{-1} \vk_* .
    \end{align}
An interesting comparison can be made against the predictive latent
distribution using the Laplace approximation in
\refeqn{eqn:laplace:pred}. In particular, with the Laplace
approximation, the latent mean depends on the first derivative $\vu$
of the observation log-likelihood (at the mode), whereas with the
variational method, the latent mean depends on the {\em expectation}
of the first derivative, $\hat{\vgamma} = \EV[\vu]$, using
\refeqn{eqn:KL:df_dm_u}.
Similarly, with the Laplace approximation, the effective noise term
$\mW$ is the inverse of the 2nd derivative of the observation
log-likelihood, and with KLD, this term depends on an estimate of
the 2nd derivative by differencing the 1st derivative around $m$,
given by \refeqn{eqn:KL:df_dv_u}.

\subsection{Summary}

In this section, we have %
studied closed-form Taylor approximation, and discussed its connections with output-transformed GPs.
We also discussed other popular approximate inference methods in the context of GGPMs.
Using the general EFD form for the likelihood, we can identify the specific
quantities required for each
algorithm in terms of parameters $\calE$, as summarized in Table \ref{tab:summary_approxinf}.
For example,
EP requires the expectations of $\theta(\eta)$ and $b(\theta(\eta))$
under the approximate predictive distribution, whereas KL
minimization requires these expectations under a Gaussian
distribution.
This result has two practical consequences:
1) the implementation of
likelihood functions is simplified,
since only the derivatives and expectations  of simple functions in $\calE$ need to be implemented;
2) we elucidate the expectations that may require  numerical approximation.
Furthermore, different likelihood and link functions can be combined in novel ways without much additional implementation effort.

\comments{
\begin{table}\scriptsize
\centering
\begin{tabular}{@{}lc|cc@{}}
\hline
Appro. method & lik. & posterior \& marginal & marginal derivatives
\\
\hline Taylor
  & general
  & $\pdd{^k}{\eta^k}\log p(y|\eta), \ k=\{1,2\}$ & $\pdd{}{\phi}\log p(y|\eta)$ \\
  & GGPM
  & $\sd{b}, \sdd{b}, \sd{\theta}, \sdd{\theta}$ & $\sd{a}, \sd{c}$
\\\hline
Laplace
  & general
  & $\pdd{^k}{\eta^k}\log p(y|\eta), \ k=\{1,2\}$ & $\pdd{^3}{\eta^3}\log p(y|\eta), \pdd{}{\phi}\pdd{^2}{\eta^2}\log p(y|\eta), \pdd{}{\phi}\log p(y|\eta)$\\
  & GGPM
  & $\sd{b}, \sdd{b}, \sd{\theta}, \sdd{\theta}$ & $\sd{a}, \sd{c}, \sddd{b},\sddd{\theta}$
\\\hline
EP
  & general
  & $\log \hat{Z}_i, \pdd{^k}{\eta_i^k}\log \hat{Z}_i, k=\{1,2\}$ & $\pdd{}{\phi} \log \hat{Z}_i$ \\
  & GGPM
  & $\EV_q[\eta_i], \var_q(\eta_i)$ & $\sd{c}, \sd{a}, \EV_q[\theta(\eta)], \EV_q[b(\theta(\eta))]$
\\\hline
KLD
  & general
  & $\EV[\eta^k\log p(y|\eta)]$, $k\in\{0,1,2\}$ & $\EV[\pdd{}{\phi}\log p(y|\eta)]$\\
  & GGPM
  & $\EV[\eta^k\theta(\eta)], \EV[\eta^kb(\theta(\eta))]$, $k\in\{0,1,2\}$ & $\sd{c}, \sd{a}$
\\
 \hline
\end{tabular}\caption{The required calculations for each approximation}
\end{table}
}

\begin{table}[htb]
\scriptsize
\centering
\begin{tabular}{@{}ll|cc@{}}
\hline
\multicolumn{2}{@{}l|}{Approximation} & general likelihood& GGPM likelihood
\\
\hline Taylor
  & posterior \& marginal
  & $\pdd{^k}{\eta^k}\log p(y|\eta), \ k=\{1,2\}$
  & $\sd{b}, \sdd{b}, \sd{\theta}, \sdd{\theta}$
  \\
  & marginal derivatives
  & $\pdd{}{\phi}\log p(y|\eta)$
  & $\sd{a}, \sd{c}$
\\\hline
Laplace
  & posterior \& marginal
  & $\pdd{^k}{\eta^k}\log p(y|\eta), \ k=\{1,2\}$
  & $\sd{b}, \sdd{b}, \sd{\theta}, \sdd{\theta}$
  \\
  & marginal derivatives
  & $\pdd{^3}{\eta^3}\log p(y|\eta), \pdd{}{\phi}\pdd{^k}{\eta^k}\log p(y|\eta), k\in\{0,2\}$
  & $\sd{a}, \sd{c}, \sddd{b},\sddd{\theta}$
\\\hline
EP
  & posterior \& marginal
  & $\log \hat{Z}_i, \pdd{^k}{\eta_i^k}\log \hat{Z}_i, k=\{1,2\}$
  & $\EVscript_q[\eta_i], \var_q(\eta_i)$
  \\
  & marginal derivatives
  & $\pdd{}{\phi} \log \hat{Z}_i$
  & $\sd{c}, \sd{a}, \EVscript_q[\theta(\eta)], \EVscript_q[b(\theta(\eta))]$
\\\hline
KLD
  & posterior \& marginal
  & $\EVscript[\eta^k\log p(y|\eta)]$, $k\in\{0,1,2\}$
  & $\EVscript[\eta^k\theta(\eta)], \EVscript[\eta^kb(\theta(\eta))]$, $k\in\{0,1,2\}$
\\
  & marginal derivatives
  & $\EVscript[\pdd{}{\phi}\log p(y|\eta)]$
  & $\sd{c}, \sd{a}$
\\
 \hline
\end{tabular}\caption{The required calculations of the likelihood function for approximate inference.
The
expectations $\EV_q$ and $\var_q$ can be calculated from derivatives
of $\log \hat{Z}_i$.
} \label{tab:summary_approxinf}
\end{table}

\subsection{Implementation Details}

The GGPM was implemented in MATLAB
by extending the GPML toolbox \citep{GPMLcode} to
include implementations for: 1) the generic exponential family
distribution using the parameters $\{a(\phi), b(\theta), c(y,\phi),
\theta(\eta), T(y)\}$; 2) the closed-form Taylor approximation for
inference; 3) the EP and KLD moments and the predictive distributions, approximated using numerical integration when necessary. Empirically, we found that
EP was sensitive to the accuracy of the approximate integrals,
and exhibited convergence problems when less accurate approximations were used (e.g.
Gaussian-Hermite quadrature). Hyperparameters (dispersion and kernel
parameters) were optimized by maximizing the marginal likelihood,
using the existing scaled conjugate gradient method in GPML.
The code will be made available\footnote{http://visal.cs.cityu.edu.hk/downloads/}.

\section{Comparison of approximate posteriors}
\label{1D_example}

In this section, we provide a theoretical and experimental comparison of the approximate posteriors from Section \ref{text:appinf}.
In particular, we show that the efficacy of an approximate inference method is
influenced by properties of the likelihood, evaluation metrics, and datasets.

\subsection{Ordering of posterior means and predictive means}

\begin{figure}[tp]
\vspace{-0.2in} \centering
\begin{tabular}{c}
 \includegraphics[scale=0.36]{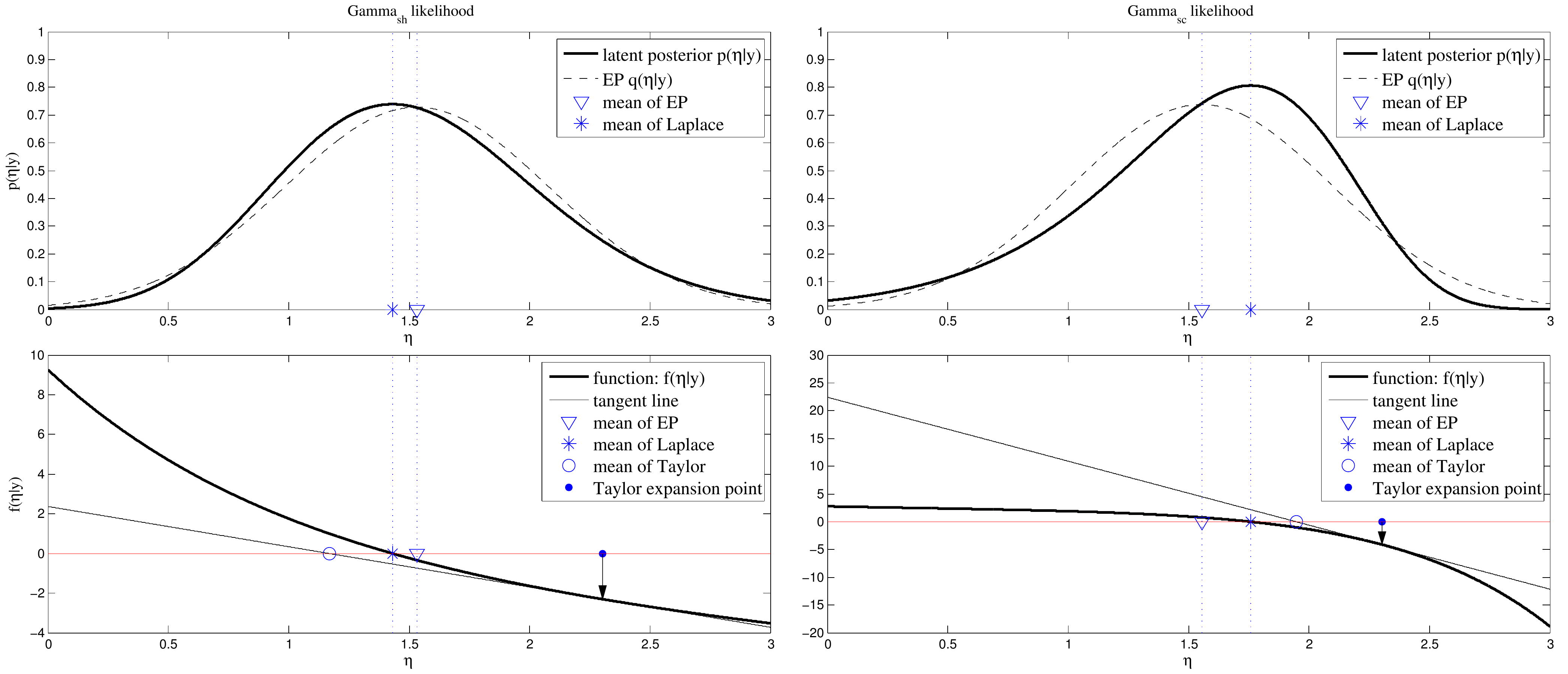}
\end{tabular}
\caption{Comparing the approximate posterior means of Taylor,
Laplace and EP methods for the Gamma-shape (left) and Gamma-scale (right)
likelihood functions. The first row shows the true latent
posterior $p(\eta|y)$ and the EP approximation $q(\eta|y)$. The
second row shows the first derivative of the log-posterior
$f(\eta|y) = \pdd{}{\eta} \log p(\eta|y)$. The mean of the Laplace
approximation is the zero-crossing point of $f(\eta|y)$.  The mean
of the Taylor approximation is the zero-crossing of the tangent line
at the expansion point (one iteration of Newton's method).
}
\label{fig:Gamma_T_L}
\end{figure}

We first compare the latent posteriors of the Taylor, Laplace, and EP approximations for one latent variable.
Consider an example using the Gamma-shape likelihood,
and the corresponding true latent posterior $p(\eta|y) \propto p(y|\eta)p(\eta)$ in Figure~\ref{fig:Gamma_T_L} (top-left).
The first derivative of the log-posterior, $f(\eta|y) = \pdd{}{\eta}\log p(\eta|y)$, is plotted in Figure~\ref{fig:Gamma_T_L} (bottom-left).
Note that the derivative of the Gamma-shape log-likelihood, $u(\eta,y)=\nu(ye^{-\eta}-1)$, is convex and monotonically decreasing for all $y>0$.

\begin{claim}
\label{claim:postmeans}
If the derivative of the observation log-likelihood, $\pdd{}{\eta}\log p(y|\eta)$, is convex and monotonically decreasing, then the means of the 1-D approximate posteriors are ordered according to
    \begin{align}
    \mu_{TA} < \mu_{LA} < \mu_{EP},
    \label{eqn:posterior-order}
    \end{align}
where $\mu_{TA}$, $\mu_{LA}$, and $\mu_{EP}$ are the latent means for the Taylor approximation, Laplace approximation, and EP, respectively.
The ordering is reversed when $\pdd{}{\eta}\log p(y|\eta)$ is concave and decreasing.

\begin{proof}
The log posterior is $\log p(\eta|y) \propto \log p(y|\eta) + \log p(\eta)$, and the derivative is
    \begin{align}
    f(\eta|y) = \pdd{}{\eta}\log p(y|\eta) + \pdd{}{\eta} \log p(\eta)
    \label{eqn:feta1d}
    \end{align}
The first term on the RHS of \refeqn{eqn:feta1d} is assumed to be convex and monotonically decreasing, while the second term is a linear function with negative slope (derivative of the Gaussian log-likelihood).
Hence, $f(\eta|y)$ is also convex and monotonically decreasing.
The mean of the Laplace approximation is the mode of the true posterior, i.e., the zero crossing of $f(\eta|y)$, and is marked with a star (*) in Figure \ref{fig:Gamma_T_L}.
The Taylor approximation is equivalent to one iteration of Newton's method on $f(\eta|y)$, starting at the expansion point $\teta$.  Hence, geometrically, the mean of the Taylor approximation is the zero-crossing of the tangent line to $f(\eta|y)$ at the expansion point (marked with a circle (o) in Figure \ref{fig:Gamma_T_L}).
Since $f(\eta|y)$ is convex and monotonically decreasing,
the zero-crossing point of any tangent line is always less than the zero-crossing point of $f(\eta|y)$.  Therefore, the Taylor mean is always smaller than the Laplace mean.

Because $f(\eta|y)$ is monotonically decreasing and convex, the posterior $p(\eta|y)$ is skewed to the right, and its mean is larger than the mode.
Since EP matches the mean of the approximation to the mean of the true posterior, the EP mean must be larger than the mode (i.e., the Laplace mean).
\end{proof}
\end{claim}

\setlength{\myw}{0.23\linewidth}

\setlength{\myw}{0.23\linewidth}
\begin{figure}[thbp]
\vspace{-0.3in} \centering
\begin{tabular}{ccc}
& \footnotesize{varying $\phi$}
& \footnotesize{varying $k_w$}
\\
\raisebox{0.8in}{\footnotesize{TA vs. LA}} &
 \includegraphics[scale=0.50]{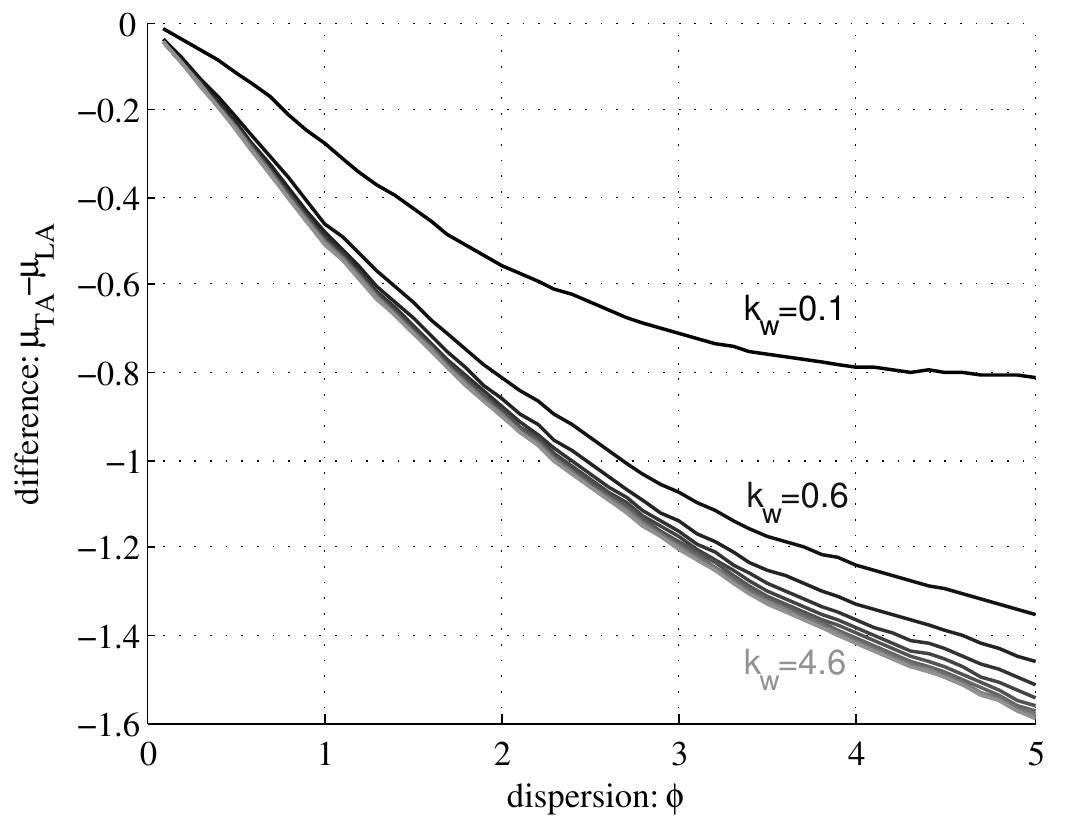}
 & \hspace{-0.2in}
 \includegraphics[scale=0.50]{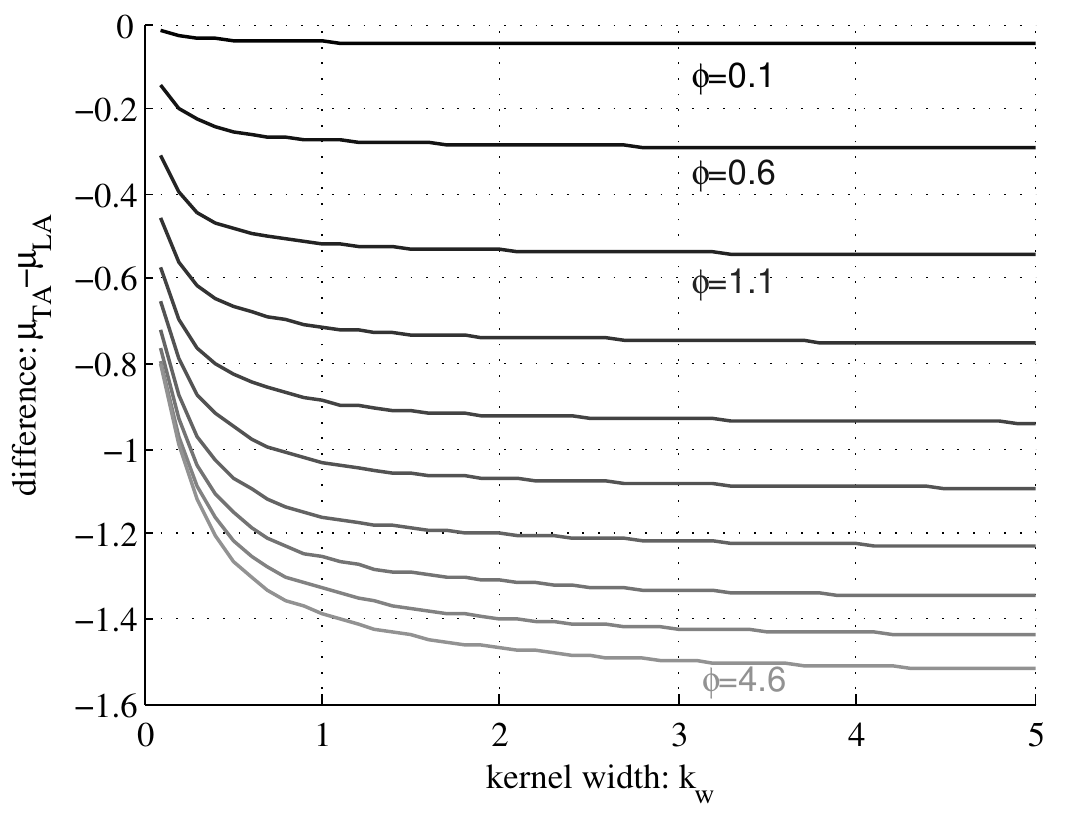}
 \\
\raisebox{0.8in}{\footnotesize{LA vs. EP}} &
 \includegraphics[scale=0.50]{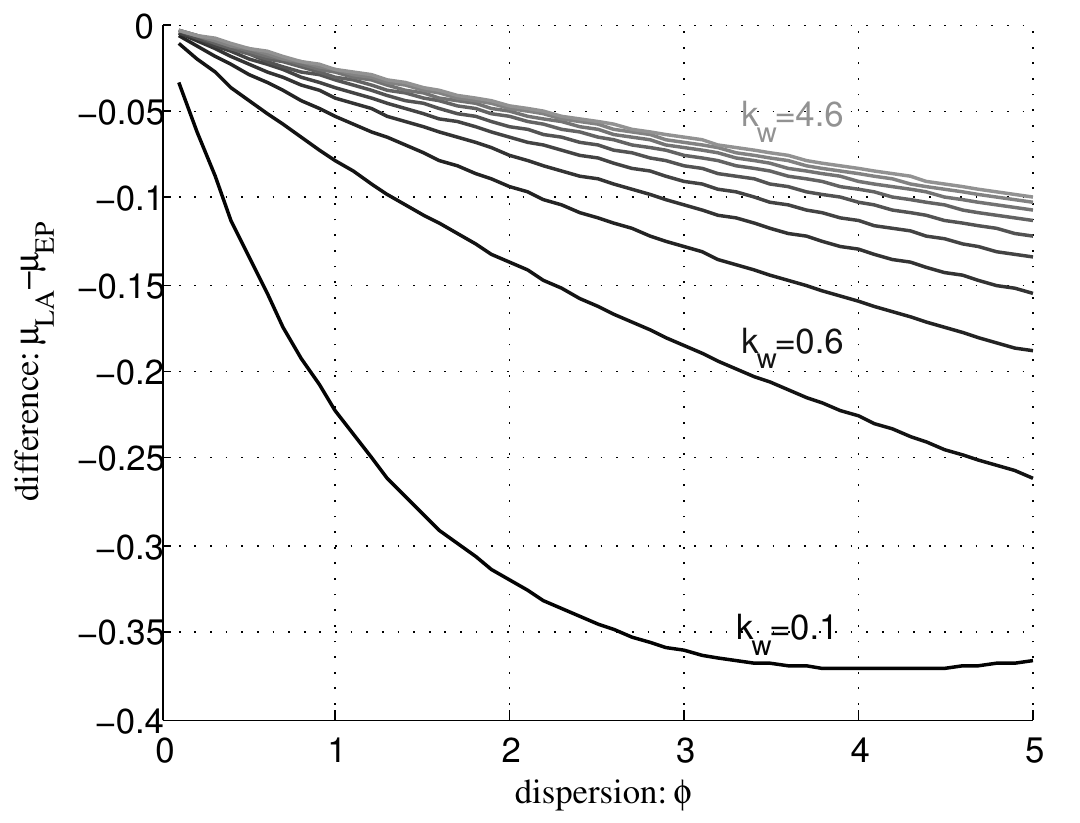}
 & \hspace{-0.2in}
 \includegraphics[scale=0.50]{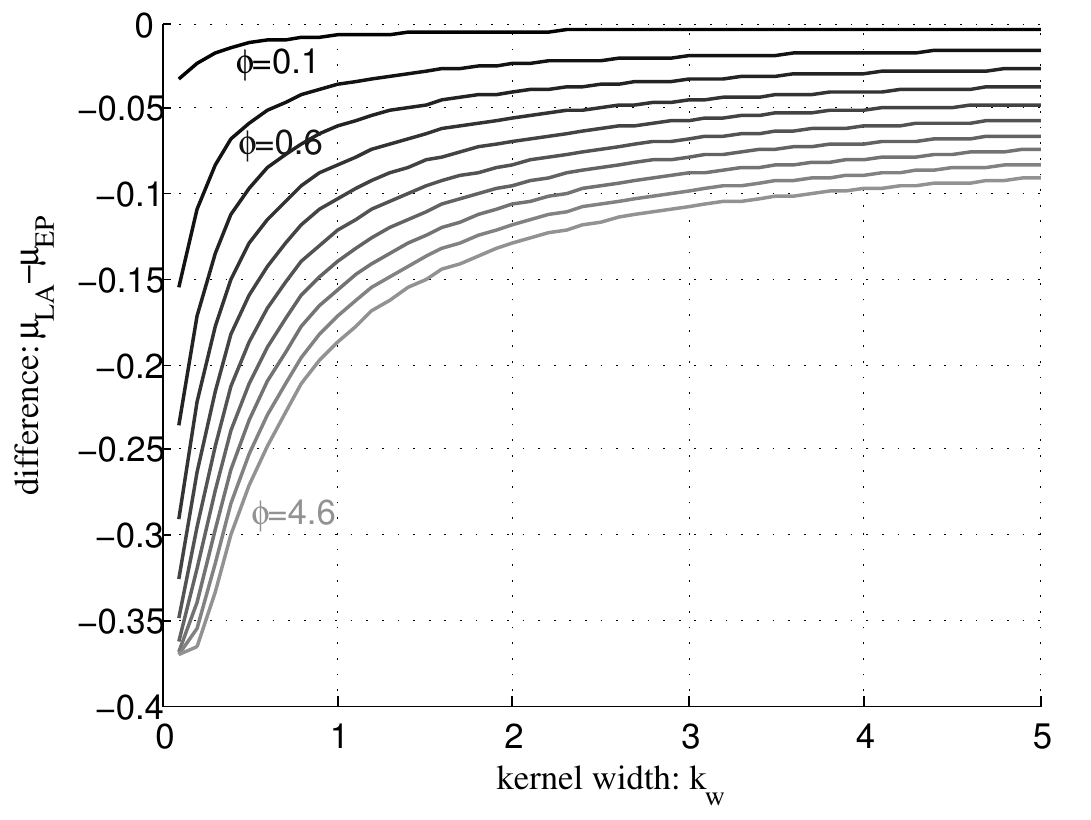}
 \vspace{-0.1in}
\end{tabular}%
\caption{The difference between approximate posterior means for the Gamma-shape
likelihood function.
The top row compares Taylor ($\mu_{TA}$) vs. Laplace ($\mu_{LA}$), while
the bottom row compares Laplace ($\mu_{LA}$) vs. EP ($\mu_{EP}$).
In each row, the two plots show the difference w.r.t the dispersion parameter $\phi$
and the RBF kernel bandwidth  $k_{w}$.
}
\label{fig:figDiffTALA}
\end{figure}

Claim \ref{claim:postmeans} suggests that the posterior means follow a particular order for the {\em 1-dimensional}  posterior.  For the general multi-dimensional case, it is difficult to prove a similar result, since the dimensions of the latent posterior are correlated through the kernel matrix\footnote{When the kernel matrix is diagonal, i.e., the kernel function is a delta function,  it is easy to show that the ordering in Claim \ref{claim:postmeans} will hold for each dimension.}.
Nonetheless, we can empirically show that the ordering in \refeqn{eqn:posterior-order} holds for multivariate posteriors on average.

To this end, we ran a synthetic experiment using the Gamma-shape likelihood, with dispersion parameter $\phi \in [0.1, 5]$,  and RBF kernel function with scale $K_s = 2$ and bandwidth $K_w\in [0.1, 5]$.
For a given bandwidth $K_w$ and dispersion $\phi$ pair, 100 different functions are first randomly sampled from a Gamma-shape GGPM. For each function, 40 points are used for training, and the multivariate means of the approximate posteriors $\vmu_{TA}$, $\vmu_{LA}$, and $\vmu_{EP}$ are calculated using Taylor, Laplace, and EP, respectively. The differences between these means are averaged over all 100 trials and plotted in Figure~\ref{fig:figDiffTALA}.

For all parameter settings, the ordering holds for the multivariate means on average, i.e., the TA mean is always less than the LA mean, which is always less than the EP mean.
Note that, as the dispersion level increases, the difference between the three means also increases.
Increasing the dispersion level will ``stretch'' the observation log-likelihood term. This will scale down its derivative, and as a result, the zero-crossing point of the tangent line ($\mu_{TA}$) will move further from the zero-crossing of $f(\eta|y)$ (i.e., $\mu_{LA}$).  Similarly, ``stretching'' the observation log-likelihood also moves the mean further from the mode, thus increasing the difference between $\mu_{LA}$ and $\mu_{EP}$.
For the kernel bandwidth, increasing the bandwidth also stretches the prior term.
This scales down the derivatives of $f$, and as a result, the difference between TA and LA also increases in a similar way.
On the other hand, as the bandwidth increases, the difference between LA and EP {\em decreases}.  Increasing the bandwidth squashes the prior, making it more uniform.  As a result, the prior has less influence on the mean of the posterior, compared to the likelihood term, and the EP mean converges to the mean of the likelihood term.

\begin{figure}[tb]
 \centering
\begin{tabular}{c}
 \scriptsize (a) The first example: EP having the best performance \\
 \includegraphics[scale=0.31]{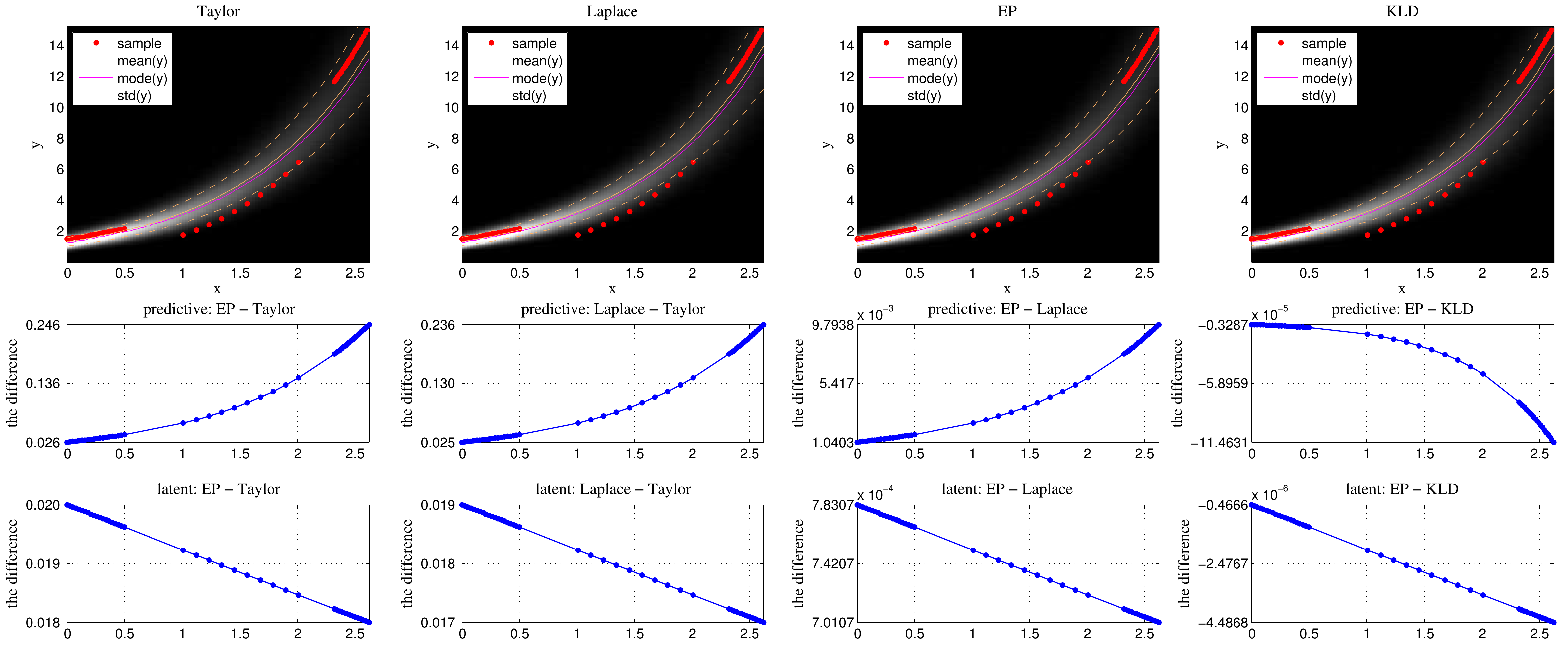} \\
 \scriptsize (b)  The second example: Taylor having the best performance \\
 \includegraphics[scale=0.31]{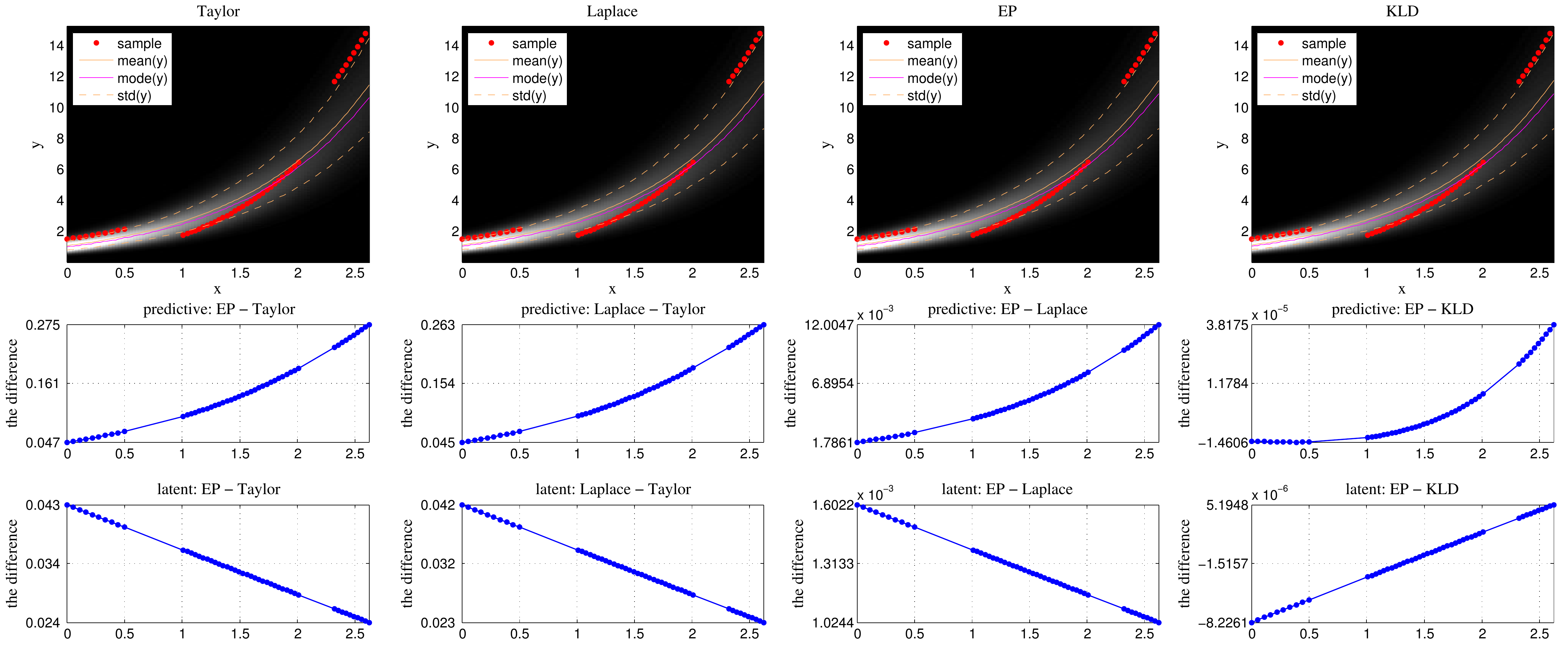} \\
\end{tabular}
\caption{Two 1D examples of comparing different inference methods.
In each example the top row shows the learned Gamma\shape-GGPM
regression models with four different inference methods:
Taylor, Laplace, EP, KLD. The middle row
shows the difference between predictive distributions, while the
bottom row shows the difference between latent functions.}
\label{fig:EP_Taylor_best}
\end{figure}

The ordering of the posterior means also suggest  that the predictive means will follow a similar order,
 i.e., the predictions using TA will be less than those of LA, and the predictions
using EP will always be larger than LA.
This is difficult to prove theoretically, since the the variance of the posterior affects the predictive mean,
but has been observed empirically in toy examples, as well as in experiments on real data in Section \ref{text:experiments}.
This is demonstrated in Figure~\ref{fig:EP_Taylor_best}, which uses a Gamma-shape GGPM with
the four inference methods. The training samples are distributed in
three distinct regions and have the similar trends to an exponential
function.
All the four inference methods can well capture the exponential trend.
However, given an input $x_*$, the latent values for TA, LA, and EP exhibit the ordering of Claim \ref{claim:postmeans}, as seen in the second row of Figure~\ref{fig:EP_Taylor_best}(a).  In addition, the predictive means also follow the ordering.
KLD and EP methods have
almost overlapped latent means.
EP and KLD both optimize the KL divergence between the true
posterior $p$ and the approximate $q$. The difference is that EP
optimizes $D(p||q)$ and KLD optimizes $D(q||p)$. For a unimodal
distribution $p$, minimizing either $D(q||p)$ or $D(p||q)$ will correctly capture
the mean of $p$.

\setlength{\myw}{0.23\linewidth}

\subsection{Effect on prediction error}

The systematic ordering between the approximate posterior and predictive means suggests that differences in prediction accuracy between approximation methods are influenced by the distribution of the test data.
If the test data has more points ``below the mean curve'', then TA will have better accuracy because it systematically underpredicts.  On the other hand, if there are more points ``above the mean curve'', then EP will have better accuracy.
This effect is illustrated in Figure~\ref{fig:EP_Taylor_best}b.
Because of the configuration of the data,
the predictive mean function passes above the middle points and below the extremal points.
Thus TA will have the smallest prediction error for the middle region and largest error at the extremes.  In contrast, EP will have lowest error on points near the ends, and largest error on the middle region.  If the test data contains more points in the middle, as in Figure~\ref{fig:EP_Taylor_best}b, then the Taylor approximation will have lower average predictive error.  In contrast, if the test data contains more points at the extremes, as in Figure~\ref{fig:EP_Taylor_best}a, then EP will have lower average error.

\begin{table}[!thbp]
\vspace{-0.01in} \centering \scriptsize
\begin{tabular}{@{}l||cc|cc||cc|cc@{}}
\multicolumn{1}{c||}{} & \multicolumn{4}{c||}{Gamma-shape} & \multicolumn{4}{c}{Gamma-scale}
\\
\cline{1-9}
 & \multicolumn{2}{c|}{First Example (a)}& \multicolumn{2}{c||}{Second Example (b)}
 & \multicolumn{2}{c|}{First Example (a)} & \multicolumn{2}{c}{Second Example (b)}\\
 Inference  & MAE & \NLP & MAE &
\NLP  & MAE & \NLP & MAE & \NLP
 \\ \hline
Taylor &$0.894$&$1.293$&$\bf 1.123$&$1.429$ & $\bf 0.581$&$1.274$&$0.900$&$1.309$\\
Laplace &$0.818$&$\bf 1.282$&$1.144$&$\bf 1.424$ &$0.633$&$\bf 1.243$&$0.715$&$\bf 1.241$\\
EP & $\bf 0.815$&$\bf 1.282$&$1.145$&$\bf 1.424$ &$0.636$&$\bf 1.243$&$\bf 0.713$&$\bf 1.241$\\
KLD & $\bf 0.815$&$\bf 1.282$&$1.145$&$\bf 1.424$ &$0.636$&$\bf 1.243$&$\bf 0.713$&$\bf 1.241$\\
 \hline
\end{tabular}\caption{Average errors for the two 1D examples.}
\label{tab:simExam}
\end{table}

To quantify this difference in predictive accuracy, a test set was generated by randomly sampling points in the neighborhood of each training point.
Each inference method was evaluated using two measures: 1) the mean absolute error (MAE), which measures the goodness-of-fit; 2)  the mean negative log predictive density evaluated
at the test points (\NLP), which measures how well the model predicts the entire density~\citep{Snelson04WGP}.
Evaluation results are presented in Table~\ref{tab:simExam}.
For the first example (Figure~\ref{fig:EP_Taylor_best}a), there are more test points in the end regions, and  EP achieves the lowest MAE of $0.815$ versus $0.894$ for TA.
On the other hand, in the second example (Figure~\ref{fig:EP_Taylor_best}b), where most of the test data is in the middle region, TA is the most accurate among the four methods (MAE of $1.123$ versus $1.145$ of EP).
Finally, if the likelihood is changed from Gamma-shape to Gamma-scale, the MAE results of the four inference methods are reversed for the two examples (see Table~\ref{tab:simExam} right), since the derivative of the log-likelihood of the Gamma-scale is concave.
If we evaluate the results by \NLP, the Taylor method always
performs worse than the other three methods.

From these examples and Claim \ref{claim:postmeans},  %
it can be concluded that the performances of different
inference methods are highly affected by dataset distributions,
likelihood functions and evaluation metrics.
The systematic ordering of the latent posterior means can affect the prediction accuracy if the test set is unbalanced or skewed.
Real-life datasets are often noisy and high-dimensional,
and it is unlikely for a single inference method to dominate for all datasets and metrics.
\section{Initializing hyperparameter estimation}\label{multiple_local_minima}

\begin{figure}[tbhp]
\centering
\begin{tabular}{r@{}c@{}c@{}l}
\raisebox{0.75in}{\footnotesize{Taylor}} &
\includegraphics[scale=0.45]{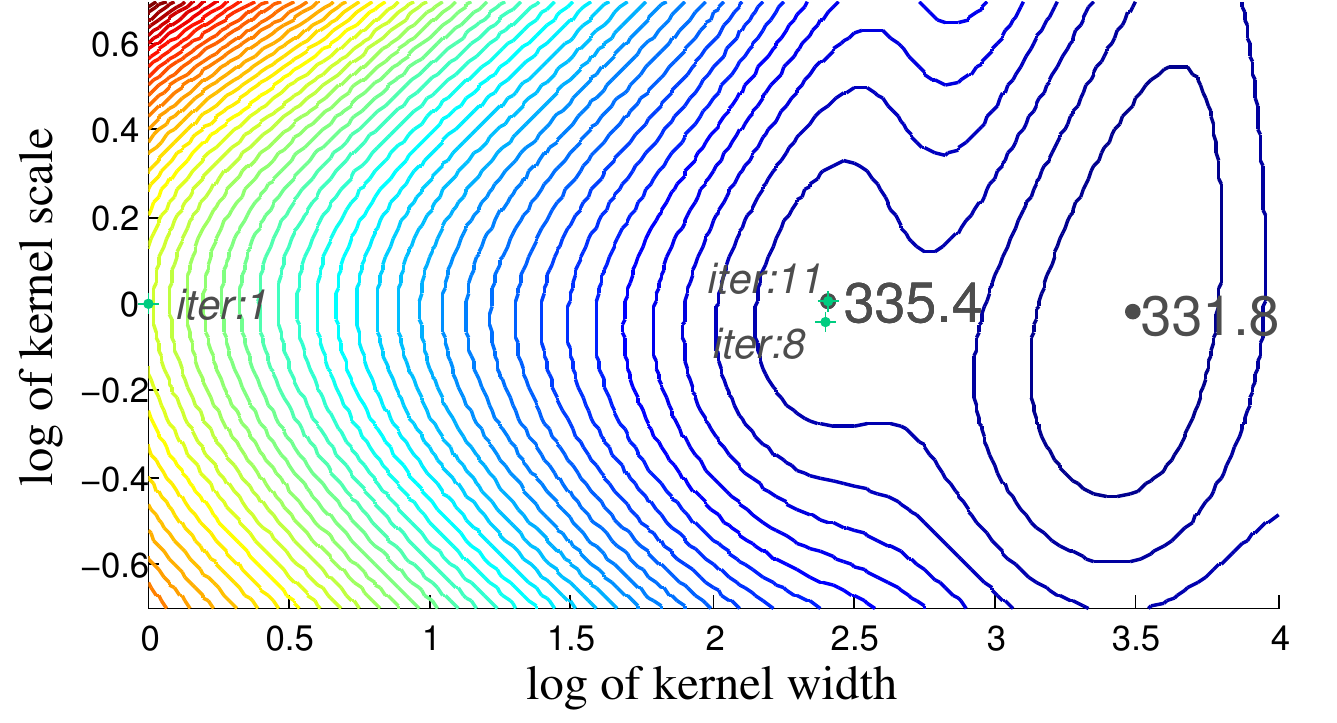}\vspace{-0.00in}&
 \includegraphics[scale=0.45]{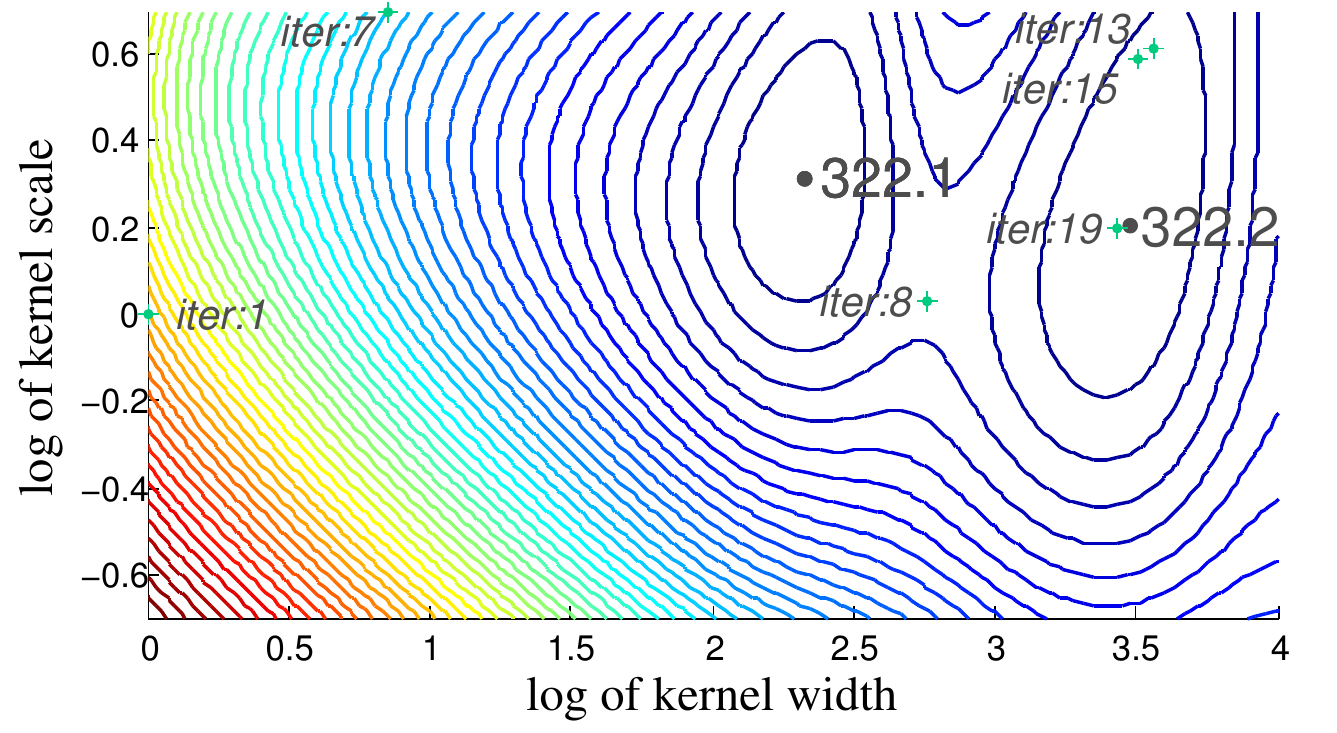}\vspace{-0.05in}
 & \raisebox{0.75in}{\footnotesize{Laplace}}
 \\
\raisebox{0.75in}{\footnotesize{EP}}
 &
\includegraphics[scale=0.45]{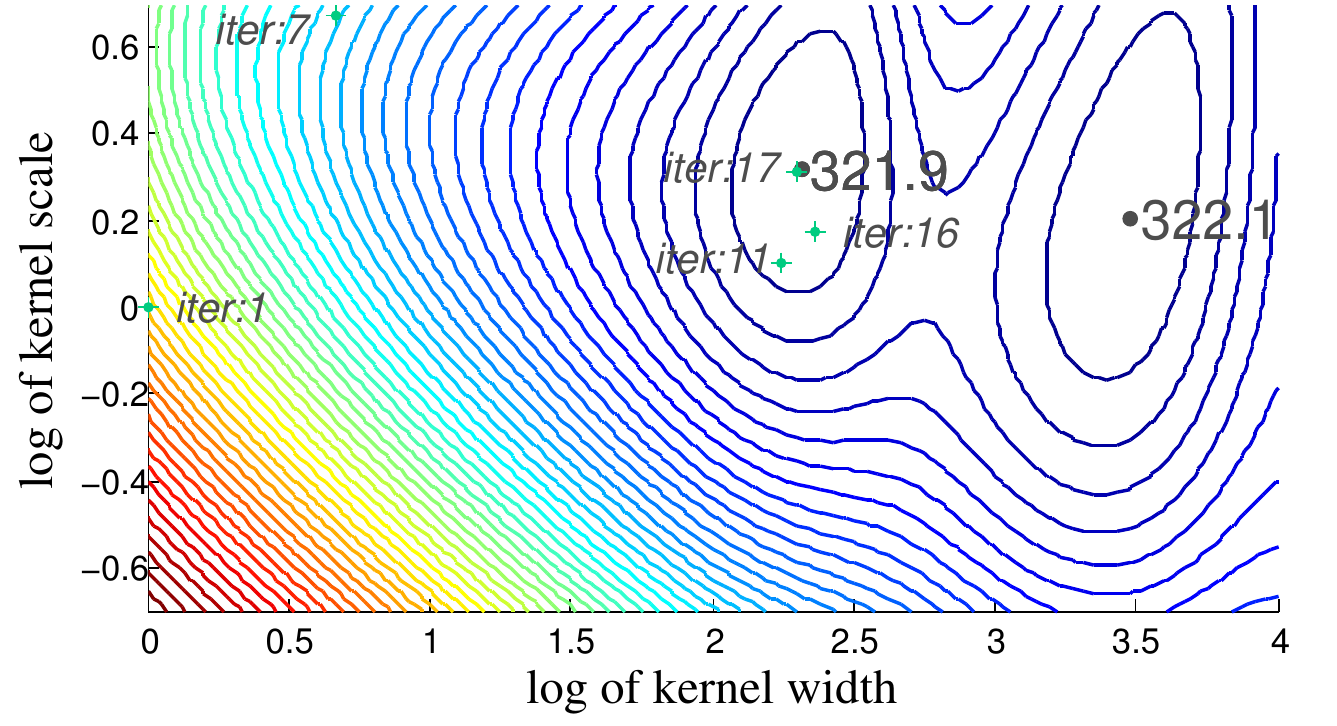}\vspace{-0.00in}&
\includegraphics[scale=0.45]{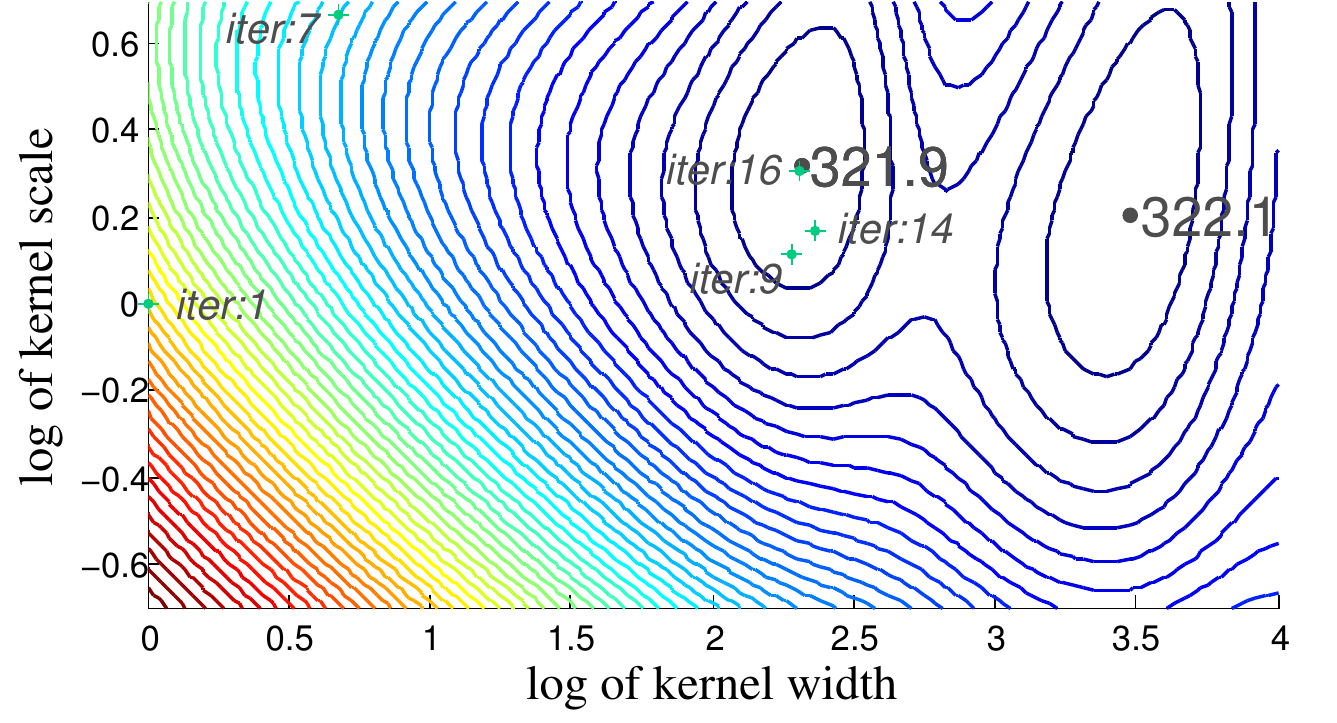}\vspace{-0.00in}
& \raisebox{0.75in}{\footnotesize{KLD}}
\end{tabular}
\caption{Contour plot of the negative log marginal likelihood
as a function of the RBF kernel width and scale hyperparameters for four approximate
inference methods.  The iterations of scaled conjugate gradient
minimization with initialization (0, 0) are plotted in green.}
\label{fig:rainfall_nlZ}
\end{figure}

As with GPR, the estimation of GGPM hyperparameters by maximizing the marginal likelihood often suffers from the problem of multiple local optima.  %
Figure \ref{fig:rainfall_nlZ} shows the negative log-marginal
likelihood (NML), as a function of the RBF kernel width and scale hyperparameters,
for the four inference methods
on the rainfall dataset (see Section \ref{text:binexp} for description).
The NML surface for all four likelihood functions have
at least two local optima, and
the Laplace, EP and KLD methods produce very similar surfaces.

A common approach to hyperparameter estimation is to initialize the log-hyperparameters to zero, and then optimize using the scaled conjugate gradient method \citep{GPMLcode}.
However, in the example in Figure \ref{fig:rainfall_nlZ}, using the same initialization strategy leads to {\em different} hyperparameter estimates from each inference method.
As illustrated in Figure \ref{fig:rainfall_nlZ},
TA, KLD, and EP converge to similar local optimum on the left, whereas Laplace converges to a different local minimum on the right.
Hence, a more robust initialization method is required to better explore the search space, and to ensure that a good optimum can be found for each inference method.
One strategy is to run the optimization procedure many times using a large set  of random initializations.  However, for some inference algorithms, e.g. EP, the computational burden will be large.

\setlength{\myw}{0.23\linewidth}

\begin{figure}[htbp]
\vspace{-0.1in} \centering
\begin{tabular}{ccc}
{\footnotesize (a)} &
{\footnotesize (b)} &
{\footnotesize (c)} \\
 \includegraphics[scale=0.4]{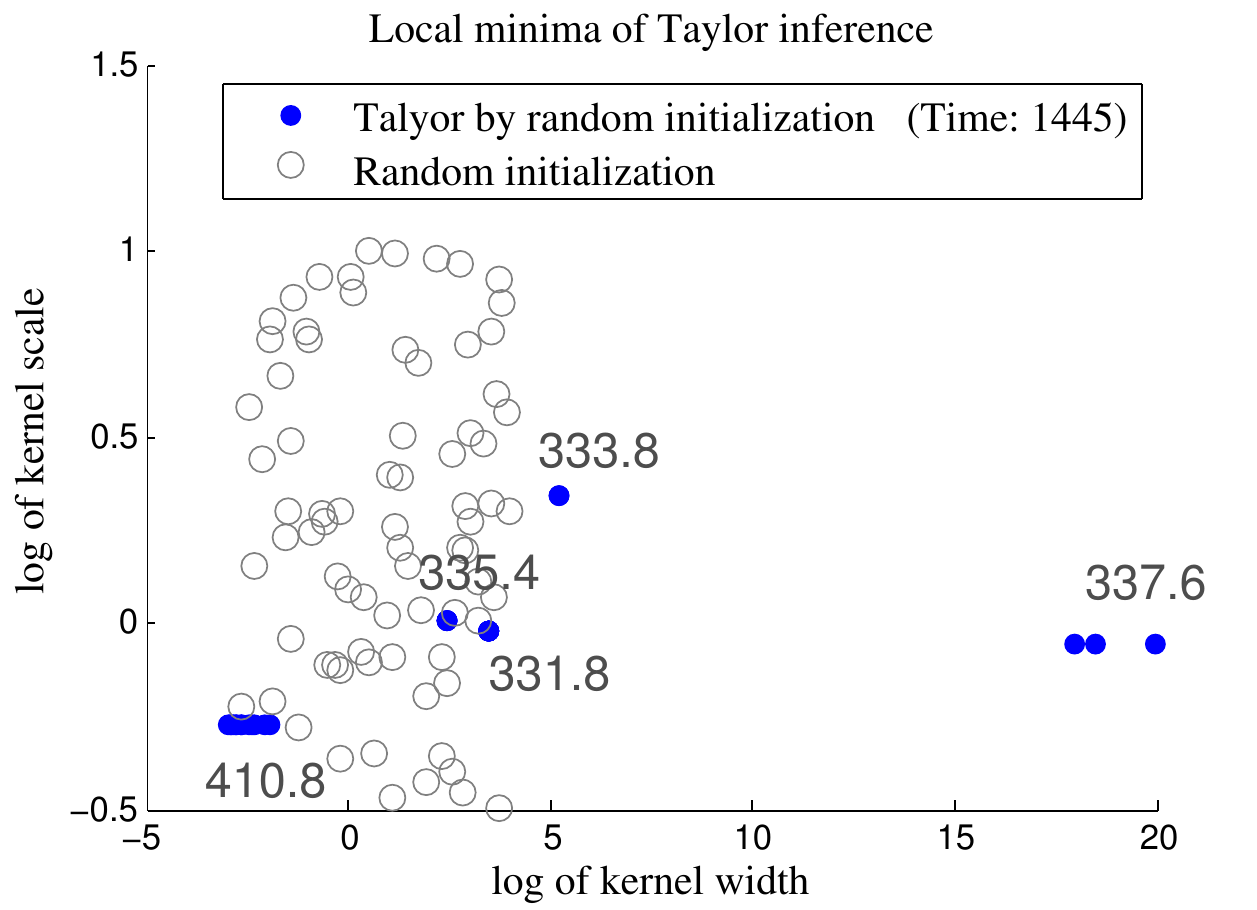}
 & \hspace{-0.2in}
 \includegraphics[scale=0.4]{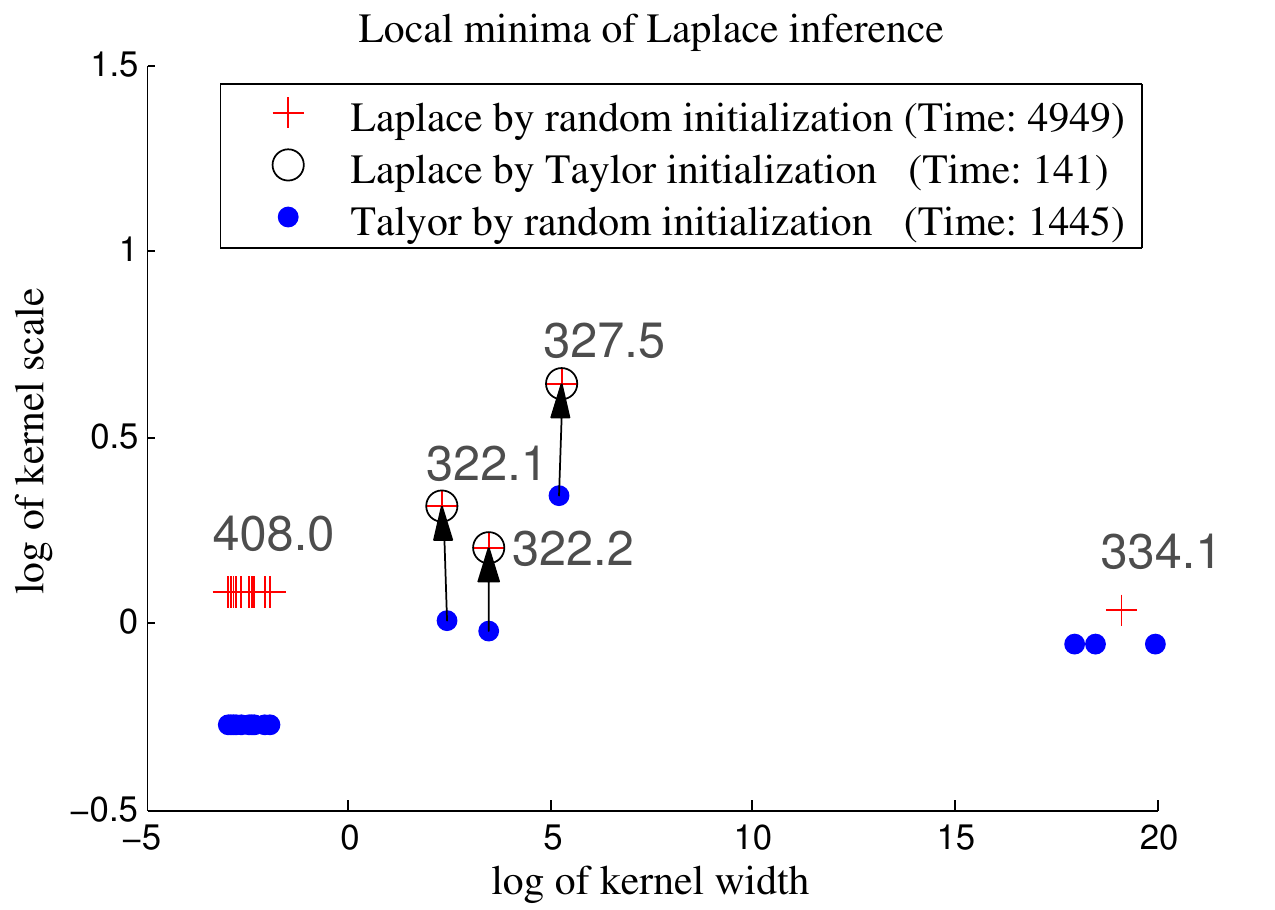}
& \hspace{-0.2in}
 \includegraphics[scale=0.4]{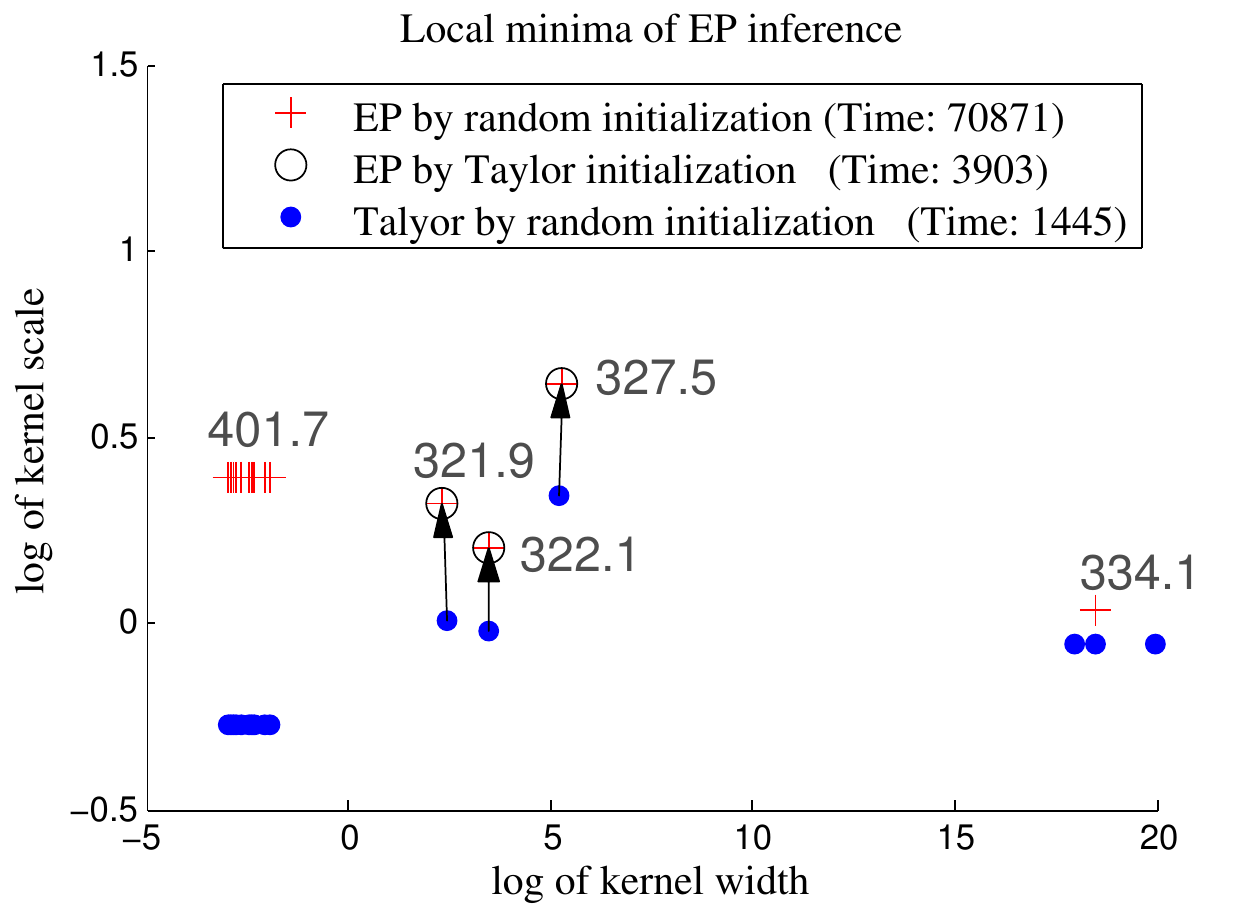} \vspace{-0.1in}
\end{tabular}%
\caption{%
Illustration of using Taylor inference to
initialize other inference methods for hyperparameter estimation.
(a) Candidate local optima are found using Taylor method with 50 random initializations;
(b) Comparison of local optima of Laplace method initialized
by Taylor and 50 random points. (c) Comparison of local optima of EP
method initialized by Taylor and 50 random points.}
\label{fig:figEffTay}
\end{figure}

We propose an efficient initialization strategy that uses Taylor inference to quickly find a few good initializations for the other inference methods (Laplace, KLD, and EP).
The procedure is illustrated in Figure~\ref{fig:figEffTay}.
Taylor inference is used to optimize the hyperparameters using
50 random initializations (see Figure~\ref{fig:figEffTay}a),
resulting in convergence to several local optima  with different marginal likelihoods.
The top 3 unique local optima are used as the initializations for the Laplace and EP methods, and the results are presented in Figures~\ref{fig:figEffTay}b and \ref{fig:figEffTay}c.
In both cases, the Taylor-initialized Laplace and EP can recover the same local optima as the randomly-initialized versions, but with a significant reduction in computational cost (3 times faster for Laplace, and 13 times faster for EP).
The hyperparameter resulting in the largest marginal likelihood can then be selected as the estimate.

\setlength{\myw}{0.23\linewidth}

Table~\ref{tab:comIni1} shows the quantitative comparison between
random-initialization and Taylor-initialization on three dataset (servo, auto-mpg and housing from Section \ref{text:nnrexp}).
We compare the two initialization methods using the relative
change in MAE, $\Delta\mathrm{MAE} =
\tfrac{\mathrm{MAE}_{T}-\mathrm{MAE}_{R}}{\mathrm{MAE}_R}$,
where $\mathrm{MAE}_T$ and $\mathrm{MAE}_R$
are the MAEs when using Taylor-initialization and
random-initialization, respectively.  The relative change in \NLP~is calculated in an analogous way.
In all cases, the relative changes in MAE are small (within 0.002),
and not statistically significant (paired $t$-test, $p > 0.15$).
A similar conclusion holds for the relative change in \NLP.
The Taylor-initialization yields a significant reduction in
computational cost.  For example, Taylor-initialized EP was about 26
times faster than random-initialized EP.
Furthermore, with the help of Taylor initialization, we did not encounter any convergence problems with EP
(similarly observed for the real experiments in Section \ref{text:experiments}).
These experiments demonstrate that Taylor-initialization can speedup hyperparameter estimation for other inference methods, while maintaining the same quality as fully random initialization.

\begin{table}[h] \centering{
\scriptsize \vspace{-0.1in}
  \begin{tabular}{@{}l|c@{\hspace{0.1in}}c@{\hspace{0.1in}}c||c@{\hspace{0.1in}}c@{\hspace{0.1in}}c@{}}
\cline{2-7}
 \multicolumn{1}{c}{} & \multicolumn{3}{c||}{Laplace} & \multicolumn{3}{c}{EP}\\
\cline{1-7} Likelihood & $\Delta$MAE  & $\Delta$\NLP & speedup &
$\Delta$MAE & $\Delta$\NLP & speedup
 \\\hline
\multirow{2}{*}{Gamma\shape}&$0.000\pm0.0001$&$0.000\pm0.0000$
&\multirow{2}{*}{$2.97\pm0.901$} & $0.000\pm0.0001$&$0.000\pm0.0023$
&\multirow{2}{*}{$24.92\pm7.566$}\\
 & ( 0.9987) & ( 0.4531) & & ( 0.1889) & ( 0.4090) & \\ \hline

\multirow{2}{*}{Gamma\scale}&$-0.002\pm0.0127$&$0.002\pm0.0077$
&\multirow{2}{*}{$3.17\pm1.109$}
&$-0.001\pm0.0026$&$-0.001\pm0.0019$
&\multirow{2}{*}{$26.86\pm6.487$}\\
 & ( 0.3474) & ( 0.2597) & & ( 0.2056) & ( 0.1270) &  \\ \hline

\multirow{2}{*}{Inv.Gauss}&$-0.000\pm0.0012$&$0.001\pm0.0023$
&\multirow{2}{*}{$3.44\pm1.590$} &$0.002\pm0.0164$&$0.013\pm0.0534$
&\multirow{2}{*}{$28.78\pm9.861$}\\
 & ( 0.3296) & ( 0.2709) & & ( 0.5010) & ( 0.2103)  & \\ \hline

\multirow{2}{*}{all      }&$-0.001\pm0.0074$&$0.001\pm0.0046$
&\multirow{2}{*}{$3.20\pm1.236$} &$0.000\pm0.0095$&$0.004\pm0.0312$
&\multirow{2}{*}{$26.85\pm8.158$}\\
 & ( 0.3005) & ( 0.1567) &&  ( 0.6372) & ( 0.2158) & \\ \hline
\end{tabular}
}\caption{Comparison between random-initialization and
Taylor-initialization in terms of relative change in MAE and \NLP, and the speedup factor.
Parenthesis denote the $p$ values using a paired $t$-test. The differences are not statistically significant.} \label{tab:comIni1}
\end{table}

\section{Experiments}
\label{text:experiments}

In this section, we present experiments using GGPMs
and inference methods on a wide variety of real-world datasets.
In Section \ref{text:binexp}, we consider finite counting data and the Binomial-GGPM.
In Section \ref{text:nnrexp}, we experiment with regression to non-negative reals (Gamma-GGPM, Inverse-Gaussian GGPM).
Finally, Section \ref{text:betaexp} presents results on range data (Beta-GGPM), and Section \ref{text:countexp} considers counting data and Poisson-GGPMs\footnote{In this paper, we do not present results using GP regression and classification, which have been extensively studied in \cite{GPML,Kuss05GPC,Nickisch2008GPC}.}.

In the following experiments, we use the Taylor initialization method
from Section \ref{multiple_local_minima} to speed up hyperparameter estimation
using the other inference methods (LA, EP, KLD).
For all inference methods, the best hyperparameter is selected as the
one with the largest marginal likelihood on the training set.
Our results indicate that the performances of different inference methods are highly affected by
likelihoods, evaluation metrics and datesets. Each inference method,
even the Taylor method, can have the best performance. Furthermore,
our hypothesis testing results show that the difference between EP
and KLD is not statistically significant;
Laplace approximation and EP often perform comparably.

\comments{
\setlength{\myw}{0.22\linewidth}
\begin{figure}[tbp]
\vspace{-0.2in} \centering
\begin{tabular}{@{\hspace{-0.1in}}c}
\\
\includegraphics[scale =
0.62]{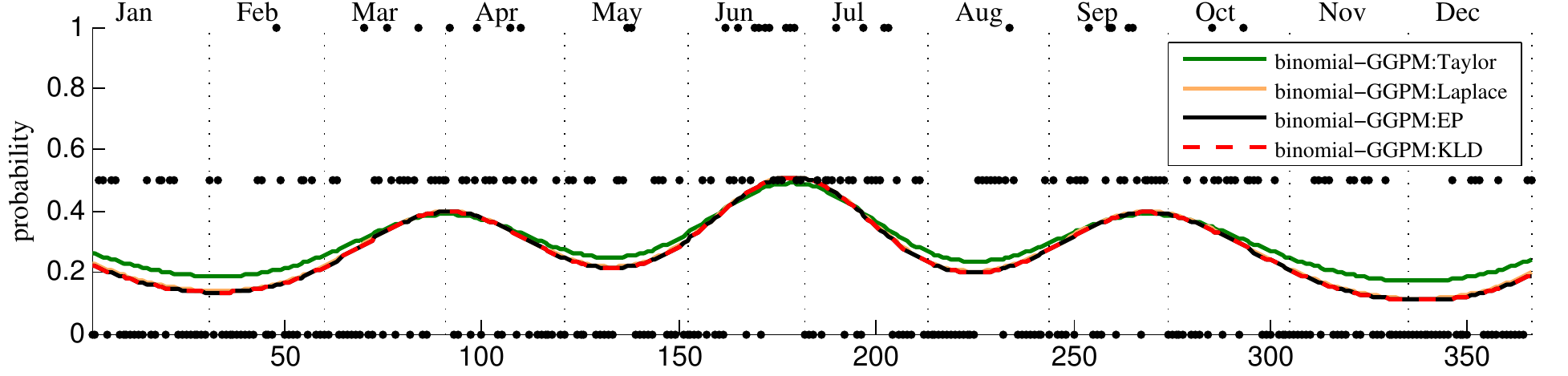}\vspace{-0.11in}\\ \scriptsize(a)
\\
\includegraphics[scale = 0.62]{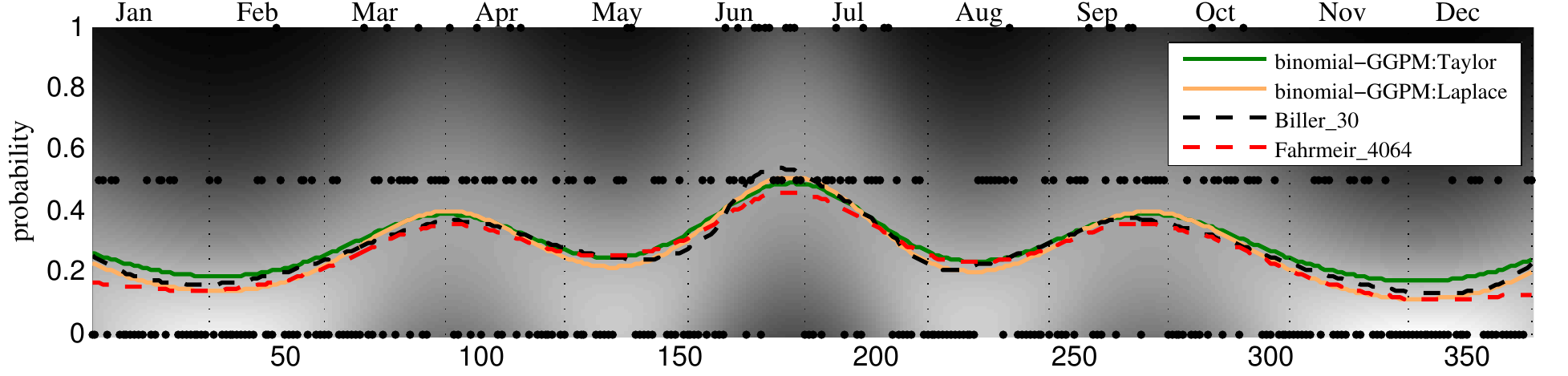}
\vspace{-0.11in}\\ \scriptsize(b)\\
\end{tabular}
\caption{Estimated smoother binomial mean function of rainfall in
Tokyo and comparison with other methods. (a) Estimated binomial mean
function by GGPMs. (b) Comparing GGPMs to Biller's and Fahrmeir's
methods. } \label{fig:rainfall_smooth} \vspace{-0.2in}
\end{figure}

\setlength{\myw}{0.22\linewidth}
\begin{figure}[!htbp]
\vspace{-0.2in} \centering
\begin{tabular}{@{\hspace{-0.1in}}c}
\\
\includegraphics[scale =
0.62]{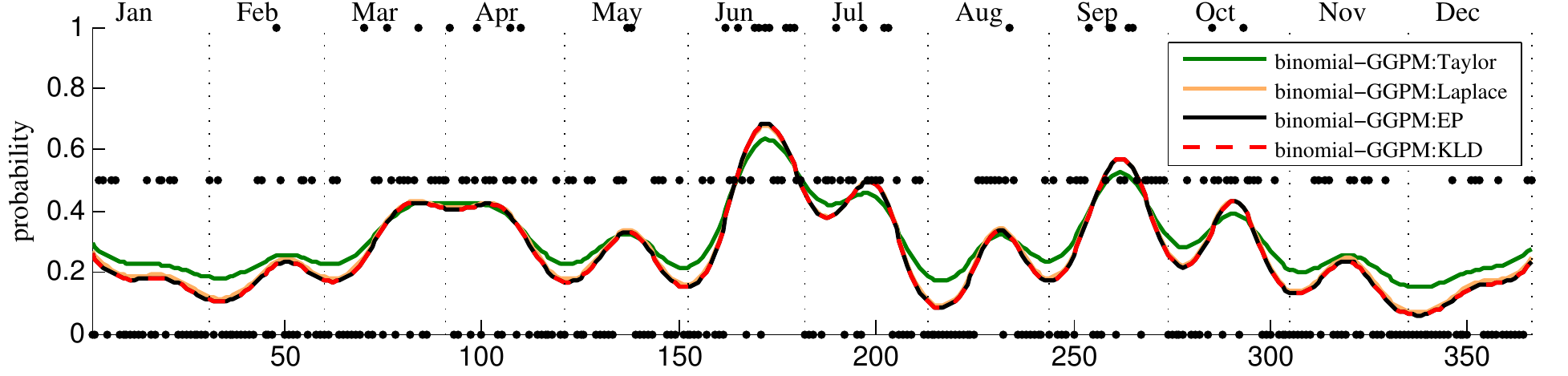}\vspace{-0.1in}\\ \scriptsize(a)
\\
\includegraphics[scale = 0.62]{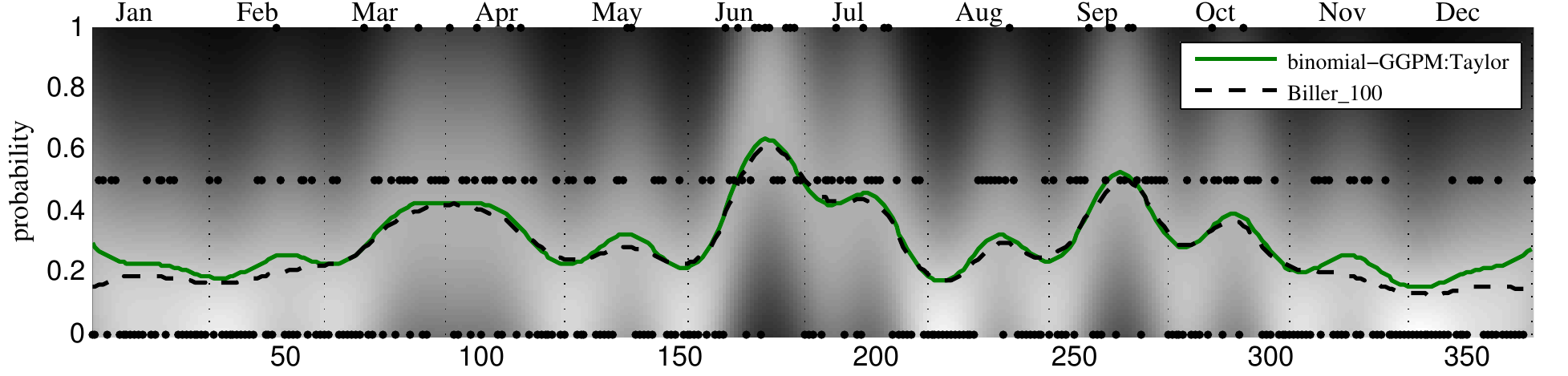}
\vspace{-0.1in}\\ \scriptsize(b)
\\
\includegraphics[scale = 0.62]{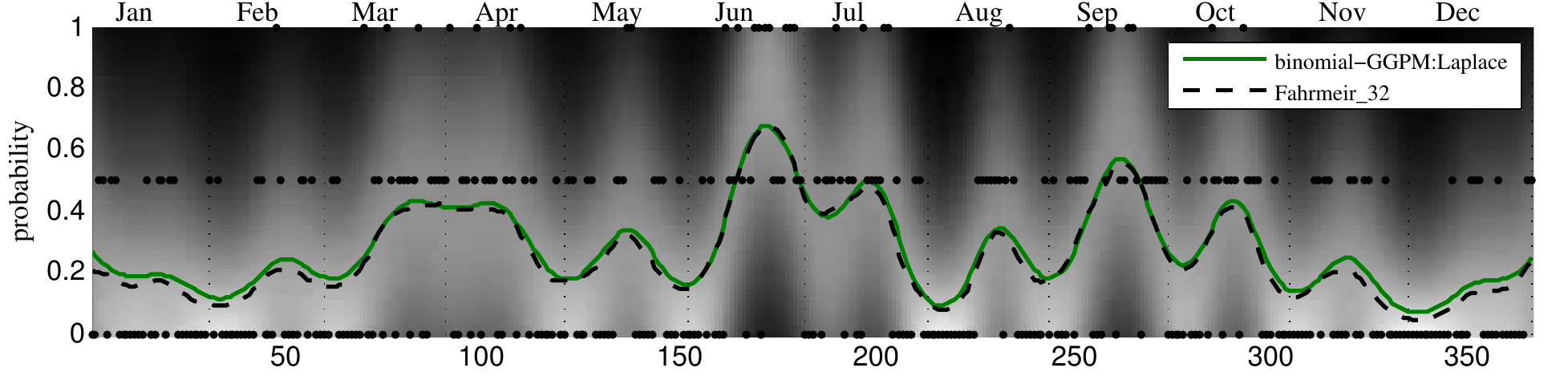}\vspace{-0.1in}\\ \scriptsize(c)
\\
\end{tabular}
\caption{Estimated rougher binomial mean function of rainfall in
Tokyo and comparison with other methods. (a) Estimated binomial mean
function by GGPMs. (b) Comparing GGPMs with Taylor inference to
Biller's method. (c) Comparing GGPMs with Laplace inference to
Fahrmeir's method.} \label{fig:rainfall_rough}
\end{figure}
}

\setlength{\myw}{0.22\linewidth}
\begin{figure}[tbp]
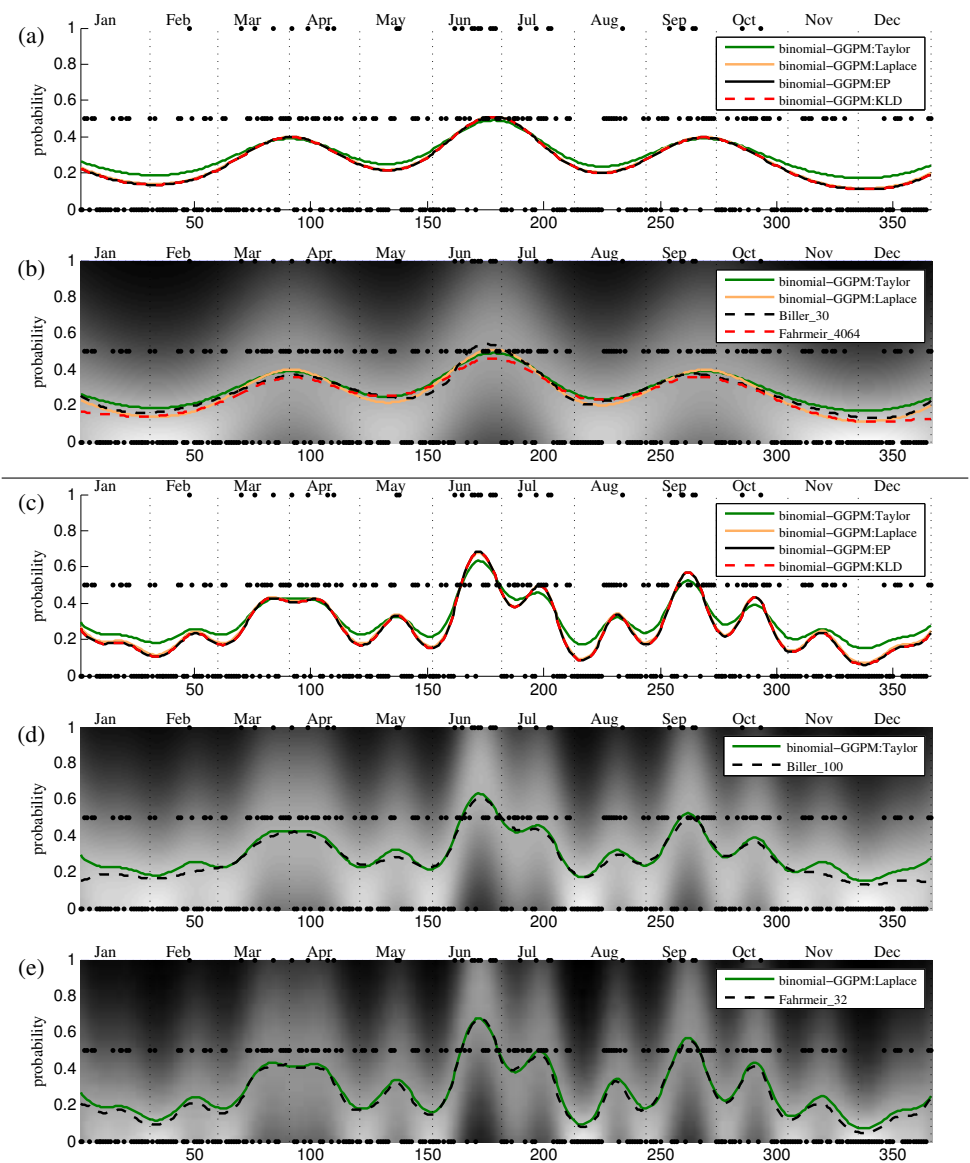

\vspace{-0.2in} \centering
\begin{tabular}{c@{\hspace{-0.1in}}c}
\raisebox{1in}{\footnotesize (a)} &
\includegraphics[scale = 0.62]{images/figSmoother-eps-converted-to.pdf}
\\
\raisebox{1in}{\footnotesize (b)} &
\includegraphics[scale = 0.62]{images/binoGGPM_RainFall_Laplace-eps-converted-to.pdf}
\\
\hline
\raisebox{1in}{\footnotesize (c)} &
\includegraphics[scale =0.62]{images/figRough-eps-converted-to.pdf}
\\
\raisebox{1in}{\footnotesize (d)} &
\includegraphics[scale = 0.62]{images/binoGGPM_RainFall_Taylor-eps-converted-to.pdf}
\\
\raisebox{1in}{\footnotesize (e)} &
\includegraphics[scale = 0.62]{images/binoGGPM_RainFall_EP-eps-converted-to.pdf}
\end{tabular}
\caption{Two  binomial-GGPM mean functions on Tokyo rainfall data, and comparisons with other methods:
(a) binomial-GGPM mean function using large kernel width (smoother function), and
(b) comparison to \cite{biller2000adaptive} and \cite{fahrmeir1994multivariate};
(c) binomial-GGPM using small kernel width (rougher function), with
(d) comparison of Taylor inference to \cite{biller2000adaptive} and
(e) comparison of Laplace inference to \cite{fahrmeir1994multivariate}.
} \label{fig:rainfall_both} \vspace{-0.2in}
\end{figure}

\subsection{Binomial Example}
\label{text:binexp}
In this section we apply binomial-GGPM to the Tokyo rainfall
dataset\footnote{http://www.stat.uni-muenchen.de/service/datenarchiv/tokio/tokio\_e.html}.
The dataset records the number of occurrences of rainfall over
1mm in Tokyo for every calendar day in 1983 and 1984. The rainfall
occurrence for a given calendar day follow a binomial distribution. We
assign the occurrence times ``0'', ``1'', and ``2'' to the outputs
$y\in\{0, 0.5, 1\}$ of the binomial model.
The input feature is the calendar day (0 to 365), and the RBF kernel was used\footnote{Since the input feature is the calendar day, a cyclic kernel which wraps around from 365 to 0 can be used to better model the correlation between days. In this paper, we do not adopt cyclic kernel to follow the same settings as the reference methods.}.

\begin{figure}[tp]
\vspace{-0.2in} \centering
\begin{tabular}{ccc}
 \includegraphics[scale=0.42]{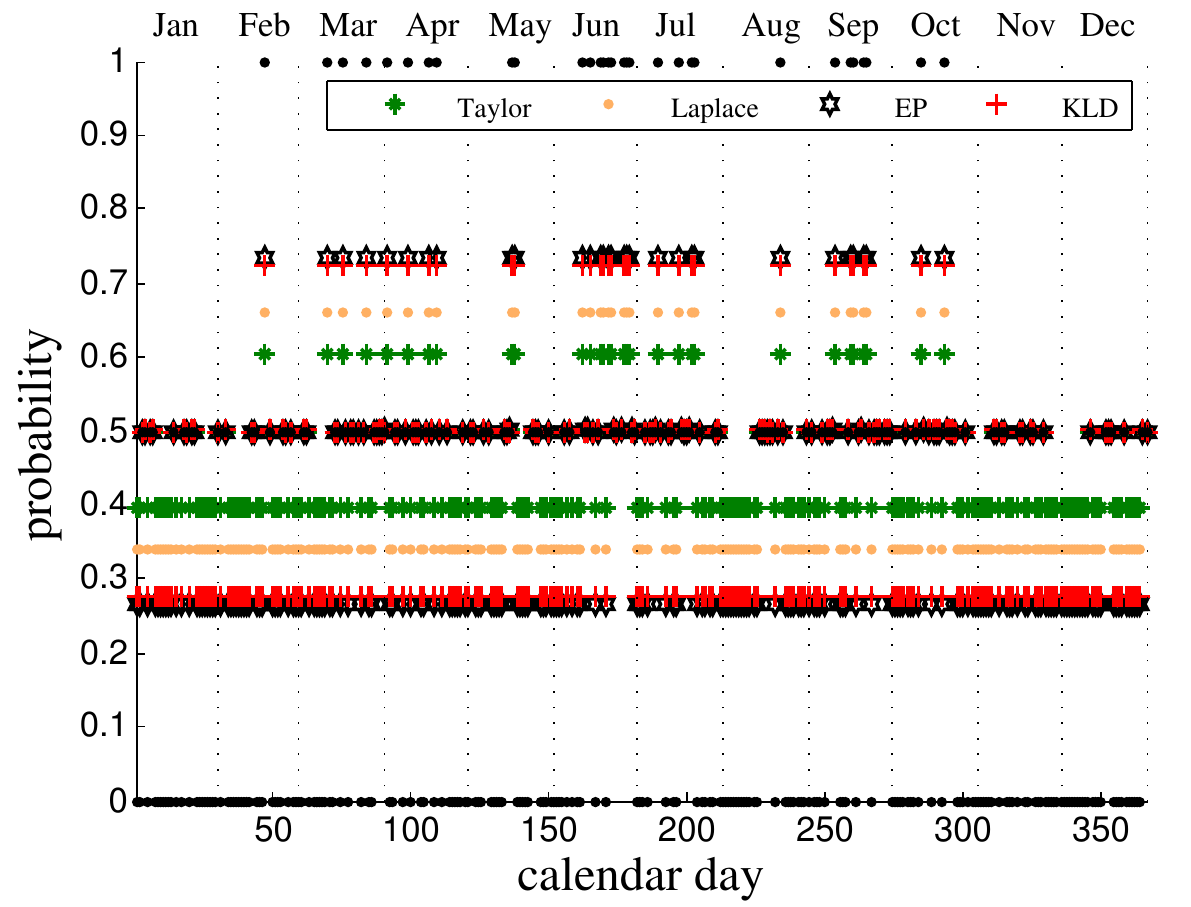}
 & \hspace{-0.2in}
 \includegraphics[scale=0.42]{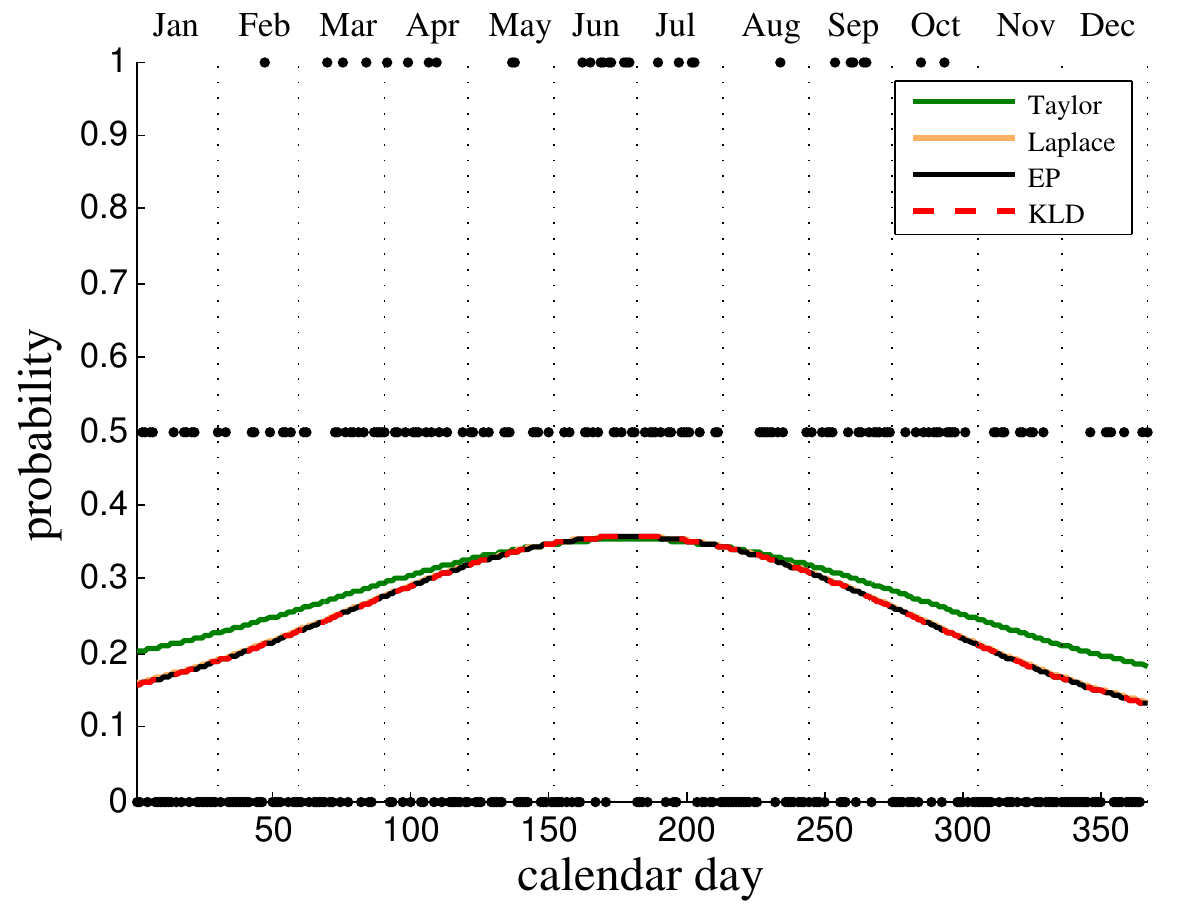}
& \hspace{-0.2in}
 \includegraphics[scale=0.42]{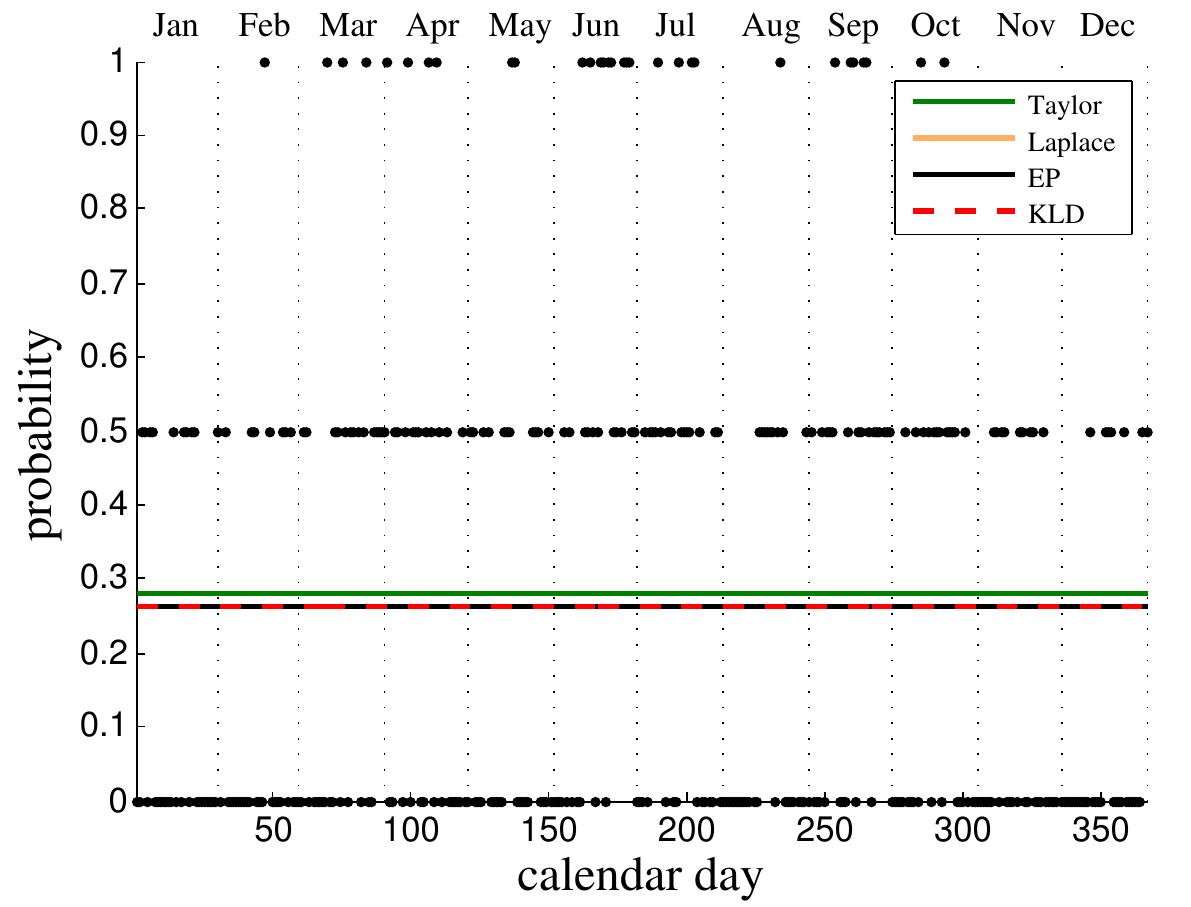}\vspace{-0.1in}
\\ \scriptsize(a) & \scriptsize(b) & \scriptsize(c)\\
\end{tabular}\vspace{-0.1in}
\caption{Estimated binomial-GGPM mean functions on Tokyo rainfall,
corresponding to three bad local optima (too big or small kernel
widths) as shown in Figure~\ref{fig:figEffTay}. }
\label{fig:figRF_other}
\end{figure}

As presented earlier, Figure \ref{fig:rainfall_nlZ} shows the negative log marginal
likelihood (NML) as a function of the RBF kernel width and scale,
and Figure \ref{fig:figEffTay} shows the results of Taylor-initialization, which yielded 5 local minima in the NML that correspond to different interpretations of the data.
The best two interpretations (largest marginal likelihoods) are presented in
Figures \ref{fig:rainfall_both}a and \ref{fig:rainfall_both}c, using the four inference methods.
Figure \ref{fig:rainfall_both}a uses a larger
kernel width, resulting in a smoother function that
shows the clear seasonal pattern in Tokyo,
as described by \citet{kitagawa1987non}: dry winter (Dec., Jan., and Feb.),
rainy season in late June to mid-July, stable hot
summer in late July through Aug., and generally fine but with an
occasional typhoon in Sept. and Oct.
It would be difficult to identify these trends by only looking at the original data.
Finally, Figure~\ref{fig:figRF_other} depicts the curves for the remaining three bad local minima, where
the kernel bandwidth is either too small or too large.
The Laplace, EP and KLD methods
have almost overlapped estimates for all the five local optima.
The Taylor method also captures similar trends as the
other three methods.

We compare the binomial-GGPMs with two spline-based regression
models. The first model~\citep{fahrmeir1994multivariate} is an
extension of GLM that replaces the linear function with a cubic spline. In
this model a parameter $\lambda_S$ is used to control the tradeoff
between data-fit and smoothness of the cubic function.
In \cite{fahrmeir1994multivariate}, $\lambda_S$ was estimated by cross-validation, resulting in two
local minima, $\lambda_S = 4064$ and $\lambda_S =32$. The larger $\lambda_S = 4064$ yields a relatively smoother curve, which is very close to the smoother estimate using binomial-GGPM (see Figure
\ref{fig:rainfall_both}b). The curve with $\lambda_S=32$ is
quite similar to the rougher estimate of binomial-GGPM using Laplace inference.
The difference of the two local optima can be attributed to different
smoothness levels.

\setlength{\myw}{0.23\linewidth}

The second model~\citep{biller2000adaptive} is a fully Bayesian approach to
regression splines with automatic knot placement. A Poisson
distribution with parameter $\lambda_N$ is placed over the number of
knots.
Experimental results~\citep{biller2000adaptive}  show that the rainfall dataset is very
sensitive to the choice of $\lambda_N$. With a smaller $\lambda_N$
the resulting function is very smooth, while increasing the value of
$\lambda_N$ allows a more flexible function. The estimate with
$\lambda_N = 100$ is similar to the rougher curve estimated by
Taylor %
(see Figure \ref{fig:rainfall_both}d).
When the value is decreased to $\lambda_N=30$, the curve becomes %
similar to the smoother Binomial-GGPM estimate (see Figure \ref{fig:rainfall_both}b).
From
the two comparisons, we can see the shapes of resulted curves are
highly affected by the adoption of optimal hyperparameters. If we
use the negative marginal likelihood as a metric, the smoother curve
will be favored by the Taylor method, while the rougher curve will
be selected by the other three inference methods.

\begin{table}[tbhp]
\centering {\footnotesize
\begin{tabular}{@{}l|ccc|cc@{}}
\hline
 Dataset Name & $X_{dim}$ & $Y_{min}$ & $Y_{max}$ & $N_{train}$ & $N_{test}$
 \\\hline
  servo & 4 & 0.13 & 7.10 & 70 & 97 \\
  auto-mpg & 7 & 9.00 & 46.60 & 100 & 298 \\
 housing & 12 & 5.00 & 50.00 & 200 & 306 \\
 abalone & 8 & 1.00 & 29.00 & 1000 & 3177 \\
\hline
\end{tabular}
} \caption{Datasets for non-negative real regression.}
\label{tab:nnrnInfo}
\end{table}

\begin{table}[h] \centering{
\scriptsize \vspace{-0.1in}
  \begin{tabular}{@{}l@{}c|ccc||ccc@{}} %
\cline{3-8}
 \multicolumn{2}{c}{} & \multicolumn{3}{c||}{abalone dataset} &\multicolumn{3}{c}{housing dataset}\\
\cline{1-8} Lik$^{*}$ & Inf. & MAE & MSE & \NLP & MAE & MSE
& \NLP
 \\\hline
GP&Exact& $1.60\pm 0.029$ & $4.66\pm 0.232$ & $2.19\pm 0.012$&$2.40\pm0.136$&$11.27\pm1.716$&$2.60\pm0.047$\\
LGP&Exact&$1.47\pm0.016$&$4.12\pm0.125$&$2.00\pm0.009$&$2.22\pm0.141$&$10.39\pm1.858$&$2.57\pm0.040$\\
WGP&Exact&$1.52\pm0.024$&$4.36\pm0.169$&$\bf 1.94\pm0.010$&$2.22\pm0.138$&$\bf 10.24\pm1.386$&$2.54\pm0.191$\\
  \hline
GA\shape&Taylor&$1.49\pm0.021$&$4.26\pm0.238$&$1.99\pm0.007$&$\bf 2.21\pm0.140$&$10.46\pm1.841$&$2.59\pm0.044$\\
GA\shape&Laplace&$1.55\pm0.022$&$4.55\pm0.232$&$2.01\pm0.007$&$2.21\pm0.144$&$10.41\pm1.885$&$2.58\pm0.041$\\
GA\shape&EP&$1.55\pm0.022$&$4.57\pm0.237$&$2.01\pm0.007$&$2.22\pm0.146$&$10.46\pm1.897$&$2.58\pm0.040$\\
GA\shape&KLD&$1.55\pm0.022$&$4.53\pm0.185$&$2.01\pm0.007$&$2.22\pm0.146$&$10.46\pm1.891$&$2.58\pm0.040$\\
\hline
GA\scale&Taylor&$1.67\pm0.023$&$5.10\pm0.173$&$2.12\pm0.012$&$2.23\pm0.159$&$10.27\pm2.282$&$2.52\pm0.038$\\
GA\scale&Laplace&$1.53\pm0.024$&$4.41\pm0.268$&$2.06\pm0.010$&$2.23\pm0.146$&$10.35\pm2.104$&$\bf 2.52\pm0.036$\\
GA\scale&EP&$1.53\pm0.025$&$4.39\pm0.274$&$2.06\pm0.010$&$2.22\pm0.141$&$10.30\pm2.046$&$2.52\pm0.037$\\
GA\scale&KLD&$1.53\pm0.025$&$4.35\pm0.300$&$2.05\pm0.010$&$2.22\pm0.142$&$10.29\pm2.071$&$2.52\pm0.038$\\
\hline
INV&Taylor&$\bf 1.42\pm0.022$&$\bf 4.11\pm0.354$&$1.99\pm0.014$&$2.27\pm0.156$&$11.26\pm2.080$&$2.75\pm0.069$\\
INV&Laplace&$1.54\pm0.024$&$4.73\pm0.440$&$1.98\pm0.008$&$2.28\pm0.163$&$11.05\pm2.073$&$2.72\pm0.064$\\
INV&EP&$1.55\pm0.025$&$4.78\pm0.449$&$1.99\pm0.008$&$2.32\pm0.171$&$11.39\pm2.172$&$2.71\pm0.059$\\
INV&KLD&$1.56\pm0.019$&$5.04\pm0.576$&$1.99\pm0.007$&$2.32\pm0.170$&$11.36\pm2.149$&$2.71\pm0.059$\\
\hline
\end{tabular}
\begin{tabular}{@{}l@{}c|ccc||ccc@{}}
\cline{3-8}
 \multicolumn{2}{c}{} & \multicolumn{3}{c||}{auto-mpg dataset} &\multicolumn{3}{c}{servo dataset}\\
\cline{1-8}Lik$^{*}$ & Inf. & MAE & MSE & \NLP & MAE &
MSE&\NLP
 \\\hline
GP&Exact&$2.11\pm0.053$&$8.69\pm0.349$&$2.48\pm0.038$&$0.43\pm0.047$&$0.33\pm0.058$&$1.06\pm0.033$\\
LGP&Exact&$2.10\pm0.090$&$8.71\pm0.655$&$2.36\pm0.044$&$0.27\pm0.035$&$ 0.18\pm0.042$&$0.32\pm0.044$\\
WGP&Exact&$\bf 2.08\pm0.061$&$\bf 8.45\pm0.672$&$2.39\pm0.055$&$\bf 0.25\pm0.031$&$0.18\pm0.043$&$0.13\pm0.113$\\
\hline
GA\shape&Taylor&$2.09\pm0.072$&$8.71\pm0.540$&$2.36\pm0.043$&$0.26\pm0.035$&$0.20\pm0.048$&$0.11\pm0.084$\\
GA\shape&Laplace&$2.09\pm0.073$&$8.64\pm0.506$&$\bf 2.36\pm0.040$&$0.27\pm0.037$&$0.20\pm0.047$&$0.09\pm0.072$\\
GA\shape&EP&$2.09\pm0.074$&$8.65\pm0.503$&$2.36\pm0.040$&$0.27\pm0.037$&$0.20\pm0.048$&$\bf 0.09\pm0.070$\\
GA\shape&KLD&$2.09\pm0.075$&$8.66\pm0.498$&$2.36\pm0.041$&$0.27\pm0.037$&$0.20\pm0.048$&$0.09\pm0.071$\\
\hline
GA\scale&Taylor&$2.10\pm0.076$&$8.73\pm0.536$&$2.40\pm0.036$&$0.28\pm0.024$&$0.21\pm0.041$&$0.19\pm0.040$\\
GA\scale&Laplace&$2.09\pm0.064$&$8.74\pm0.496$&$2.40\pm0.035$&$0.26\pm0.023$&$0.19\pm0.033$&$0.18\pm0.043$\\
GA\scale&EP&$2.09\pm0.064$&$8.75\pm0.503$&$2.40\pm0.036$&$\bf 0.25\pm0.023$&$\bf 0.18\pm0.036$&$0.17\pm0.048$\\
GA\scale&KLD&$2.09\pm0.065$&$8.74\pm0.495$&$2.40\pm0.036$&$0.25\pm0.024$&$0.18\pm0.036$&$0.17\pm0.048$\\
\hline
INV&Taylor&$2.11\pm0.092$&$8.88\pm0.890$&$2.39\pm0.055$&$0.33\pm0.045$&$0.34\pm0.123$&$0.41\pm0.141$\\
INV&Laplace&$2.08\pm0.071$&$8.58\pm0.624$&$2.36\pm0.043$&$0.35\pm0.046$&$0.41\pm0.139$&$0.41\pm0.155$\\
INV&EP&$2.08\pm0.069$&$8.59\pm0.589$&$2.36\pm0.041$&$0.37\pm0.050$&$0.45\pm0.139$&$0.38\pm0.146$\\
INV&KLD&$\bf 2.08\pm0.065$&$8.59\pm0.567$&$2.36\pm0.042$&$0.37\pm0.051$&$0.45\pm0.147$&$0.40\pm0.150$\\
\hline
\end{tabular}
 }
 \caption{Average errors for servo, auto-mpg, housing and abalone datasets. (\small $^{*}$likelihood abbreviations: \textbf{LGP} - GP on the log-transformed outputs;
 \textbf{WGP} - warped GP;
 \textbf{GA\shape} - Gamma\shape-GGPM;
 \textbf{GA\scale} - Gamma\scale-GGPM;
 \textbf{INV} -  Inv. Gaussian-GGPM)}
\label{tab:abalone}
\end{table}

\subsection{Non-negative Real Numbers Experiments}
\label{text:nnrexp}

In this section, we perform four experiments on non-negative real
numbers regression using GGPM with the Gamma
and Inverse Gaussian likelihoods.
We use the following four UCI datasets:
1) abalone\footnote{http://archive.ics.uci.edu/ml/datasets/Abalone} -- predict the
age of abalone from physical measurements;
2) housing\footnote{http://archive.ics.uci.edu/ml/datasets/Housing} -- predict housing
values in suburbs of Boston;
3) auto-mpg\footnote{http://archive.ics.uci.edu/ml/datasets/Auto+MPG}
-- estimate city-cycle fuel consumption in miles per gallon(mpg);
4) servo\footnote{http://archive.ics.uci.edu/ml/datasets/Servo}
-- predict the rise time of a servo-mechanism in terms of two gain settings and two choices of mechanical linkages.
The four datasets are summarized in Table~\ref{tab:nnrnInfo}, which
lists the output range ($Y_{min}$, $Y_{max}$), the input dimension ($X_{dim}$), and the size of the training and test sets ($N_{train}$,
$N_{test}$).

In all experiments, we follow the same testing protocol. A given
number of training samples are randomly selected from the dataset,
and the remaining data is used for testing. This process is repeated
10 times, and the means and standard deviations of MAE and
\NLP~reported.  To test statistical significance, we
use the Friedman test \citep{howell2010fundamental}, which is a
non-parametric test on the differences of several related samples,
based on ranking.
To find candidate hyperparameters in each trial, we use the Taylor-initialization procedure
described in the Section~\ref{multiple_local_minima}.
Then the candidate hyperparameters %
are used as the initialization for the other three inference methods.
We also implemented three other typical GP models: standard GPR,
GPR on the log-transformed data, and the warped GP \citep{Snelson04WGP}.

Experimental results are presented in Table~\ref{tab:abalone}.
The non-negative GGPMs typically perform better than standard GPR.
As expected from Section \ref{text:specialcases}, the log-transformed GP performs similarly to the
Gamma\shape-GGPM with Taylor inference.  However, there are small performance differences due to
the different marginal likelihoods used to estimate the hyperparameters; the former is based on the standard GP marginal in \refeqn{eqn:GPmarg}, while the latter is based on the Taylor marginal in \refeqn{eqn:cfa:marginal} that includes an extra penalty term on the dispersion hyperparameter.
The learned warping functions
of WGP for the four datasets are also log-like, and hence
WGP also has similar performance.
The Inv.-Gaussian-GGPM with Taylor inference can also be viewed as using a
standard GP on the log of the outputs, but with observation noise that is output dependent (see Section \ref{text:specialcases}).
We can benefit from
this property for the auto-mpg and abalone datasets, since Inv.-
Gaussian-GGPM with Taylor inference achieves the smallest fitting
errors. While for the servo and housing dataset, this likelihood is
worse than the Gamma\shape likelihood.

Next, we rank each likelihood function based on the resulting performance in either NLP or MAE,
and use a Friedman test to determine significance\footnote{We pool trials over all inference methods.}.
Table~\ref{tab:ft_lik} shows
the average rankings and $p$ values for various likelihood combinations.
Looking at \NLP, there is a best ranked likelihood function that is statistically significant in all cases except for one ($p<0.001$).  For auto-mpg, the Gamma\shape has slightly better ranking than Inv.-Gaussian, and the difference is marginally significant ($0.05<p<0.06$).
Since each likelihood can have the smallest average rankings, all the three GGPMs
should be taken into consideration when a new dataset is given.

Looking at MAE performance, there are also significant differences in ranking between likelihoods on each dataset.
Note though that in many cases the difference between the top two likelihoods are not significant (e.g., housing, auto-mpg, servo).
This is mainly because the relative performance of different inference methods changes between likelihood functions, e.g., on servo, Taylor inference performs better compared to the other inference methods using the Gamma\shape, but the ranking is reversed when using the Gamma\scale.
Again this indicates that the effects of
likelihood functions to MAE are dataset dependent, so
that the choice of likelihood function is important.

\setlength{\myw}{0.03in}
\begin{table}[tbh]
\centering {\scriptsize
\begin{tabular}{@{}l|c@{\hspace{\myw}}c@{\hspace{\myw}}c@{\hspace{\myw}}cc@{\hspace{\myw}}c@{\hspace{\myw}}cc@{\hspace{\myw}}c@{\hspace{\myw}}cc@{\hspace{\myw}}c@{\hspace{\myw}}c@{}}\cline{1-14}
 \multicolumn{1}{c}{} & \multicolumn{13}{c}{(a) Using \NLP~as measurement}\\
\cline{1-14}
 dataset & GA\shape, &GA\scale, &INV & ($p$) & GA\scale, &INV & ($p$) & GA\shape, &GA\scale & ($p$) & GA\shape, &INV & ($p$)
 \\\hline
  abalone& 1.900, &3.000, &{\bf 1.100} & (0.0000)& 2.000, &{\bf 1.000} & (0.0000) & {\bf 1.000}, &2.000 & (0.0000)& 1.900, &{\bf 1.100} & (0.0000)\\
  housing& 2.000, &{\bf 1.000}, &3.000 & (0.0000)& {\bf 1.000}, &2.000 & (0.0000)& 2.000, &{\bf 1.000} & (0.0000)& {\bf 1.000}, &2.000 & (0.0000)\\
  auto-mpg& {\bf 1.400}, &2.800, &{\bf 1.800} & (0.0000)& 1.850, &{\bf 1.150} & (0.0000) & {\bf 1.050}, &1.950 & (0.0000)&1.350, &1.650 & (0.0578)\\
  servo& {\bf 1.125}, &1.875, &3.000 & (0.0000)& {\bf 1.000}, &2.000 & (0.0000)& {\bf 1.125}, &1.875 & (0.0000)& {\bf 1.000}, &2.000 & (0.0000)\\
\hline\hline
 \multicolumn{1}{c}{} & \multicolumn{13}{c}{(b) Using MAE as measurement}\\
\cline{1-14}
 dataset & GA\shape, &GA\scale, &INV& ($p$) & GA\scale, &INV& ($p$) & GA\shape, &GA\scale & ($p$)& GA\shape, &INV& ($p$)
 \\\hline
  abalone& 2.375, &{\bf 1.563}, &2.063 & (0.0012)& {\bf 1.313}, &1.688 & (0.0163)& 1.750, &{\bf 1.250} & (0.0016)& 1.625, &1.375 & (0.1138)\\
  housing& {\bf 1.750}, &{\bf 1.837}, &2.413 & (0.0053) & {\bf 1.300}, &1.700 & (0.0114)& 1.462, &1.538 & (0.6310)& {\bf 1.288}, &1.712 & (0.0065)\\
  auto-mpg& {\bf 1.816}, &2.395, &{\bf 1.789} & (0.0117)& 1.684, &{\bf 1.316} & (0.0231)& {\bf 1.289}, &1.711 & (0.0094)& 1.526, &1.474 & (0.7456)\\
  servo& {\bf 1.528}, &{\bf 1.472}, &3.000 & (0.0000)& {\bf 1.000}, &2.000 & (0.0000)& 1.528, &1.472 & (0.7389) &  {\bf 1.000}, &2.000 & (0.0000)\\
\hline
\end{tabular}

} \caption{Average rankings and $p$ values (in parenthesis)  for
different likelihood combinations using the Friedman test.
In each grouping, bolded rankings differ significantly from non-bold rankings ($p<0.05$), but not from each other.
} \label{tab:ft_lik}
\end{table}

\setlength{\myw}{0.03in}
\begin{table}[tbh]
\centering {\scriptsize
\begin{tabular}
{@{}ll|c@{\hspace{\myw}}c@{\hspace{\myw}}c@{\hspace{\myw}}c@{\hspace{\myw}}cc@{\hspace{\myw}}c@{\hspace{\myw}}cc@{\hspace{\myw}}c@{\hspace{\myw}}cc@{\hspace{\myw}}c@{\hspace{\myw}}c@{}}\cline{1-16}
 \multicolumn{1}{c}{} & \multicolumn{15}{c}{(a) Using \NLP~ as measurement}\\
 \hline
 Dataset & Lik & TA,&LA,&EP,&KLD &($p)$ & TA,&LA&($p)$ & LA,&EP&($p)$ & EP,&KLD&($p)$ \\
\hline
abalone &  INV& 3.050, &  {\bf 1.650}, &  2.500, &  2.800 &(0.0683)& 1.700, &  1.300 &(0.2059)& {\bf 1.200}, &  1.800 &(0.0339)&1.400, &  1.600 &(0.4795)
\\
housing &  GA$_{sc}$&2.200, &  2.300, &  2.700, &  2.800 &(0.6537)& 1.400, &  1.600 &(0.5271)& 1.400, &  1.600 &(0.5271)&1.500, &  1.500 &(1.0000)
\\
auto-mpg &  GA\shape& 3.400, &  2.300, &   2.050, &  2.250 &(0.0488)&1.800, &  1.200 &(0.0578) &1.600, &  1.400 &(0.4142)& 1.450, &  1.550 &(0.5637)
\\
servo &  GA\shape& 3.050, &  1.900, &  2.500, &  2.550 &(0.2318)& 1.800, &  1.200 &(0.0578)&1.400, &  1.600 &(0.5271)&1.500, &  1.500 &(1.0000)
\\
all & -  & 2.925, &  {\bf 2.038}, &  {\bf 2.438}, & {\bf  2.600} &(0.0137) & 1.675, &  {\bf 1.325} &(0.0269)& 1.400, &  1.600 &(0.1701)& 1.462, &  1.538 &(0.5127)
\\
 \hline
 \hline
 \multicolumn{1}{c}{} & \multicolumn{15}{c}{(b) Using MAE as measurement}\\
\cline{1-16}
Dataset & Lik & TA,&LA,&EP,&KLD &($p)$ & TA,&LA&($p)$ & LA,&EP&($p)$ & EP,&KLD&($p)$
 \\\hline
abalone &  GA\scale& 4.000, &  3.000, &  {\bf 1.750}, &  {\bf 1.250} &(0.0000)& 2.000, & {\bf 1.000} &(0.0016)& 2.000, &  {\bf 1.000} &(0.0016)& 1.750, &  1.250 &(0.0588)
\\
housing &  GA\shape&2.350, &  2.050, &  2.550, &  3.050 &(0.3411)& 1.500, &  1.500 &(1.0000)& 1.350, &  1.650 &(0.3173)& 1.350, &  1.650 &(0.3173)
\\
auto-mpg &  INV& 3.300, &  1.800, &  2.300, &  2.600 &(0.0694)& 1.800, &  1.200 &(0.0578)&  1.300, &  1.700 &(0.2059)&  1.400, &  1.600 &(0.5271)
\\
servo &  GA\scale& 4.000, &  3.000, &  {\bf 1.700}, &  {\bf 1.300} &(0.0000)& 2.000, &  {\bf 1.000} &(0.0016)& 2.000, &  {\bf 1.000} &(0.0016)& 1.700, &  1.300 &(0.1025)
\\
all & -  & 3.413, &  2.462, &  {\bf 2.075}, &  {\bf 2.050} & (0.0000) &  1.825, & {\bf 1.175} &(0.0000) & 1.663, &  {\bf 1.337} &(0.0374)& 1.550, &  1.450 &(0.4795)\\
  \hline
\end{tabular}
} \caption{
Average rankings and $p$ values (in parenthesis)  for
different inference methods using the Friedman test.
Only the best performing likelihood function is considered.
In each grouping, bolded rankings differ significantly from non-bold rankings ($p<0.05$), but not from each other.}
\label{tab:ft_inf}
\end{table}

We next compare approximate inference methods for the best-performing likelihood functions (according to average rankings in Table \ref{tab:ft_lik}) on each dataset.
Table~\ref{tab:ft_inf} shows the average rankings and the corresponding $p$ values using a Friedman test.
Looking the ranking based on \NLP~in Table~\ref{tab:ft_inf}a, EP and KLD have similar rankings (within 0.2) on each dataset, and any differences are not statistically significant.
Similarly, LA and EP also have similar rankings (within 0.2)  for each dataset except on abalone.  Over all datasets, LA has the best ranking (2.038), but this result is not statistically significant, suggesting that the rank orderings of LA, EP, and KLD are not consistent.
Finally, LA has a statistically better ranking than TA, when pooling over all datasets, although the difference is not large; LA outperforms TA  about two-thirds of the time.

The results are similar when looking at the rankings based on MAE in Table~\ref{tab:ft_inf}b.
First looking at the rankings over all datasets, as before, EP and KLD have almost identical rankings in MAE,
but now EP has better MAE than LA about two-thirds of the time (statistically significant).
Finally, LA typically dominates TA in MAE ranking, over all datasets.
Interestingly, there are some datasets (e.g. housing, auto-mpg), where there is no significant difference between the inference algorithms.  In other words, the best inference algorithm may change in each trail, based on the particular training and test set.

In summary,
the performances of inference methods are highly affected by likelihoods, datasets
and evaluation metrics. For a given dataset,
the choice of likelihood has a large impact on the predictive density (\NLP), with only one likelihood usually dominating, and less so on the prediction error (MAE), where typically more than one likelihood can achieve low error.
Given the ``correct'' likelihood, the performance of inference methods tends to be similar, e.g., LA, EP, and KLD have similar rankings when evaluated with \NLP, and EP and KLD have similar ranking for MAE.
However, given the ``wrong'' likelihood, the performance of the inference algorithms can be highly affected by the dataset and the evaluation metrics.

\subsection{Range Data Experiments}
\label{text:betaexp}

In this experiment, we consider conversion of
device-dependent RGB values to device- and illuminant-independent
reflectance spectra.
In~\citet{Heikkinen2008}, this conversion is cast as a
regularized regression problem, where the input can be RGB or HSV color values,
and the output is reflectance values at sampled wavelengths.
In particular, the reflectance spectra is first scaled from the [0~1] interval to [-1~1], and then mapped to a real value via the inverse hyperbolic tangent (arctanh) function,  then a regularized regression framework is applied.
The method in~\citet{Heikkinen2008} is equivalent to applying standard GPR to the logit-transformed spectral values, as discussed in Section \ref{text:specialcases}.
Note that the hyperparameters are fixed in~\citet{Heikkinen2008}, whereas using the GP interpretation, the hyperparameters can be estimated automatically using maximum marginal likelihood.

Since the regression output is constrained to the unit interval, we consider Beta-GGPM for spectra reflectance regression,
and perform experiments on the Munsell dataset, consisting of 1269 RGB/spectral pairs.
Following the protocol of \citet{Heikkinen2008}, we used 669 for training and 600 for testing, and results are averaged over 10 trials.
We also used a smaller training set, consisting of 20\% of the original training set.
Hyperparameters are learned using maximum marginal likelihood.

\begin{table}[h]
\centering {\scriptsize
\begin{tabular}{@{}lc|ccc@{}}
\cline{3-5}
 \multicolumn{2}{c}{} & \multicolumn{3}{c}{(a) Full training dataset}\\
\cline{1-5}Model & Inference & Avg. Error & Max. Error & Std. Error
 \\\hline
 GP & Exact & $0.0090\pm 0.00032$ & $\bf{0.0919}\pm 0.0195$ & $0.0102\pm 0.00100$\\
  \hline
 arctanh+GP & Exact  & $\bf{0.0087}\pm 0.00036$ & $0.0946\pm 0.0205$ & $\bf{0.0101}\pm 0.00098$\\
  \hline
 Beta-GGPM & Taylor & $0.0088\pm 0.00037$ & $0.0961\pm 0.0195$ & $0.0103\pm 0.00098$\\
 Beta-GGPM & Laplace & $0.0088\pm 0.00038$ & $0.0965\pm 0.0199$ & $0.0103\pm 0.00100$\\
 Beta-GGPM & EP  & $\bf{0.0087}\pm 0.00038$ & $0.0946\pm 0.0208$ & $\bf{0.0101}\pm 0.00100$\\
\hline %
\end{tabular}
\begin{tabular}{@{}lc|ccc@{}}
\cline{3-5}
 \multicolumn{2}{c}{} & \multicolumn{3}{c}{(b) Small training dataset}\\
\cline{1-5}Model & Inference & Avg. Error & Max. Error & Std. Error
 \\\hline
GP&Exact&$0.0132\pm0.00115$&$0.0999\pm0.0179$&$0.0132\pm0.00221$\\
\hline
arctanh+GP&Exact&$0.0123\pm0.00074$&$\bf 0.0965\pm0.0186$&$\bf 0.0124\pm0.00141$\\
 \hline
Beta-GGPM&Taylor&$0.0123\pm0.00080$&$0.0970\pm0.0170$&$\bf 0.0124\pm0.00147$\\
Beta-GGPM&Laplace&$0.0123\pm0.00079$&$0.0973\pm0.0172$&$\bf 0.0124\pm0.00143$\\
Beta-GGPM&EP&$\bf 0.0122\pm0.00077$&$0.0967\pm0.0183$&$\bf 0.0124\pm0.00141$\\
\hline
\end{tabular}
}
\caption{Average errors for the Munsell dataset.
}
\label{tab:spectra}
\end{table}

The experimental results are presented in Table~\ref{tab:spectra}.
First, the standard GP performs worse than Beta-GGPM, due to the mismatch between output domain and actual outputs.
This effect is more pronounced when less training data is available; when using the smaller training set,
the average error drops around 8\% for the Beta-GGPM versus the GP.
Next, Beta-GGPM with Taylor inference and
Gauss-GGPM+arctanh have almost the same values for the three error
metrics, this is consistent to our conclusion that Beta-GGPM with
Taylor inference can be viewed as using a standard GP on the logit
transformation of the outputs.
Finally, the three inference methods (TA, LA, EP) perform similarly for Beta-GGPM on this dataset, in terms of average error. Table~\ref{tab:ft_spectra} shows the Friedman test results for the different inference combinations.
For the full training set, LA and EP have similar average rankings, with TA ranked 0.6 worse.  However, the differences are not statistically significant, due to the similar average error values. For the reduced training set, EP has the best ranking, followed by TA and then LA. Again, the rankings are not statistically significant,
although EP is marginally better than LA ($0.05<p<0.06$).

\setlength{\myw}{0.03in}
\begin{table}[!tbhp]
\centering {\scriptsize
\begin{tabular}
{@{}l|c@{\hspace{\myw}}c@{\hspace{\myw}}c@{\hspace{\myw}}cc@{\hspace{\myw}}c@{\hspace{\myw}}cc@{\hspace{\myw}}c@{\hspace{\myw}}cc@{\hspace{\myw}}c@{\hspace{\myw}}c@{}}
 \cline{2-14}
 \multicolumn{1}{c}{} & \multicolumn{13}{c}{(a) Full training dataset}\\
 \cline{1-14}
 Metrics & TA,&LA,&EP&($p$) & TA,&LA&($p$) & TA,&EP&($p$) & LA,&EP&($p$)
 \\\hline
  Avg. Error& 2.450, & 1.850, & 1.700 &(0.0724)& 1.750, & \bf{1.250} &(0.0253) & 1.700, & 1.300 &(0.1024)& 1.600, & 1.400 &(0.4142)\\
  Max. Error& 1.900, & 2.500, & 1.600 &(0.1225)&1.300, & 1.700 &(0.2059)& 1.600, & 1.400 &(0.5271)& 1.800, & 1.200 &(0.0578)\\
  Std. Error& 2.500, & 2.050, & 1.450 &(0.0581) & 1.700, & 1.300 &(0.2059) & 1.800, & 1.200 &(0.0578)& 1.750, & 1.250 &(0.0956)\\
  \hline
\end{tabular}

\begin{tabular}
{@{}l|c@{\hspace{\myw}}c@{\hspace{\myw}}c@{\hspace{\myw}}cc@{\hspace{\myw}}c@{\hspace{\myw}}cc@{\hspace{\myw}}c@{\hspace{\myw}}cc@{\hspace{\myw}}c@{\hspace{\myw}}c@{}}
 \cline{2-14}
 \multicolumn{1}{c}{} & \multicolumn{13}{c}{(b) Small training dataset}\\
 \cline{1-14}
 Metrics & TA,&LA,&EP&($p$) & TA,&LA&($p$) & TA,&EP&($p$) & LA,&EP&($p$)
 \\\hline
  Avg. Error& 2.050, & 2.400, &1.550 &(0.1316)&1.350, &1.650 &(0.3173) &1.700, &1.300 &(0.2059)&1.750, &1.250 &(0.0588)\\
  Max. Error& 1.800, & 2.000, & 2.200 &(0.6703)&1.400, & 1.600 &(0.5271)& 1.400, & 1.600 &(0.5271)& 1.400, & 1.600 &(0.5271)\\
  Std. Error& 1.700, & 2.350, & 1.950 &(0.3225) & 1.250, & 1.750 &(0.0956) & 1.450, & 1.550 &(0.7389)& 1.600, & 1.400 &(0.5271)\\
  \hline
\end{tabular}
} \caption{Beta-GGPM: Average rankings of different inference methods and $p$ values using the Friedman test. Bolded rankings differ significantly from non-bold rankings ($p<0.05$).
} \label{tab:ft_spectra}
\end{table}

\subsection{Counting experiments}
\label{text:countexp}
We perform two counting
experiments using GGPMs with Poisson-based likelihoods.  In all
cases, predictions are based on the mode of the distribution for
GGPMs, and the rounded, truncated mean for GPR.
In the first experiment, we perform crowd counting using the UCSD crowd counting dataset\footnote{Data set at http://visal.cs.cityu.edu.hk/downloads/\#ucsdpeds-feats} from
\citet{Chan2008cvpr}.
The dataset contains 30-dimensional features extracted from images and the corresponding number of people in each image, for two different directions (right and left motion).  The goal is to predict the number of people using just the image features.  The right crowd contains more people (average of 14.69 per image) than the left crowd (average of 9.98).
The dataset consists of 2000 feature/count pairs for each direction, and following \citet{Chan2008cvpr}, we use 800 for training and 1200 of testing.  We predict using the Poisson and COM-Poisson GGPMs and the exponential mean mapping (canonical link function), as well as the versions using the linear mean mapping (linearized link function) from Section \ref{text:linearizedmean}. The compound linear plus RBF kernel was used for all models.

The crowd counting results are presented in Table \ref{tab:crowds}.
On the ``right'' crowd, the Poisson- and COM-Poisson-GGPMs perform better using the exponential mapping versus the linear mapping.
This is due to the large number of people in the ``right'' crowd, which
leads to a more non-linear (exponential) trend in the feature space.
  In contrast,  the linearized link functions perform better on the ``left crowd'', indicating a more linear trend in the data (due to smaller crowd sizes and fewer occlusions).
Looking at the likelihood functions, the Poisson likelihood has higher accuracy on the ``right'' crowd,
whereas the COM-Poisson is better on the ``left'' crowd.
The main difference is that COM-Poisson provides some flexibility to control the variance of the observation noise,
which helps for the ``left'' crowd.

\comments{
\begin{table}[!thbp]
\centering \scriptsize
\begin{tabular}{@{}c|l|cc|cc||l|cc|cc@{}}
\cline{2-11}
 \multicolumn{1}{c}{} & \multicolumn{1}{c|}{} & \multicolumn{2}{c|}{Right crowd}& \multicolumn{2}{c||}{Left crowd} & \multicolumn{1}{c|}{}
 & \multicolumn{2}{c|}{Right crowd} & \multicolumn{2}{c}{Left crowd}\\
\cline{1-11} Inference & Lik$^{*}$  & MAE & \NLP & MAE &
\NLP & Lik$^{*}$ & MAE & \NLP & MAE & \NLP
 \\ \hline
Exact & GP & $1.56$ & $2.94$ & $0.85$ & $1.83$ & - & - & - & - & -\\
\hline
Taylor & PO & $\bf{1.25}$ & $\bf{2.33}$ & $1.03$ & $2.20$ & LPO   & $1.36$ & $2.34$ & $0.88$ & $2.18$\\
Laplace & PO & $1.27$ & $2.33$ & $1.03$ & $2.20$ & LPO  & $1.36$ & $2.34$ & $0.87$ & $2.18$\\
EP & PO & $1.27$ & $2.33$ & $1.03$ & $2.20$ & LPO  & $1.37$ & $2.34$ & $0.87$ & $2.18$ \\
\hline
Taylor & CPO & $1.39$ & $2.50$ & $0.94$ & $1.76$ & LCPO  & $1.48$ & $2.50$ & $0.91$ & $1.63$ \\
Laplace & CPO & $1.39$ & $2.54$ & $0.94$ & $1.77$ & LCPO  & $1.49$ & $2.51$ & $0.85$ & $\bf{1.52}$ \\
EP & CPO & $1.43$ & $2.54$ & $0.94$ & $1.76$ & LCPO  & $1.49$ & $2.52$ & $\bf{0.83}$ & $1.55$ \\
 \hline
\end{tabular}
\caption{Mean absolute errors for crowd counting. (\small
$^{*}$likelihood abbreviation: Poisson-GGPM(\textbf{PO}), Linearized
Poisson-GGPM(\textbf{LPO}), COM-Poisson-GGPM(\textbf{CPO}), Lin.
COM-Poisson-GGPM(\textbf{LCPO}))} \label{tab:crowds}
\end{table}
}

\begin{table}[thbp]
\centering \scriptsize
\begin{tabular}{@{}cl|cc|cc||cc|cc@{}}
\cline{3-10}
 \multicolumn{2}{c|}{}
 & \multicolumn{4}{c||}{Right crowd}
 & \multicolumn{4}{c}{Left crowd}\\
\hline
 & &
 \multicolumn{2}{c|}{exponential mean} &
 \multicolumn{2}{c||}{linearized mean} &
 \multicolumn{2}{c|}{exponential mean} &
 \multicolumn{2}{c}{linearized mean}
 \\
Likelihood
& Inference
& MAE & \NLP
& MAE & \NLP
& MAE & \NLP
& MAE & \NLP
 \\ \hline \hline
GP & Exact
 & - & -
 & $1.56$ & $2.94$
 & - & -
 & $0.85$ & $1.83$
 \\
\hline
Poisson & Taylor
& $\bf{1.25}$ & $\bf{2.33}$
& $1.36$ & $2.34$
& $1.03$ & $2.20$
& $0.88$ & $2.18$\\
Poisson & Laplace
& $1.27$ & $2.33$
& $1.36$ & $2.34$
& $1.03$ & $2.20$
 & $0.87$ & $2.18$\\
Poisson & EP
& $1.27$ & $2.33$
 & $1.37$ & $2.34$
& $1.03$ & $2.20$
& $0.87$ & $2.18$ \\
\hline
COM-Poisson & Taylor
& $1.39$ & $2.50$
& $1.48$ & $2.50$
& $0.94$ & $1.76$
& $0.91$ & $1.63$ \\
COM-Poisson & Laplace
& $1.39$ & $2.54$
& $1.49$ & $2.51$
& $0.94$ & $1.77$
 & $0.85$ & $\bf{1.52}$ \\
COM-Poisson & EP & $1.43$ & $2.54$
& $1.49$ & $2.52$
& $0.94$ & $1.76$
& $\bf{0.83}$ & $1.55$ \\
 \hline
\end{tabular}
\caption{Mean absolute errors for crowd counting with comparisons
between likelihood functions, inference methods, and link functions.
} \label{tab:crowds}
\vspace{-0.3in}
\end{table}

\setlength{\myw}{0.04in}
\begin{table}[thbp]
\centering \scriptsize
\begin{tabular}{@{}c@{\hspace{\myw}}l@{\hspace{\myw}}|c@{\hspace{\myw}}c@{\hspace{\myw}}|c@{\hspace{\myw}}c@{\hspace{\myw}}||c@{\hspace{\myw}}c@{\hspace{\myw}}|c@{\hspace{\myw}}c@{}}
\cline{3-10}
\multicolumn{2}{c}{}
& \multicolumn{4}{|c||}{Right crowd}
& \multicolumn{4}{c}{Left crowd}\\
\hline
 & &
 \multicolumn{2}{c|}{exponential mean} &
 \multicolumn{2}{c||}{linearized mean} &
 \multicolumn{2}{c|}{exponential mean} &
 \multicolumn{2}{c}{linearized mean}
 \\
Lik.
& Inf.
& MAE & \NLP
& MAE & \NLP
& MAE & \NLP
& MAE & \NLP
 \\ \hline \hline
GP & Exact
 & - & -
 & 1.56 $\pm$ 0.027 & 2.56 $\pm$ 0.059
 & - & -
 & 0.88 $\pm$ 0.022 & 1.69 $\pm$ 0.055
 \\
\hline
PO & TA
& \bf{1.31 $\pm$ 0.038} & {2.34 $\pm$ 0.038}
& 1.46 $\pm$ 0.036 & 2.35 $\pm$ 0.034
& 1.04 $\pm$ 0.023 & 2.19 $\pm$ 0.036
& 0.95 $\pm$ 0.024 & 2.17 $\pm$ 0.027\\
PO & LA
& 1.33 $\pm$ 0.033 & 2.34 $\pm$ 0.038
& 1.43 $\pm$ 0.030 & 2.35 $\pm$ 0.035
& 1.02 $\pm$ 0.019 & 2.19 $\pm$ 0.030
& 0.93 $\pm$ 0.019 & 2.17 $\pm$ 0.019\\
PO & EP
& 1.33 $\pm$ 0.034 & 2.34 $\pm$ 0.038
& 1.42 $\pm$ 0.032 & 2.35 $\pm$ 0.035
& 1.02 $\pm$ 0.018 & 2.19 $\pm$ 0.030
& 0.93 $\pm$ 0.018 & 2.17 $\pm$ 0.019\\
\hline
COM & TA
& 1.54 $\pm$ 0.076 & 2.61 $\pm$ 0.106
& 1.50 $\pm$ 0.046 & 2.43 $\pm$ 0.079
& 0.96 $\pm$ 0.038 & 1.83 $\pm$ 0.055
& 0.92 $\pm$ 0.045 & 1.68 $\pm$ 0.052\\
COM & LA
& 1.55 $\pm$ 0.098 & 2.22 $\pm$ 0.069
& 1.58 $\pm$ 0.037 & 2.43 $\pm$ 0.072
& 0.95 $\pm$ 0.021 & 1.81 $\pm$ 0.047
& \bf{0.86 $\pm$ 0.023} & 1.65 $\pm$ 0.043\\
COM & EP
& 1.40 $\pm$ 0.033 & \bf{2.20 $\pm$ 0.057}
& 1.54 $\pm$ 0.048 & 2.35 $\pm$ 0.034
& 0.92 $\pm$ 0.016 & 1.75 $\pm$ 0.051
& 0.86 $\pm$ 0.027 & \bf{1.62 $\pm$ 0.049}\\
 \hline
\end{tabular}
\caption{Average errors for crowd counting dataset using reduced training set. (\small likelihood abbreviations: \textbf{PO} - Poisson;
 \textbf{COM} - COM-Poisson)
} \label{tab:avg:crowds}
\vspace{-0.3in}
\end{table}

\setlength{\myw}{0.03in}
\begin{table}[tbhp]
\centering {\scriptsize
\begin{tabular}{@{}l|c@{\hspace{\myw}}c@{\hspace{\myw}}c@{\hspace{\myw}}c@{\hspace{\myw}}cc@{\hspace{\myw}}c@{\hspace{\myw}}cc@{\hspace{\myw}}c@{\hspace{\myw}}cc@{\hspace{\myw}}c@{\hspace{\myw}}c@{}}\cline{1-15}
 \multicolumn{1}{c}{} & \multicolumn{13}{c}{(a) Using \NLP~as measurement}\\
\cline{1-15}
 dataset & PO, & L-PO, &COM, &L-COM & ($p$) & PO, &L-PO & ($p$) & COM, &L-COM & ($p$) & L-PO, &L-COM & ($p$)
 \\\hline
  Right& {\bf  1.758}, &2.788, &2.030, &3.424 & (0.0000)& {\bf 1.000}, &2.000 & (0.0000) & 1.333, &1.667 & (0.0555)&{\bf 1.121}, &1.879 & (0.0000)\\
  Left& 4.000, &3.000 &2.000, &{\bf 1.000} & (0.0000)& 2.000, &{\bf 1.000} & (0.0000)& 2.000, &{\bf 1.000} & (0.0000)& 2.000, &{\bf 1.000} & (0.0000)\\
\hline\hline
\multicolumn{1}{c}{} & \multicolumn{13}{c}{(b) Using MAE as measurement}\\
\cline{1-15}
 dataset & PO, & L-PO, &COM, &L-COM & ($p$) & PO, &L-PO & ($p$) & COM, &L-COM & ($p$) & L-PO, &L-COM & ($p$)
\\\hline
  Right& {\bf 1.061}, &2.333, &3.061, & 3.545 & (0.0000)& {\bf 1.000}, & 2.000 & (0.0000) &  1.394, &1.606 & (0.2230)& {\bf 1.061}, & 1.939 & (0.0000)\\
  Left& 3.933, &2.333 &2.656, &{\bf 1.078} & (0.0000)& 2.000, &{\bf 1.000} & (0.0000)& 2.000, &{\bf 1.000} & (0.0000)& 1.944, &{\bf 1.056} & (0.0000)\\
\hline
\end{tabular}
} \caption{Average rankings and $p$ values (in parenthesis)  for
different likelihood combinations using the Friedman test.  Bolded rankings differ significantly from non-bold rankings ($p<0.05$).
(\small likelihood abbreviations: \textbf{PO} - Poisson;
\textbf{L-PO} - Linear Poisson;
 \textbf{COM} - COM-Poisson);
  \textbf{L-COM} - Linear COM-Poisson)
} \label{tab:crowd_lik}
\end{table}

To do hypothesis testing, we randomly selected 10 small training sets, consisting of 400
feature/count pairs, from the original training dataset. The learned GGPMs are evaluated on the
original test dataset, and experimental results are presented in Table 13. Table 14 shows the average
rankings and $p$ values using the Friedman test for different likelihoods combinations. We can see the
Poisson- and COM-Poisson-GGPMs perform significantly better than the corresponding linearized
models on the ``right'' crowd, and vice versa, the linearized link functions perform better on the ``left'' crowd. This
is consistent with our observations from the experiments on the original training and test datasets.

In the second experiment, the GGPM is used for age estimation of face images.
We use the FG-NET dataset\footnote{Data set at http://www.fgnet.rsunit.com},
which consists of face images of 82 people at different ages (average of 12 images per person).
The input vector into the GGPM is 150 facial features, which are extracted using active appearance
models \citep{Cootes2001AAM}, while the output is the age of the face.
We used leave-one-person-out testing as in \cite{Zhang2010mtwgp} to evaluate the performance of difference GGPMs. For each fold, the images of one person are used for testing, and all the other people are used as the training dataset.

The experiment results for age estimation are presented in Table \ref{tab:age}.
The linear Poisson-GGPM has the lowest MAE of 5.82 versus 6.12 for
standard GPR.
Table~\ref{tab:ft_age_MAE}a shows the Friedman test results for different likelihood combinations, and results indicate that the linearized Poisson GGPM significantly better ranking than the other two GGPMs.
Table~\ref{tab:ft_age_MAE}b shows the Friedman test for different inference combinations.
Although there are small differences in MAE between the inference algorithms for each likelihood, the rankings
are very similar, which indicates that no inference method dominates in terms of MAE.
In general, Laplace and EP perform similarly, and there is no statistically significant difference between them.
Moreover, Taylor approximation can outperform the other two methods for some GGPMs, e.g. Neg.
binomial-GGPM, but again the difference is not statistically significant.
Finally, Figure \ref{fig:age} presents an example prediction on a test person.

\comments{
\begin{table}[tbhp]
\vspace{-0.05in} \centering \footnotesize
\begin{tabular}{l@{\hspace{0.1in}}c@{\hspace{0.1in}}c@{\hspace{0.1in}}c}
\hline
 Method & Inference & MAE(R) & MAE(L)
 \\
 \hline
Gauss & Exact & $1.556$ & $0.853$ \\
Poisson GGPM & Taylor & $\bf{1.248}$ & $1.038$ \\
Poisson GGPM & Laplace & $1.330$ & $1.029$ \\
Poisson GGPM & EP & $1.361$ & $1.162$ \\
Linearized Poisson GGPM & Taylor  & $1.363$ & $0.880$ \\
Linearized Poisson GGPM & Laplace & $1.363$ & $0.868$ \\
Linearized Poisson GGPM & EP & $1.364$ & $0.868$ \\
COM-Poisson GGPM & Taylor & $1.388$ & $1.161$ \\
Lin. COM-Poisson GGPM & Taylor & $1.507$ & $\bf{0.848}$ \\
 \hline
\end{tabular}
\caption{Mean absolute errors for crowd counting. \NOTE{update this
table}} \label{tab:crowds} \vspace{-0.15in}
\end{table}
}

\begin{minipage}{18cm}
\begin{minipage}[h]{7cm}
\vspace{0.2in} \centering \scriptsize
\begin{tabular}{@{}l|cc@{}}
\hline
 Method & Inference & MAE
 \\
 \hline
GP & Exact   %
& $6.12$ \\
Warped GP \citep{Zhang2010mtwgp} & Exact    %
& $6.11$ \\
\hline
Poisson GGPM & Taylor  %
& $6.44$ \\
Poisson GGPM & Laplace  %
& $6.41$ \\
Poisson GGPM & EP  %
& $6.40$ \\
\hline
Linearized Poisson GGPM & Taylor   %
& $5.98$ \\
Linearized Poisson GGPM & Laplace  %
& $\bf{5.82}$\\
Linearized Poisson GGPM& EP  %
& $5.83$
\\ \hline
Neg. binomial GGPM & Taylor   %
& $6.19$ \\
Neg. binomial GGPM & Laplace  %
& $6.37$\\
Neg. binomial GGPM & EP  %
& $6.37$
\\ \hline
\end{tabular}
\makeatletter\def\@captype{table}\makeatother  \caption{MAEs for age
estimation.} \label{tab:age}
\end{minipage}
\begin{minipage}[h]{8cm}
\vspace{0.2in}\centering \psfig{file=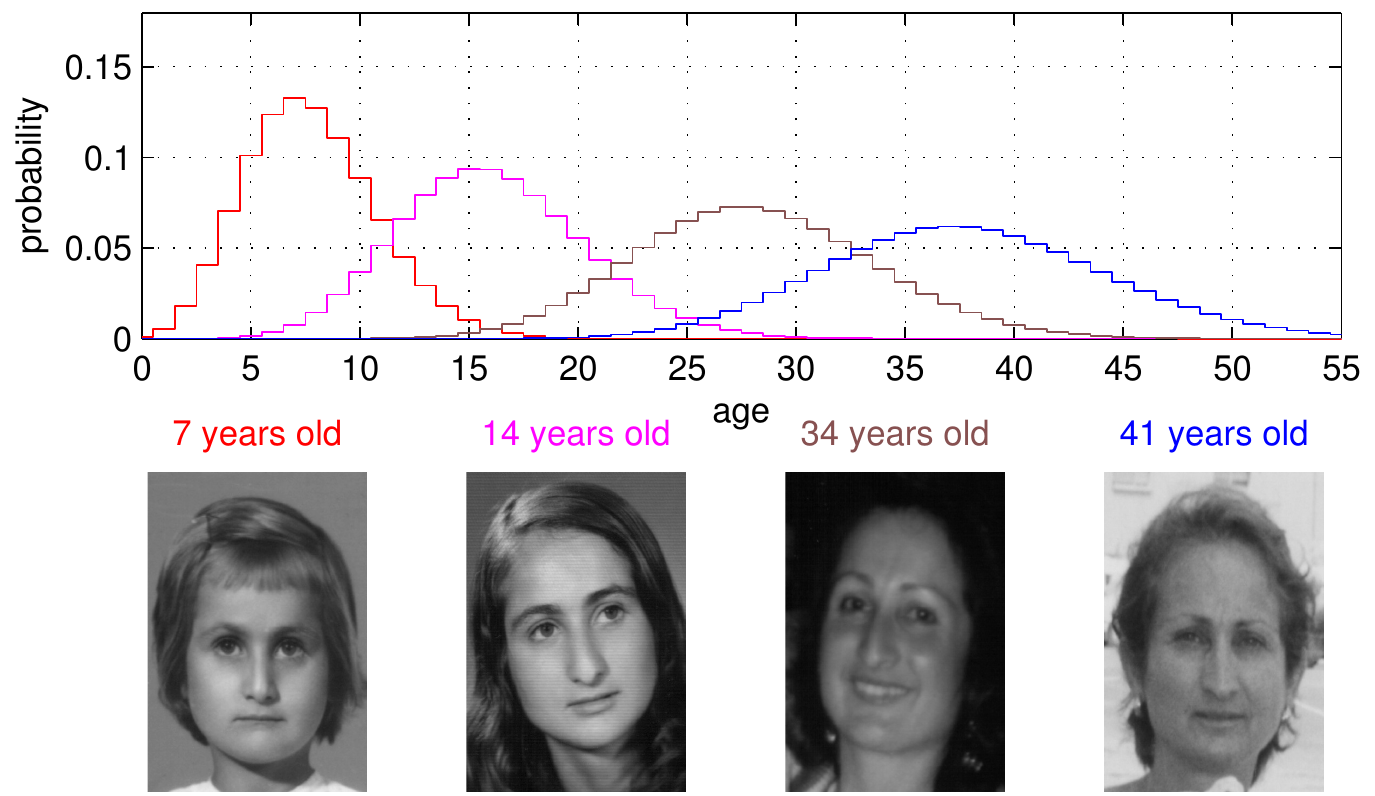, scale=0.49}
\makeatletter\def\@captype{figure}\makeatother  \caption{Examples of
predicted distributions.} \label {fig:age}
\end{minipage}
\end{minipage}

\comments{
\begin{table}[tbhp]
\centering {\scriptsize
\begin{tabular}
{@{}l|cccc@{}}\hline
 Data & Tay-Lap-EP & Tay-Lap & Tay-EP & Lap-EP
 \\\hline
  Poisson&0.2550 (2.128, 1.976, 1.896)& 0.3113 (1.555, 1.445)&0.1687 (1.573, 1.427)& 0.4233 (1.530, 1.470)\\
  Lin. Poisson&0.2094 (2.152, 1.927, 1.921)&0.2216 (1.567, 1.433)& 0.1175 (1.585, 1.415)& 0.8946 (1.494, 1.506)\\
  Neg. binomial&0.4013 (1.890, 2.073, 2.037)& 0.2100 (1.433, 1.567)& 0.4189 (1.457, 1.543)& 0.8815 (1.506, 1.494)\\
  \hline
\end{tabular}
} \caption{$p$ values and average rankings of Friedman test for
different inference combinations.} \label{tab:ft_age_MAE}
\end{table}
}

\setlength{\myw}{0.03in}
\begin{table}[tbhp]
\centering {\scriptsize
\begin{tabular}
{@{}l@{\hspace{0.28in}}|c@{\hspace{\myw}}c@{\hspace{\myw}}c@{\hspace{\myw}}cc@{\hspace{\myw}}c@{\hspace{\myw}}cc@{\hspace{\myw}}c@{\hspace{\myw}}cc@{\hspace{\myw}}c@{\hspace{\myw}}c@{}}\hline
 \multicolumn{1}{c}{} & \multicolumn{13}{c}{(a) Different likelihoods}\\
\cline{1-14}
 Inference & PO,&L-PO,&NB&($p$) & L-PO,&NB&($p$) & PO,&L-PO&($p$) & PO,&NB&($p$)
 \\\hline
  all&2.193, &\bf{1.805}, &2.002 &(0.0001) & \bf{1.433}, & 1.567&(0.0328) & 1.628, & \bf{1.372} &(0.0000)& 1.565, & \bf{1.435} &(0.0364)\\
  \hline
\end{tabular}

\begin{tabular}
{@{}l|c@{\hspace{\myw}}c@{\hspace{\myw}}c@{\hspace{\myw}}cc@{\hspace{\myw}}c@{\hspace{\myw}}cc@{\hspace{\myw}}c@{\hspace{\myw}}cc@{\hspace{\myw}}c@{\hspace{\myw}}c@{}}\hline
\multicolumn{1}{c}{} & \multicolumn{13}{c}{(b) Different inference methods}\\
\cline{1-14}
 Likelihood & TA,&LA,&EP&($p$) & TA,&LA&($p$) & TA,&EP&($p$) & LA,&EP&($p$)
\\ \hline
  Poisson& 2.128, & 1.976, & 1.896 &(0.2550)& 1.555, & 1.445 &(0.3113) & 1.573, & 1.427 &(0.1687)& 1.530, & 1.470 &(0.4233)\\
  Lin. Poisson& 2.152, & 1.927, & 1.921 &(0.2094)&1.567, & 1.433 &(0.2216)& 1.585, & 1.415 &(0.1175)& 1.494, & 1.506 &(0.8946)\\
  Neg. binomial& 1.890, & 2.073, & 2.037 &(0.4013) & 1.433, & 1.567 &(0.2100) & 1.457, & 1.543 &(0.4189)& 1.506, & 1.494 &(0.8815)\\
  \hline
\end{tabular}
} \caption{Average rankings and $p$ values using the Friedman test. Bolded rankings differ significantly from non-bold rankings ($p<0.05$).
 (\small likelihood abbreviations: \textbf{PO} - Poisson;
 \textbf{L-PO} - Lin. Poisson;
 \textbf{NB} - Neg. binomial)} \label{tab:ft_age_MAE}
\end{table}

\subsection{Summary}

Our experiments have considered a variety of likelihood functions, inference methods, and datasets.
In general, the choice of likelihood function is more important than the choice of approximate inference algorithm.
Using a likelihood function that matches the output domain and observation noise will typically lead to better performance over the standard GP.
The link function can also affect the final result, and the selection of the link function is dataset dependent, similar to the choice of the kernel function.

Looking at the inference methods, EP and KLD typically have similar performance, and Laplace inference also achieves comparable results.   Taylor inference sometimes yields the smallest MAEs, while  its \NLP~is often larger than the other three inference.  This suggests that accurate estimation in terms of NLP does not always lead to smaller fitting error in terms of MAE. The performances of different inference methods are also
 highly affected by the distribution of training and test data, due to the systematic ordering observed in the posterior means (Section \ref{1D_example}).
Finally, on many datasets there is no dominant approximate inference algorithm, yielding mixed results and average rankings with differences that are not statistically significant.

\section{Conclusions}

In this paper, we have studied approximate inference for generalized Gaussian process models.
The GGPM is a unifying framework for existing GP models, where
the observation likelihood of the GP model is itself parameterized using the exponential family distribution (EFD).
This allows approximate inference algorithms to be derived using the general form of the EFD.
A particular GP model can then be formed by setting the parameters of the EFD, which instantiates a particular observation likelihood function for an output domain.
In addition to the observation likelihood, the GGPM also has a link function that controls the mapping between the latent variable (GP prior) and the mean of the output distribution.  By appropriately setting the link function, mean trends (e.g., logarithmic) can be learned that would otherwise not be possible with standard positive-definite kernel functions.

We also study an approximate inference method based on a Taylor approximation, which is non-iterative and computationally efficient.  The Taylor approximation can justify many common heuristics in GP modeling (e.g., label regression, GPR on log-transformed outputs, and GPR on logit-transformed outputs) as principled inference with a particular likelihood function.
We also present approximate inference for GGPMs using the Laplace approximation, expectation propagation, and KL divergence minimization.  Furthermore, we demonstrate that the posterior means of the Taylor, Laplace, and EP approximations usually have a specific ordering, which are a result of the particular method of approximation.
As a consequence, since the posterior means are biased in this way, the prediction error of a particular inference algorithm heavily depends on the distributions of the training and test data.
Finally, we perform hyperparameter estimation using the Taylor approximation to initialize the other less-efficient approximate inference methods.  Our initialization procedure can greatly increase the speed of the the learning phase, while not significantly affecting the quality of the hyperparameters.  In addition, we did not notice any convergence issues of EP when using our initialization procedure.

We conduct a comprehensive set of experiments, using a variety of likelihood functions, approximate inference methods, and datasets.
In our experiments, we found that the selection of the correct likelihood function has a larger impact on the prediction accuracy than the approximate inference method.
Indeed, in many cases there was no dominant inference method, and any differences in average ranking were not statistically significant.  Whereas, the appropriate choice of likelihood function and link function improved accuracy significantly.

Finally, in this paper we have only considered univariate observations for the GGPM.  Future work will use the multivariate exponential family distribution to form a multivariate GGPM.
Such a model would encompass existing multivariate GPs, such as
GP ordinal regression \citep{Chu2005GPOR}, multi-class GP classification \citep{Girolami05VBM,Kim2006emep,Williams1998GPC}, and
semiparametric latent factor models \citep{Teh05SPLFM}.

\appendix

\section{Closed-form Taylor Approximation}
\label{app:Taylor}

This appendix contains the derivations for the closed-form Taylor approximation in Section \ref{text:Taylor}

\subsection{Joint likelihood approximation}
\label{app:Taylor:joint}

\comments{ The joint likelihood of data and parameters is,
    \begin{align}
    \lefteqn{
    \log p(\vy,\veta|\mX) = \log p(\vy|\vtheta(\veta)) + \log p(\veta|\mX)} \\
    &= \sum_{i=1}^n \left( \frac{1}{a(\phi)} \left[y_i\theta(\eta_i)  - b(\theta(\eta_i))\right] + \log h(y_i,\phi) \right)
        - \frac{1}{2} \veta^T \mK^{-1}\veta - \frac{1}{2} \log \detbar{\mK} - \frac{n}{2}\log 2\pi
    \label{eqn:cfa:joint}
    \end{align}
Note that the terms that prevents the posterior of $\veta$ from
being Gaussian are the functions $b(\theta(\eta_i))$ and
$\theta(\eta_i)$. Let us consider a second-order Taylor expansion of
the likelihood term at the point $\teta_i$,
    \begin{align}
    \log p(y_i|\theta(\eta_i)) &=
    \frac{1}{a(\phi)} \left[y_i\theta(\eta_i)  - b(\theta(\eta_i))\right] + \log h(y_i,\phi)
    \\
    &\approx \log p(y_i|\theta(\teta_i)) + \tu_i (\eta_i-\teta_i) - \frac{1}{2} \tw_i^{-1} (\eta_i-\teta_i)^2
    \label{eqn:cfa:taylor}
    \end{align}
where $\tu_i = u(\teta_i,y_i)$ and $\tw_i = w(\teta_i,y_i)$ as
defined in (19). } Summing over \refeqn{eqn:cfa:taylor}, we have the
approximation to output likelihood
    \begin{align}
    \log p(\vy|\vtheta(\veta)) &=
    \sum_{i=1}^n \log p(y_i|\theta(\eta_i))
    \approx \sum_{i=1}^n \log p(y_i|\theta(\teta_i)) + \tu_i (\eta_i-\teta_i) - \frac{1}{2} \tw_i^{-1} (\eta_i-\teta_i)^2
    \\
    &=- \frac{1}{2}(\veta-\tveta)^T\tmW^{-1}(\veta-\tveta) + \tvu^T(\veta-\tveta) + \log p(\vy|\theta(\tveta)).
    \label{A:eqn:obs-taylor}
    \end{align}
Substituting \refeqn{A:eqn:obs-taylor} into \refeqn{eqn:cfa:joint}, we obtain an approximation to the joint posterior,  %
    \begin{align}
    \begin{split}
    \log q(\vy,\veta|\mX)
    & =
     \log p(\vy|\theta(\tveta))
    - \frac{1}{2} \log \detbar{\mK} - \frac{n}{2}\log 2\pi
    \\
    &\quad\quad
    -
    \frac{1}{2}(\veta-\tveta)^T\tmW^{-1}(\veta-\tveta) + \tvu^T(\veta-\tveta)
    - \frac{1}{2} \veta^T \mK^{-1}\veta
    \end{split}
    \\
    \begin{split}
    & =
     \log p(\vy|\theta(\tveta))
    - \frac{1}{2} \log \detbar{\mK} - \frac{n}{2}\log 2\pi
    \\
    &\quad\quad
    \underbrace{-
    \frac{1}{2}(\veta-\tveta-\tmW\tvu)^T\tmW^{-1}(\veta-\tveta-\tmW\tvu)
    - \frac{1}{2} \veta^T \mK^{-1}\veta}
    + \frac{1}{2}\tvu^T\tmW\tvu.
    \end{split}
    \label{A:eqn:taylor-deriv-1}
    \end{align}
Next, we note that the bracketed term in
\refeqn{A:eqn:taylor-deriv-1} is of the form
    \begin{align}
    \lefteqn{
    (\vx-\va)^T\mB(\vx-\va) + \vx^T\mC\vx =
    \vx^T\mD\vx     - 2\vx^T\mB\va + \va^T\mB\va
    }
    \\ &=
    \vx^T\mD \vx    - 2\vx^T\mD\mD^{-1}\mB\va
    +\va^T\mB^T\mD^{-1}\mB\va
    -\va^T\mB^T\mD^{-1}\mB\va
    + \va^T\mB\va
    \\&=
    \norm{\vx - \mD^{-1}\mB\va}^2_{\mD^{-1}}
    +\va^T(\mB-\mB^T\mD^{-1}\mB)\va
    =
    \norm{\vx - \mD^{-1}\mB\va}^2_{\mD^{-1}}
    +\norm{\va}_{\mB^{-1}+\mC^{-1}}^2
    \end{align}
where $\mD = \mB+\mC$, and the last line uses the matrix inversion
lemma.  Using this property ($\vx=\veta$, $\va = \tveta+\tmW\tvu$,
$\mB=\mW^{-1}$, $\mC=\mK^{-1}$), the joint likelihood can be
rewritten as \refeqn{eqn:cfa:approxjoint}.

\comments{
    \begin{align}
    \begin{split}
    \log q(\vy,\veta|\mX)
    &=
     \log p(\vy|\theta(\tveta))
    - \frac{1}{2} \log \detbar{\mK} - \frac{n}{2}\log 2\pi
    \\
    &\quad
    -\frac{1}{2}\norm{\veta - \mA^{-1}\tmW^{-1} \tvt}^2_{\mA^{-1}}
    -\frac{1}{2}\norm{\tvt}_{\tmW+\mK}^2
    + \frac{1}{2}\tvu^T\tmW\tvu
    \end{split}
    \label{eqn:cfa:approxjoint}
    \end{align}
where $\mA = \tmW^{-1}+\mK^{-1}$, and $\tvt = \tveta+\tmW\tvu$ is
the target vector. }

\comments{
\subsection{Approximate Posterior}

Removing terms in \refeqn{eqn:cfa:approxjoint} that are not
dependent on $\veta$, the approximate posterior of $\veta$ is
    \begin{align}
    \log q(\veta|\mX, \vy) &\propto \log q(\vy,\veta|\mX)
    \propto
    -\frac{1}{2}\norm{\veta - \mA^{-1}\tmW^{-1}\tvt}^2_{\mA^{-1}}
    \\
    \Rightarrow\quad
    q(\veta|\mX, \vy)&=\Normalv{\veta}{\mA^{-1}\tmW^{-1}\tvt}{\mA^{-1}},
    \end{align}
and hence
    \begin{align}
    \hat{\vm} = \hat{\mV}\tmW^{-1}\tvt, \quad
    \hat{\mV} = \left( \tmW^{-1} + \mK^{-1}\right)^{-1}.
    \label{eqn:cfa:post}
    \end{align}
}

\subsection{Special expansion point}
\label{app:Taylor:expansion}

For the case when $\eta_i = g(T(y_i))$, we  note that
    \begin{align}
    T(y_i) - \sd{b}(\theta(g(T(y_i))))  &= T(y_i) - \sd{b}(\sd{b}^{-1}(g^{-1}(g(T(y_i))))) = 0,
    \end{align}
which yields \refeqn{eqn:specialexpansion}.

\comments{
    \begin{align}
    \tu_i &= u(g(y_i),y_i) =  0
    \quad\Rightarrow\quad
    \tilde{t}_i = \teta_i = g(y_i),
    \\
    \tw_i &=
    w(g(y_i),y_i) %
    =\frac{a(\phi)}{\sdd{b}(\theta(g(y_i))) \sd{\theta}(g(y_i))^2},
    \end{align}
}

\comments{
    \begin{align}
    [g^{-1}]'(g(y)) &= \frac{1}{g'(g^{-1}(g(y)))} = \frac{1}{g'(y)} ,
    \\
    \theta(g(y)) &= [b']^{-1}(y) ,
    \\
    \theta'(g(y)) &= [[b']^{-1}]'(y) [g^{-1}]'(g(y)) = \frac{[[b']^{-1}]'(y)}{g'(y)}  ,
    \\
    b''([b']^{-1}(y)) &= \frac{1}{[[b']^{-1}]'(y)}.
    \end{align}
}

\subsection{Approximate Marginal}
\label{app:Taylor:marginal}

The approximate marginal is obtained by substituting the approximate
joint in \refeqn{eqn:cfa:approxjoint}
    \begin{align}
    \lefteqn{
    \log p(\vy|\mX) =  \log \int \exp (\log p(\vy,\veta|\mX))d\veta
    \approx \log \int \exp (\log q(\vy,\veta|\mX))d\veta
    }
    \\
    \begin{split}
    &=
     \log p(\vy|\theta(\tveta))
    - \frac{1}{2} \log \detbar{\mK} - \frac{n}{2}\log 2\pi
    \\
    &\quad
    -\frac{1}{2}\norm{\tvt}_{\tmW+\mK}^2
    + \frac{1}{2}\tvu^T\tmW\tvu
    + \log\int e^{-\frac{1}{2}\norm{\veta - \mA^{-1}\tmW^{-1}\tvt}^2_{\mA^{-1}}}d\veta
    \end{split}
    \\
    &=
     \log p(\vy|\theta(\tveta))
    - \frac{1}{2} \log \detbar{\mK} - \frac{n}{2}\log 2\pi
    -\frac{1}{2}\norm{\tvt}_{\tmW+\mK}^2
    + \frac{1}{2}\tvu^T\tmW\tvu
    + \log (2\pi)^{\frac{n}{2}} \detbar{\mA^{-1}}^{\frac{1}{2}}
    \\
    &=
     \log p(\vy|\theta(\tveta))
    - \frac{1}{2} \log \detbar{\mK}\detbar{\mA}
    -\frac{1}{2}\norm{\tvt}_{\tmW+\mK}^2
    + \frac{1}{2}\tvu^T\tmW\tvu
    \end{align}
Looking at the determinant term,
    \begin{align}
    \log \detbar{\mA}\detbar{\mK}
    &= \log \detbar{(\tmW^{-1} + \mK^{-1}) \mK}
    = \log \detbar{\mI + \tmW^{-1}\mK}
    = \log \detbar{\tmW + \mK}\detbar{\tmW^{-1}}.
    \end{align}
Hence, the approximate marginal is \refeqn{eqn:cfa:marginal}.
\comments{
    \begin{align}
    \log q(\vy|\mX)
    &=
    -\frac{1}{2} \tvt^T(\tmW + \mK)^{-1}\tvt
    - \frac{1}{2} \log \detbar{\tmW + \mK} + r(\phi)
    \label{A:eqn:cfa:marginal}
    \end{align}
where
    \begin{align}
    r(\phi) =
    \log p(\vy|\theta(\tveta))
    + \frac{1}{2}\tvu^T\tmW\tvu
    + \frac{1}{2}\log \detbar{\tmW}.
    \end{align}
} The dispersion penalty can be further rewritten as
    $r(\phi) = \sum_{i=1}^n r_i(\phi)$,
where for an individual data point, we have
    \begin{align}
    r_i(\phi) =
        \log p(y_i|\theta(\teta_i))
        + \frac{1}{2}\tw_i \tu_i^2
        + \frac{1}{2}\log \detbar{\tw_i}.
    \end{align}

\subsubsection{Derivatives wrt hyperparameters}

The derivative of \refeqn{eqn:cfa:marginal} with respect to the
kernel hyperparameter $\alpha_j$ is
    \begin{align}
    \pdd{}{\alpha_j}    \log q(\vy|\mX) &=
    \frac{1}{2} \tvt^T\left(\mK + \tmW\right)^{-1} \pdd{\mK}{\alpha_j} \left(\mK + \tmW\right)^{-1} \tvt
    -\frac{1}{2} \tr\left[(\mK + \tmW)^{-1} \pdd{\mK}{\alpha_j}\right]
    \\
    &=\frac{1}{2} \tr\left[
    \left(   \vz \vz^T - (\mK + \tmW)^{-1}\right)
     \pdd{\mK}{\alpha_j}
    \right], \quad \vz = (\mK + \tmW)^{-1}\tvt.
    \end{align}
where $\pdd{\mK}{\alpha_j}$ is the element-wise derivative of the
kernel matrix with respect to the kernel hyperparameter $\alpha_j$,
and we use the derivative properties,
    \begin{align}
    \pdd{}{\alpha}\mA^{-1} = -\mA^{-1}\pdd{\mA}{\alpha} \mA^{-1},\quad\quad
    \pdd{}{\alpha}\log \detbar{\mA} = \tr(\mA^{-1} \pdd{\mA}{\alpha}).
    \end{align}

\subsubsection{Derivative wrt dispersion}

For the derivative with respect to the dispersion parameter, we
first note that
    \begin{align}
    \pdd{}{\phi}\tw_i &= \sd{a}(\phi)\left\{ \sdd{b}(\theta(\teta_i)) \sd{\theta}(\teta_i)^2
    -   \left[T(y_i) - \sd{b}(\theta(\teta_i))\right]\sdd{\theta}(\teta_i) \right\}^{-1}
 = \frac{\sd{a}(\phi)}{a(\phi)} \tw_i,
    \\
    \pdd{}{\phi}\tmW &= \frac{\sd{a}(\phi)}{a(\phi)} \tmW,
    \\
    \pdd{}{\phi}\tu_i &= -\frac{\sd{a}(\phi)}{a(\phi)^2} \sd{\theta}(\teta_i)\left[T(y_i) - \sd{b}(\theta(\teta_i))\right]
    = -\frac{\sd{a}(\phi)}{a(\phi)} \tu_i,
    \\
    \pdd{}{\phi}\tw_i \tu_i^2 &= \frac{\sd{a}(\phi)}{a(\phi)} \tw_i\tu_i^2 + 2 \tw_i \tu_i \left(-\frac{\sd{a}(\phi)}{a(\phi)} \tu_i\right) = -\frac{\sd{a}(\phi)}{a(\phi)} \tw_i\tu_i^2,
    \\
    \pdd{}{\phi}\log p(y_i|\theta(\teta_i))
    &=
    -\frac{\sd{a}(\phi)}{a(\phi)^2} \left[T(y_i)\theta(\teta_i) - b(\theta(\teta_i)) \right] + \sd{c}(\phi, y_i)
    = -\frac{\sd{a}(\phi)}{a(\phi)} \tv_i + \sd{c}(\phi, y_i),
    \label{A:eqn:cfa:marg:deriv:logp}
    \end{align}
where $\tv_i = \frac{1}{a(\phi)} \left[y_i\theta(\teta_i) -
b(\theta(\teta_i)) \right]$, and $\sd{c}(\phi, y_i) = \pdd{}{\phi}
c(\phi, y_i)$. Thus,
    \begin{align}
    \pdd{}{\phi} r_i(\phi) &= -\frac{\sd{a}(\phi)}{a(\phi)}\tv_i
    + \sd{c}(\phi, y_i) - \frac{1}{2}\frac{\sd{a}(\phi)}{a(\phi)} \tw_i\tu_i^2 + \frac{1}{2}\frac{1}{\tw_i}\frac{\sd{a}(\phi)}{a(\phi)} \tw_i
    \\
    &= \frac{\sd{a}(\phi)}{a(\phi)}\left(\frac{1}{2} - \tv_i - \frac{1}{2}\tw_i\tu_i^2\right)
    + \sd{c}(\phi, y_i).
    \end{align}
Summing over $i$,
    \begin{align}
    \pdd{}{\phi}r(\phi) &= \sum_i \frac{\sd{a}(\phi)}{a(\phi)}\left(\frac{1}{2} - \tv_i - \frac{1}{2}\tw_i\tu_i^2\right) + \sd{c}(\phi, y_i)
    \\
    &= \frac{\sd{a}(\phi)}{a(\phi)} \left(\frac{n}{2} - \vone^T\tvv - \frac{1}{2} \tvu^T\tmW\tvu\right)
    + \sum_{i=1}^n \sd{c}(\phi, y_i).
    \end{align}
Also note that $\tvt$ is not a function of $\phi$, as the term
cancels out in $\tmW\tvu$. Finally,
    \begin{align}
    \begin{split}
    \lefteqn{
    \pdd{}{\phi}\log q(\vy|\mX)}
    \\ &=
    \frac{1}{2} \tvt^T\left(\mK + \tmW\right)^{-1} \pdd{\tmW}{\phi} \left(\mK + \tmW\right)^{-1} \tvt
    -\frac{1}{2} \tr((\mK + \tmW)^{-1} \pdd{\tmW}{\phi}) + \pdd{}{\phi} r(\phi)
    \end{split}
    \\
    &=\frac{1}{2} \tr\left[
    \left(   \vz \vz^T - (\mK + \tmW)^{-1}\right)
     \pdd{\tmW}{\phi}
    \right] +  \frac{\sd{a}(\phi)}{a(\phi)} \left(\frac{n}{2} - \vone^T\tvv - \frac{1}{2} \tvu^T\tmW\tvu\right) + \sum_{i=1}^n \sd{c}(\phi, y_i)
        \\
    &=\frac{\sd{a}(\phi)}{a(\phi)} \left\{
    \frac{1}{2} \tr\left[
    \left(   \vz \vz^T - (\mK + \tmW)^{-1}\right)\tmW \right] +  \frac{n}{2} - \vone^T\tvv - \frac{1}{2} \tvu^T\tmW\tvu  \right\}
    + \sum_{i=1}^n \sd{c}(\phi, y_i)
    \\
    &=\frac{\sd{a}(\phi)}{a(\phi)} \left\{
    \frac{1}{2} \vz^T\tmW \vz - \frac{1}{2} \tr \left[(\mK + \tmW)^{-1}\tmW\right]
    +  \frac{n}{2} - \vone^T\tvv - \frac{1}{2} \tvu^T\tmW\tvu  \right\}
    + \sum_{i=1}^n\sd{c}(\phi, y_i)
    \end{align}

\subsubsection{Derivative wrt dispersion when $b_{\phi}(\theta)$}
\label{app:taylor_bphi}

We next look at the special case where the term $b_\phi(\theta)$ is
also a function of $\phi$. We first note that
    \begin{align}
    \begin{split}
    \pdd{}{\phi}\tw_i &= \sd{a}(\phi)\left\{ \sdd{b}_{\phi}(\theta(\teta_i)) \sd{\theta}(\teta_i)^2
    -   \left[T(y_i) - \sd{b}_{\phi}(\theta(\teta_i))\right]\sdd{\theta}(\teta_i) \right\}^{-1}
    \\
    &\quad
    -
    a(\phi)
     \frac{\sd{\theta}(\teta_i)^2\pdd{}{\phi} \sdd{b}_{\phi}(\theta(\teta_i))
    +  \sdd{\theta}(\teta_i) \pdd{}{\phi}\sd{b}_{\phi}(\theta(\teta_i)) }{\left\{ \sdd{b}_{\phi}(\theta(\teta_i)) \sd{\theta}(\teta_i)^2
    -   \left[T(y_i) - \sd{b}_{\phi}(\theta(\teta_i))\right]\sdd{\theta}(\teta_i) \right\}^{2}}
    \end{split}
    \\
    &= \frac{\sd{a}(\phi)}{a(\phi)} \tw_i - \frac{1}{a(\phi)}\tw_i^2\left[\sd{\theta}(\teta_i)^2\pdd{}{\phi} \sdd{b}_{\phi}(\theta(\teta_i))
    +  \sdd{\theta}(\teta_i) \pdd{}{\phi}\sd{b}_{\phi}(\theta(\teta_i))\right],
    \\
    \pdd{}{\phi}\tu_i &= -\frac{\sd{a}(\phi)}{a(\phi)^2} \sd{\theta}(\teta_i)\left[T(y_i) - \sd{b}_{\phi}(\theta(\teta_i))\right]
    -\frac{\sd{\theta}(\teta_i)}{a(\phi)}\pdd{}{\phi}\sd{b}_{\phi}(\theta(\teta_i))
    = -\frac{\sd{a}(\phi)}{a(\phi)} \tu_i -\frac{\sd{\theta}(\teta_i)}{a(\phi)}\pdd{}{\phi}\sd{b}_{\phi}(\theta(\teta_i)),
    \\
    \pdd{}{\phi}(\tw_i \tu_i^2) &=
    \pdd{\tw_i}{\phi} \tu_i^2 +2 \tw_i \tu_i\pdd{\tu_i}{\phi},
    \quad
    \pdd{}{\phi}\tilde{t}_i = \pdd{}{\phi}(\teta_i + \tw_i\tu_i)
    = \tw_i\pdd{\tu_i}{\phi} + \tu_i\pdd{\tw_i}{\phi},
    \end{align}
    \begin{align}
    \pdd{}{\phi}\log p(y_i|\theta(\teta_i))
    &=
    -\frac{\sd{a}(\phi)}{a(\phi)^2} \left[T(y_i)\theta(\teta_i) - b_{\phi}(\theta(\teta_i)) \right]
    -\frac{1}{a(\phi)}\pdd{}{\phi}b_{\phi}(\theta(\teta_i))
    + \sd{c}(\phi, y_i)
    \\
    &= -\frac{\sd{a}(\phi)}{a(\phi)} \tv_i + \sd{c}(\phi, y_i)
        -\frac{1}{a(\phi)}\pdd{}{\phi}b_{\phi}(\theta(\teta_i)).
    \label{A:eqn:cfa:marg:deriv:logp2}
    \end{align}
Thus,
    \begin{align}
    \pdd{}{\phi} r_i(\phi) &=
    \pdd{}{\phi}\log p(y_i|\theta(\teta_i))
    + \frac{1}{2}\left[\pdd{\tw_i}{\phi} \tu_i^2 +2 \tw_i \tu_i\pdd{\tu_i}{\phi}\right]
     + \frac{1}{2}\frac{1}{\tw_i}\pdd{\tw_i}{\phi}
    \end{align}
For the first term in \refeqn{eqn:cfa:marginal},
    \begin{align}
    \pdd{}{\phi} \left[\tvt^T(\tmW + \mK)^{-1}\tvt\right]
    &=
    \tvt^T\pdd{(\tmW + \mK)^{-1}}{\phi}\tvt
    +2
    \tvt^T(\tmW + \mK)^{-1}\pdd{\tvt}{\phi}
\\
    &=
    -\tvt^T\left(\mK + \tmW\right)^{-1} \pdd{\tmW}{\phi} \left(\mK + \tmW\right)^{-1} \tvt
    +2
    \tvt^T(\tmW + \mK)^{-1}\pdd{\tvt}{\phi}.
    \end{align}
For the second term in \refeqn{eqn:cfa:marginal},
    \begin{align}
    \pdd{}{\phi} \log \detbar{\mK + \tmW} =
    \tr((\mK + \tmW)^{-1} \pdd{\tmW}{\phi}) .
    \end{align}
Finally, we have
    \begin{align}
    \begin{split}
    \pdd{}{\phi}\log q(\vy|\mX)
    &=
    \frac{1}{2}\tvt^T\left(\mK + \tmW\right)^{-1} \pdd{\tmW}{\phi} \left(\mK + \tmW\right)^{-1} \tvt
    -
    \tvt^T(\tmW + \mK)^{-1}\pdd{\tvt}{\phi}
    \\
    &\quad -\frac{1}{2} \tr((\mK + \tmW)^{-1} \pdd{\tmW}{\phi})
    +\sum_i \pdd{}{\phi}r_i(\phi)
    .
    \end{split}
    \end{align}

\section{Taylor approximation for specific GGPMs}\label{text:Taylor:DeFun}

This appendix contains the derivatives and target functions used to derive the special cases of Taylor approximation in Section \ref{text:specialcases}.

\subsubsubsection{Binomial/Bernoulli}
The derivative functions are
    \begin{align}
    u(\eta,y) = N(y-\tfrac{e^\eta}{1+e^\eta}),
    \ \
    w(\eta,y) %
    = \tfrac{(1+e^\eta)^2}{N e^\eta}.
    \nonumber
    \end{align}
Thus, for a given expansion point $\teta_i$, the target and effective noise are
    \begin{align}
    \tilde{t}_i = \teta_i + \tfrac{(1+e^{\teta_i})^2}{e^{\teta_i}} (y_i - \tfrac{e^{\teta_i}}{1+e^{\teta_i}} ), \
    \tilde{w}_i = \tfrac{(1+e^{\teta_i})^2}{N e^{\teta_i}}.
    \nonumber
    \end{align}

\subsubsubsection{Poisson}
The
derivative functions are
    \begin{align}
    u(\eta,y) = y-e^{\eta},
    \ \
    w(\eta,y) %
    = e^{-\eta}.
    \nonumber
    \end{align}
Thus, given an expansion point $\teta_i$, the target and effective noise are
    \begin{align}
    \tilde{t}_i = \teta_i + (y_i e^{-\teta_i}-1), \quad
    \tilde{w}_i = e^{-\teta_i}.
    \end{align}

\subsubsubsection{Gamma}
The derivatives of the Gamma\shape likelihood (mean parameter, shape hyperparameter) are
    \begin{align}
    u(\eta,y) = \nu e^{-\eta}(y-e^{\eta}) = \nu(ye^{-\eta}-1),
    \ \
    w(\eta,y) %
    = \frac{1}{\nu}(1+ye^{-\eta}-1)^{-1}=\frac{1}{\nu ye^{-\eta}}.
    \nonumber
    \end{align}
Thus, given an expansion point $\teta_i$, the target and effective noise are
    \begin{align}
    \tilde{t}_i = \teta_i + \tfrac{1}{\nu y_i e^{-\teta_i}}\nu (y_i e^{-\teta_i}-1) = \teta_i+1-\tfrac{1}{y_i e^{-\teta_i}}, \quad
    \tilde{w}_i = \tfrac{1}{\nu y_i e^{-\teta_i}}.
    \end{align}

\subsubsubsection{Inverse Gaussian}
The derivatives of the Inverse Gaussian likelihood are
    \begin{align}
    u(\eta,y) = \phi e^{-\eta} [y-(2e^{-\eta})^{-1/2}],
    \ \
    w(\eta,y)
    = \phi \{(8e^{\eta})^{-1/2}+[y-(2e^{-\eta})^{-1/2}]e^{-\eta}\}^{-1}.
    \nonumber
    \end{align}
Using the canonical expansion point, $\teta_i = \log(2y_i^2)$,
yields
\begin{align}
    u(\teta_i,y_i) =    %
     0,
    \ \
    w(\teta_i,y_i) %
     = 4\phi y_i.
    \end{align}
\subsubsubsection{Beta}
Consider an agnostic choice of the expansion point, $\teta_i=0$, and hence $\ttheta_i = \theta(\teta_i) = \frac{1}{2}$.  We also have,
    \begin{align}
    \sd{\theta}(\teta_i) = \frac{e^{\teta_i}}{(1+e^{\teta_i})^2} = \frac{1}{4},
    \quad
    \sdd{\theta}(\teta_i) = \frac{e^{\teta_i}(e^{\teta_i}-1)}{(1+e^{\teta_i})^3} = 0.
    \end{align}
Looking at the 1st and 2nd derivatives of $b(\theta)$ at $\ttheta_i$, we have
    \begin{align}
    \sd{b}(\ttheta_i) &= \psi_0(\tfrac{\ttheta_i}{\phi}) - \psi_0(\tfrac{1-\ttheta_i}{\phi})
    = \psi_0(\tfrac{1}{2\phi}) - \psi_0(\tfrac{1}{2\phi}) = 0, \\
    \sdd{b}(\ttheta_i) &= \tfrac{1}{\phi}\psi_1(\tfrac{\ttheta_i}{\phi}) + \tfrac{1}{\phi}\psi_1(\tfrac{1-\ttheta_i}{\phi})
    = \tfrac{2}{\phi} \psi_1(\tfrac{1}{2\phi}).
    \end{align}
Using the above results, we can now calculate the derivative functions at $\teta_i=0$,
    \begin{align}
    \tu_i = u(\teta_i,y_i) &= \frac{1}{\phi} \sd{\theta}(\teta_i) \left[ T(y_i) - \sd{b}(\ttheta_i)\right]
        = \frac{1}{4\phi} T(y_i) = \frac{1}{4\phi}\log \frac{y_i}{1-y_i},
        \\
    \tw_i = w(\teta_i,y_i) &= \frac{\phi}{\sdd{b}(\ttheta_i) \sd{\theta}(\teta_i)^2 - 0}
        = \frac{\phi}{\tfrac{2}{\phi}\psi_1(\tfrac{1}{2\phi})\tfrac{1}{4^2}}
        = \frac{8\phi^2}{\psi_1(\tfrac{1}{2\phi})}.
    \end{align}
which yields the targets,
    \begin{align}
    \tilde{t}_i = \teta_i + \tw_i \tu_i = \frac{2\phi}{\psi_1(\tfrac{1}{2\phi})} \log \frac{y_i}{1-y_i}.
    \end{align}

\section*{Acknowledgements}
The authors thank CE Rasmussen and CKI Williams for  the GPML code \citep{GPMLcode}.
This work was supported by City University of Hong Kong (internal grant 7200187), and
by the Research Grants Council of the Hong Kong Special Administrative Region, China (CityU 110610 and CityU 123212)

\vskip 0.2in
\bibliography{abc_all,GGPRrefs}

\end{document}